\pdfoutput=1

\documentclass[11pt]{article}

\usepackage[preprint]{acl}

\usepackage{times}
\usepackage{latexsym}

\usepackage[T1]{fontenc}

\usepackage[utf8]{inputenc}

\usepackage{microtype}

\usepackage{inconsolata}

\usepackage{array}
\usepackage{xspace}
\usepackage{booktabs}
\usepackage{amsmath}
\usepackage{amssymb}
\usepackage{graphicx}
\usepackage{adjustbox}
\usepackage{multirow}
\usepackage{comment}
\usepackage{color, colortbl}
\usepackage{arydshln}
\usepackage{hhline}
\usepackage{array}
\usepackage{makecell}

\usepackage{tcolorbox}
\usepackage{graphicx}
\usepackage{enumitem}
\usepackage{adjustbox}
\usepackage{xcolor}
\usepackage{float} 
\usepackage{caption} 

\newtcolorbox{promptbox}[1]{colback=white!95!black,colframe=black,fonttitle=\bfseries\small,fontupper=\footnotesize,title=#1}

\newcommand{\rparagraph}[1]{\vspace{1.2mm}\noindent\textbf{#1.}}
\newcommand{\iparagraph}[1]{\vspace{0.7mm}\noindent\textit{#1.}}

\usepackage{xspace}

\newcommand{\its}{{\textsc{i2s}}\xspace}
\newcommand{\sti}{{\textsc{s2i}}\xspace}
\newcommand{\tts}{{\textsc{t2s}}\xspace}
\newcommand{\stt}{{\textsc{s2t}}\xspace}

\usepackage{todonotes}

\title{MVL-SIB: A Massively Multilingual Vision-Language Benchmark for Cross-Modal Topical Matching}

  \author{%
  Fabian David Schmidt\textsuperscript{1}\thanks{Equal contribution.}, 
  Florian Schneider\textsuperscript{2}\footnotemark[1], 
  Chris Biemann\textsuperscript{2}, 
  Goran Glavaš\textsuperscript{1} \\
  \textsuperscript{1}Center for Artificial Intelligence and Data Science, University of Würzburg, Germany \\
  \textsuperscript{2}Language Technology Group, University of Hamburg, Germany \\
  \texttt{fabian.schmidt@uni-wuerzburg.de, florian.schneider-1@uni-hamburg.de}\\
  \textbf{Dataset:} \href{https://huggingface.co/datasets/WueNLP/mvl-sib}{MVL-SIB}
}

\begin{document}
\maketitle
\begin{abstract}

Existing multilingual vision-language (VL) benchmarks often only cover a handful of languages. Consequently, evaluations of large vision-language models (LVLMs) predominantly target high-resource languages, underscoring the need for evaluation data for low-resource languages. To address this limitation, we introduce MVL-SIB, a massively multilingual vision-language benchmark that evaluates both cross-modal and text-only topical matching across 205 languages---over 100 more than the most multilingual existing VL benchmarks encompass. 
We then benchmark a range of of open-weight LVLMs together with GPT-4o(-mini) on MVL-SIB. Our results reveal that LVLMs struggle in cross-modal topic matching in lower-resource languages, performing no better than chance on languages like N'Koo. Our analysis further reveals that VL support in LVLMs declines disproportionately relative to textual support for lower-resource languages, as evidenced by comparison of cross-modal and text-only topical matching performance. 
We further observe that open-weight LVLMs do not benefit from representing a topic with more than one image, suggesting that these models are not yet fully effective at handling multi-image tasks.
By correlating performance on MVL-SIB with other multilingual VL benchmarks, we highlight that MVL-SIB serves as a comprehensive probe of multilingual VL understanding in LVLMs.
\end{abstract}
\section{Introduction}
\label{sec:intro}

\begin{figure}[ht]
\begin{adjustbox}{width=\columnwidth,center}
\begin{tcolorbox}[
    colback=white!95!black,    
    colframe=black,            
    title=Images-To-Sentences (\its), 
    fonttitle=\bfseries\Large,       
    boxrule=0.5pt,             
    arc=4pt,                   
    outer arc=4pt,
    width=\textwidth,          
    boxsep=1pt,
    left=7pt,
    right=7pt,
    enlarge left by=0mm,
    enlarge right by=0mm,
    before skip=0.5em,           
    after skip=0.5em,            
]

Which sentence best matches the topic of the images? The images and the sentences each belong
to one of the following topics: "entertainment", "geography", "health", "politics", "science and technology", "sports", or "travel". Choose one sentence from A, B, C, or D. Output only 
a single letter!

\medskip

\# Images

\begin{center}
    \begin{minipage}{0.18\textwidth}
        \centering
        \includegraphics[width=\linewidth]{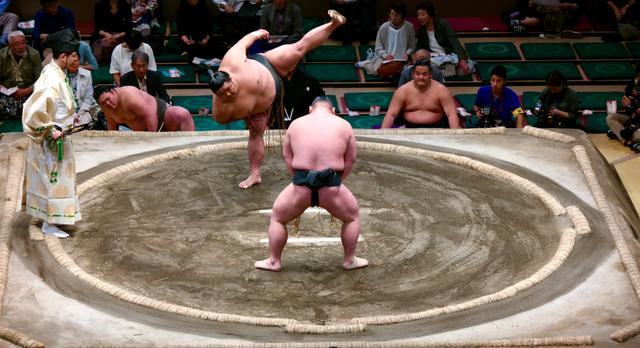}
    \end{minipage}
    \hfill
    \begin{minipage}{0.18\textwidth}
        \centering
        \includegraphics[width=\linewidth]{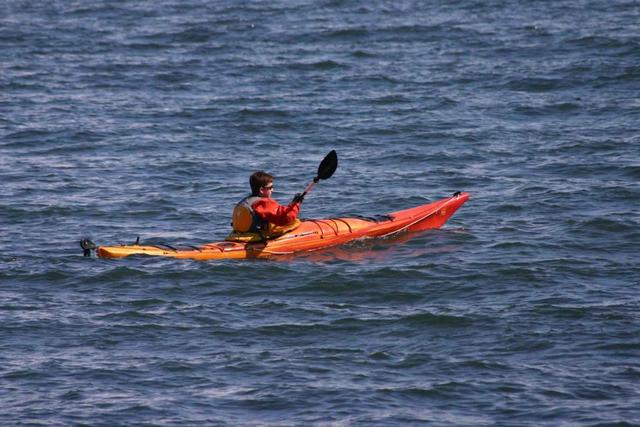}
    \end{minipage}
    \hfill
    \begin{minipage}{0.18\textwidth}
        \centering
        \includegraphics[width=\linewidth]{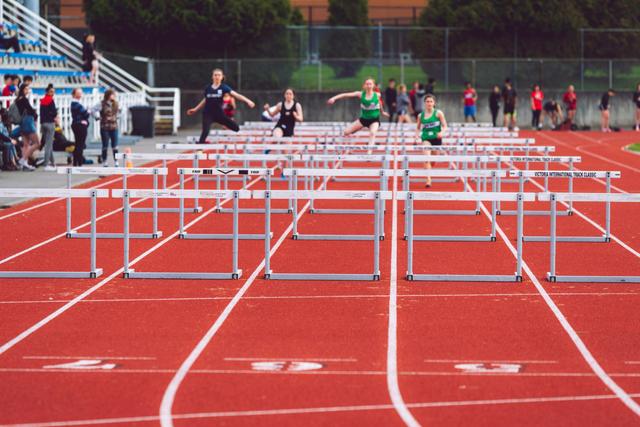}
    \end{minipage}
    \hfill
    \begin{minipage}{0.18\textwidth}
        \centering
        \includegraphics[width=\linewidth]{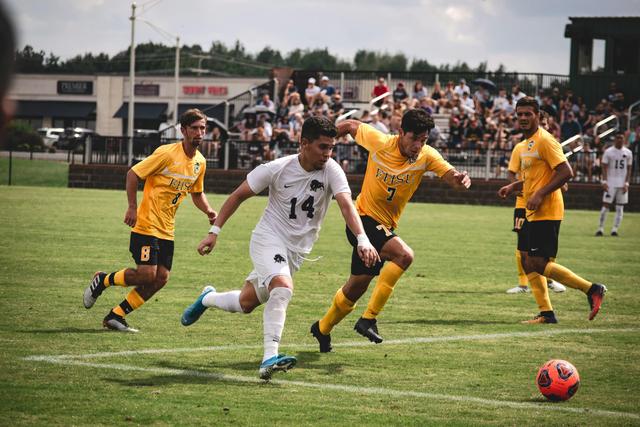}
    \end{minipage}
    \hfill
    \begin{minipage}{0.18\textwidth}
        \centering
        \includegraphics[width=\linewidth]{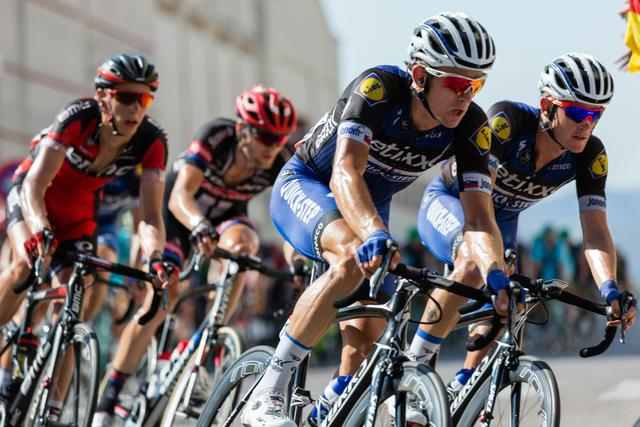}
    \end{minipage}
\end{center}

\medskip

\# Sentences

\begin{enumerate}[label=\Alph*., itemsep=0pt, topsep=0pt]
    \item \textasciigrave\textasciigrave\textasciigrave Maroochydore führte am Ende die Rangfolge an, mit sechs Punkten Vorsprung vor Noosa als Zweitem.\textasciigrave\textasciigrave\textasciigrave
    \item \textasciigrave\textasciigrave\textasciigrave 
    Es wurden keine schwere Verletzungen gemeldet, jedoch mussten mindestens fünf der zur Zeit der Explosion Anwesenden aufgrund von Schocksymptomen behandelt werden.\textasciigrave\textasciigrave\textasciigrave
    \item \textasciigrave\textasciigrave\textasciigrave 
    Finnland ist ein großartiges Reiseziel für Bootstouren. Das „Land der tausend Seen“ hat auch Tausende von Inseln – in den Seen und in den Küstenarchipelen.\textasciigrave\textasciigrave\textasciigrave
    \item \textasciigrave\textasciigrave\textasciigrave 
    Es ist auch nicht erforderlich, dass Sie eine lokale Nummer von der Gemeinde erhalten, in der Sie leben. Sie können eine Internetverbindung über Satellit in der Wildnis v on Chicken in Alaska erhalten und eine Nummer auswählen, die vorgibt, dass Sie im sonnigen Arizona sind.\textasciigrave\textasciigrave\textasciigrave
\end{enumerate}

Your answer letter: 
\end{tcolorbox}
\end{adjustbox}
\vspace{-0.7cm}
\caption{Cross-modal topic matching `Images-To-Sentence' for German with $k{=}5$ reference images.} 
\vspace{-0.8cm}
\label{fig:intro-prompt-i2s}
\end{figure}

 \begin{figure*}[ht]
   \centering
   \adjustbox{max width=\textwidth,trim={0 0 0 0.4cm},clip}{%
     \includegraphics{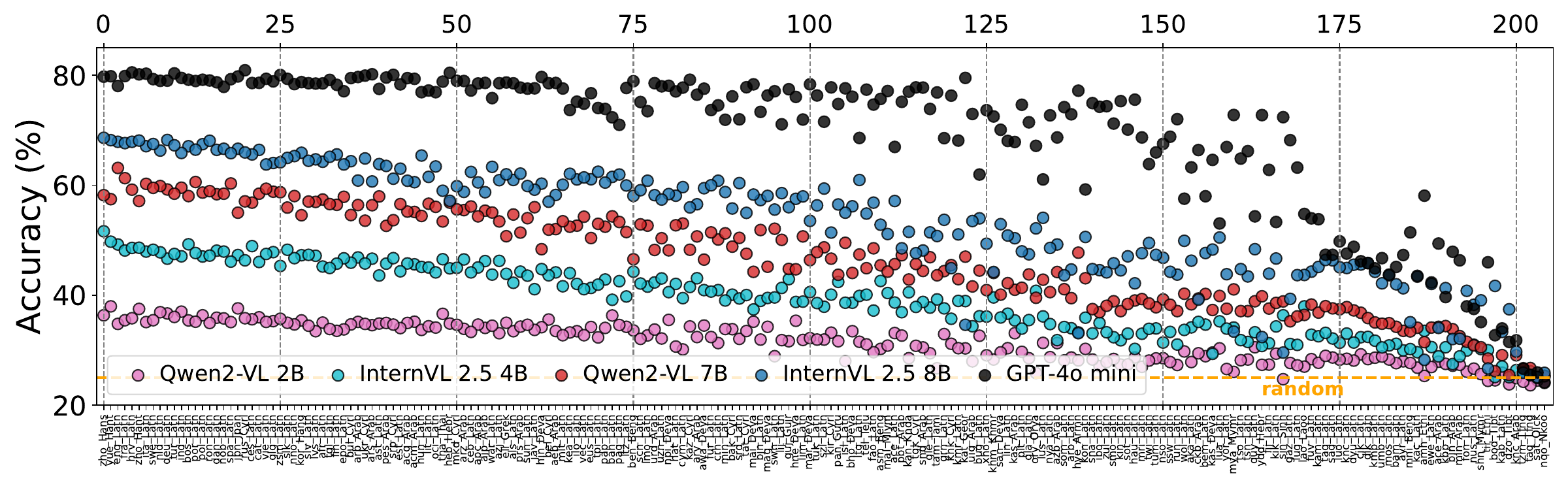}%
   }
   \vspace{-0.8cm}
   \caption{\textbf{Images-To-Sentences~@~$k{=}3$.} The English prompt describes the cross-modal topic matching task, lists all topics, and provides both $k{=}3$ reference images and 4 sentences in the corresponding language \{\texttt{eng\_Latn}, $\dots$, \texttt{nqo\_Nkoo}\}. LVLMs must select the sentence of 4 options that topically fits $k{=}3$ reference images. The sentences spanning 205 languages and 7 topics are drawn from SIB-200 \cite{adelani-etal-2024-sib}, while images for the topics were hand-selected (cf. Appendix \ref{app:images-per-topic}). An example prompt is shown in Appendix \ref{app:images-to-sentences}; further details are in \S\ref{sec:experimental-setup}. \\ \textbf{Plot.} The x-axis orders the languages of the candidate sentences \{\texttt{eng\_Latn}, $\dots$, \texttt{nqo\_Nkoo}\}, respectively, by descending performance (y-axis). The top x-axis indicates the running index of each language $L_i$ ($i \in \{1, \dots, 205\}$).}
   \vspace{-0.3cm}
   \label{fig:intro-img2sent}
 \end{figure*}

Large Vision-Language Models (LVLMs) extend Large Language Models (LLMs) to take images as inputs, leveraging their advanced language capabilities for vision-language (VL) tasks like image captioning and visual question answering (VQA). However, LVLMs are typically trained mainly on English data, leading to significant limitations despite the base LLMs' multilingual abilities. They may fail to follow instructions or struggle to interpret text within images in Non-English languages \cite{schneider-sitaram-2024-m5,tang2024mtvqa}. \\
Although many multilingual VL benchmarks exist, they typically cover at most 10 languages \cite[\textit{inter alia}]{bugliarello-etal-2022-iglue,liu-etal-2021-visually,tang2024mtvqa}. Only concurrent work has scaled VL evaluation to 100 languages using machine translation (MT) with human post-editing \cite{vayani2024alm}. Nevertheless, benchmarks constructed using semi-manual MT cannot support truly low-resource languages adequately, as current MT models lack the necessary quality for these languages. Moreover, existing benchmarks primarily assess lower-level VL semantics through concrete text-image relationships, such as those found in VQA. This underscores the need for VL benchmarks that cover truly low-resource languages and evaluate more abstract VL interactions. 

To address these challenges, we introduce the massively multilingual vision-language SIB (MVL-SIB) dataset, which extends the topic labels of the multi-way parallel sentences from SIB-200 \cite{adelani-etal-2024-sib} by associating each topic with hand-selected images. MVL-SIB evaluates cross-modal image-text topic matching in 205 languages: LVLMs must select one of 4 candidate sentences that best matches the topic of the reference images (`images-to-sentence', cf. Figure \ref{fig:intro-prompt-i2s}) or, conversely, choose one of 4 candidate images corresponding to the topic of the reference sentences (`sentences-to-image'). Figure~\ref{fig:intro-img2sent} displays the `images-to-sentence' performance for across all languages in MVL-SIB, sorted in descending order.
Notably, GPT-4o-mini performs robustly on the top 125 languages. However, beyond that range its performance declines sharply, falling to chance levels in the lowest-resource languages, such as N'Koo. 
Bridging these gaps is crucial for developing genuinely inclusive VL technology.


\rparagraph{Contributions} \textbf{1)} MVL-SIB supports parallel VL evaluation in 205 languages on professionally translated texts, a 105 more languages than any other VL benchmark. The tasks, images-to-sentence and sentences-to-images with prefix and postfix images in context, respectively, allows for fine-grained analysis of VL interactions. We also define corresponding text-only tasks by replacing the images with the topic label to compare the VL support (by language) of LVLMs against the text-only support of their underlying LLMs. Both tasks allow to vary the number of included images to analyze the shift from single to multi-image support in LVLMs.
\textbf{2)} We thoroughly evaluate LVLMs on cross-modal image-text topic matching, finding that task performance is closely associated with both model size and the size of pre-training corpora of the respective languages. 
We further find that only GPT-4o-mini seizes on multiple references in both cross-modal tasks. Open-weights LVLMs, moreover, favor one of the two tasks, highlighting the asymmetry in their VL support.
\textbf{3)} We analyze the relationship between stand-alone text and vision-language support in LVLMs by also benchmarking LVLMs on text-only topic matching. The performance gap between matching sentences to reference images or the topic tends to be larger the better the LVLM supports the underlying language. Conversely, the spread in performance between picking the fitting image or topic for reference sentences increases the worse the vision-language support of the LVLM for the evaluated language is.
\textbf{4)} We correlate MVL-SIB with established multilingual VL benchmarks on the languages shared between the respective pairs of datasets, showing that the MVL-SIB tasks align well with all VL tasks except OCR. We further show that images-to-sentence and sentence-to-image probe distinct aspects of VL interaction, as certain benchmarks correlate more strongly with one task than with the other. This analysis shows that MVL-SIB constitutes a reliable and comprehensive VL benchmark for the lower-resource languages that are not covered by other datasets.

\section{Related Work}
\label{sec:related}

\rparagraph{Multilingual Vision-Language Models} Researchers have extended more English-centric LVLMs like BLIP-2 or LLaVA by continuing to train on multilingual data. Google’s PaLI models were the earliest closed-weight models trained on multilingual captions and VQA data; their open-weight PaliGemma followed a comparable strategy. Meanwhile, modern LLMs (e.g., Qwen 2.5, Llama 3, Gemma 2, Aya) have improved in multilingual tasks but frequently fail to respond consistently in non-English languages, particularly in low-resource settings \cite{schneider-sitaram-2024-m5}. 
However, foundational models tend to focus on higher-resource languages and do not fully account for broader linguistic contexts in vision-language tasks. mBLIP is the first open multilingual LVLM trained on image captions and a small set of translated instruct data in 98 languages \cite{geigle-etal-2024-mblip}. Pangea incorporates multicultural dimensions by blending machine-translated data, existing multilingual resources, and synthetic data, on 39 languages \cite{yue2024pangea}. Most recently, \citet{geigle2025centurio} studied the composition of training data for multilingual adaptation of LVLMs, observing that only 25-50\% of the data needs to be in English. The authors apply their findings when training `Centurio', which achieves state-of-the-art performance on 14 multilingual VL benchmarks.

\rparagraph{Multilingual Vision-Language Benchmarks}
Existing datasets span VQA, natural language inference (NLI), image captioning, outlier detection, and culturally grounded QA: 

\iparagraph{VQA} xGQA extends the questions of the GQA dataset into 8 languages (5 scripts), but the answers in English \cite{pfeiffer-etal-2022-xgqa}. MaXM offers short-form QA fully in 7 languages, pairing culturally aligned images with same-language QA pairs \cite{changpinyo-etal-2023-maxm}. MTVQA focuses on text-heavy VQA in 9 languages \cite{tang2024mtvqa}.

\iparagraph{Culturally-grounded VQA} CVQA collects culturally diverse images and queries in 31 languages with multiple country-specific variants \cite{romero2024cvqa}. The concurrent ALM-Bench covers 100 languages via MT with GPT-4o followed by human post editing with both generic and cultural multiple-choice and `true or false' questions, as well as free-form VQA \cite{vayani2024alm}.

\iparagraph{Visual Reasoning \& NLI}
XVNLI evaluates cross-lingual visual NLI on 5 languages \cite{bugliarello-etal-2022-iglue}. Binary reasoning tasks include MaRVL \cite{liu-etal-2021-visually} and M5B-VGR \cite{schneider-sitaram-2024-m5}, each using linguistically specific images and textual statements. M5B-VLOD presents an outlier detection challenge, where a statement holds true for all but one image \cite{schneider-sitaram-2024-m5}.

\iparagraph{Multiple-Choice QA}
Babel-ImageNet \cite{geigle-etal-2024-babel} translates ImageNet labels into nearly 300 languages for multiple-choice object classification.  M3Exam and xMMMU also feature multiple-choice VQA in 9 and 7 languages, respectively.

\vspace{1.2mm}
\noindent MVL‐SIB fills the gaps in existing multilingual VL benchmarks. It provides test data professionally translated to 205 languages, covering over 105 more languages than other benchmarks for which MT cannot synthesize reliable data for. Other VL datasets that eschew MT, such as culturally-grounded VQA benchmarks,  typically construct more language-specific non-parallel data, that does not support comparative evaluation across languages. The cross-modal topic matching tasks can be also framed text-only (cf. \S\ref{sec:tasks}), replacing the images that represent topics with the explicit topic labels. Thereby, MVL-SIB enables to ablate the vision-language support from the textual support for a language. Finally, the benchmark allows to vary the number of images provided LVLMs to analyze the support for multi‐image reasoning.

%
\section{Dataset and Tasks}

\subsection{Dataset}
\label{sec:dataset}

For MVL-SIB, we extend the following Flores-based datasets to create a massively multilingual, multi-way parallel VL benchmark for identifying topical associations between images and sentences.

\rparagraph{Flores} Flores is a machine translation benchmark containing 3,001 sentences from English Wikipedia paragraphs \cite{nllbteam2022language}, professionally translated into over 200 languages.\footnote{Flores splits 3,001 sentences into \textsc{dev} (997), \textsc{devtest} (1,012), and \textsc{test} (992) sets. The \textsc{test} set was not released.}

\rparagraph{SIB-200} \citet{adelani-etal-2024-sib} grouped the coarse topical annotations for sentences in the \textsc{dev} and \textsc{devtest} subsets of Flores into 7 higher-level topics.\footnote{The topics are \texttt{entertainment}, \texttt{geography}, \texttt{health}, \texttt{politics}, \texttt{science/technology}, \texttt{sports}, and \texttt{travel}.} The resulting SIB-200 dataset is a benchmark for topical classification with 1,004 parallel examples for 205 language variants.

\rparagraph{MVL-SIB} For each topic, we first manually collect 10 permissively licensed images that distinctly represent the topic (e.g., \texttt{sports}) with minimal overlap or ambiguity (cf. Appendix \ref{app:images-per-topic}). We verify that all LVLMs in our study correctly classify the topics for the images when prompted (cf. Appendix \ref{app:images-to-topics}). We next create 3 different MVL-SIB instances from each of the 1,004 SIB sentences, totaling 3,012 MVL-SIB instances.
For each MVL-SIB instance, we couple the respective SIB sentence with (1) a random selection of 5 positive images (same topic) and 4 additional sentences from the same category as the original sentence, as well as (2) 3 negative images and sentences randomly sampled from different topics compared to the starting SIB sentence. The set of sampled sentences and images by instance is maintained across languages. 

\subsection{Cross-modal \& Text-only Topic Matching}
\label{sec:tasks}

We formulate both cross-modal and text-only topic matching tasks based on MVL-SIB. In every task, we present the model with the list of topics that images and sentences may be associated with.\footnotemark[\value{footnote}] Otherwise, it would be unclear along which dimension the model should match images and sentences. The portion of the prompt that introduces the task is provided in English, while the sentences to be topically aligned with images are presented in one of the 205 languages included in MVL-SIB. LLMs reliably perform tasks described in English, even when task-related information is conveyed in other languages \cite{muennighoff2022crosslingual,romanou2024include}. This ensures a fair comparison across all 205 languages, where MT would not accurately preserve the meaning of the prompts. We detail prompts for each task in Appendix \ref{appendix:prompts}. 

\rparagraph{Cross-modal Topic Matching} 
Using our text-image samples (cf. \S\ref{sec:dataset}), we define two cross-modal topic matching tasks: Images-To-Sentence (\its) and Sentences-To-Image (\sti). In \its, the model must select, from 4 candidates, the sentence that matches the topic of $k$ reference images. Conversely, in \sti, the model chooses, from 4 options, the image that shares the topic with $k$ reference sentences. In both tasks, we present the model with $k \in \{1, 3, 5\}$ references, respectively. These tasks evaluate the model’s ability to align high-level visual and textual cues on topics.

\rparagraph{Text-only Topic Matching} 
We construct two tasks by replacing the images in \its and \sti, that represent the topics, with their corresponding labels (e.g., \texttt{sports}). The resulting unimodal tasks, Topic-To-Sentence (\tts) and Sentences-To-Topic (\stt), mirror the cross-modal tasks, \its and \sti, respectively. For \tts, we evaluate only $k{=}1$, since repeating the topic label adds no information. These baseline tasks allow us to delineate between language support and vision-language understanding in LVLMs.

\rparagraph{MVL-SIB offers 4 crucial advantages over prior benchmarks} \textbf{1)} It supports evaluation in 205 languages, covering over 100 more languages than existing benchmarks for which MT models fail to synthesize reliable evaluation data. \textbf{2)} MVL-SIB supports ablating language understanding and multimodal reasoning of LVLMs by comparatively evaluating the mirroring text-only and cross-modal topic matching tasks (cf. \S\ref{sec:tasks}). \textbf{3)} MVL-SIB enables intricate analysis of single- and multi-image VL interactions in LVLMs by allowing topics to be represented by varying numbers of images in cross-modal tasks. \textbf{4)} MVL-SIB comprises higher-level VL reasoning tasks,  pairing varied images and diverse texts to test nuanced VL understanding. 
\section{Experimental Setup}
\label{sec:experimental-setup}

\rparagraph{Models} We test state-of-the-art LVLMs Qwen2-VL \cite{wang2024qwen2vl}, InternVL 2.5 \cite{chen2024internvl}, Centurio-Qwen \cite{geigle2025centurio}, and GPT-4o(-mini) across available sizes.\footnote{We provide details on the LVLMs in Appendix \ref{app:subsec:exp-details}.} Smaller LVLMs are evaluated on all languages, while larger ones (26B+) are tested on subsets (cf. \S\ref{subsec:further-analyses}). For cross-modal topic matching, we also evaluate on mSigLIP-base \cite{zhai2023sigmoid}. Trained explicitly for semantic similarity on multilingual image-caption pairs, the ViT represents a strong baseline.\footnote{The model is available on Huggingface at: \href{https://huggingface.co/google/siglip-base-patch16-256-multilingual}{google/siglip-base-patch16-256-multilingual}.} Its prediction denotes the choice that has the highest average cosine similarity to the $k$ references.

\rparagraph{Image preprocessing} We downsample the images to 640×480 pixels, as the tasks rely on higher-level visual cues for topically associating images and texts rather than finer image details. This can significantly reduce the number of visual tokens input to LVLMs, enabling more efficient inference.

\rparagraph{Hyperparameters} We decode text greedily with temperature set to $0.0$ to ensure reproducibility.

\rparagraph{Metric} We compute the share of prompts for which responses begin with the right letter. If the label is "A", a response such as "A." is also correct.
\section{Results and Discussion}
\label{sec:results}

\begin{table*}[ht]
  \centering
  \begin{adjustbox}{width=\textwidth,center}
     \setlength{\tabcolsep}{8pt}
    \begin{tabular}{
      @{\extracolsep{\fill}} 
      l  !{\vrule}
      c c c !{\vrule}
      c c c !{\vrule}
      c c c !{\vrule}
      c c c !{\vrule}
      c c c 
    }
      \toprule
       \multicolumn{1}{c}{\textbf{Resourceness}} & \multicolumn{7}{r}{$\xleftarrow{\hspace{3cm}}$ \textbf{High} \rule[0.6ex]{3cm}{0.4pt}} & \multicolumn{1}{c}{\textbf{Mid}} & \multicolumn{7}{l}{\rule[0.6ex]{3cm}{0.4pt} \textbf{Low} $\xrightarrow{\hspace{3cm}}$} \\ \midrule
      \multicolumn{1}{c}{} &
      \multicolumn{3}{c}{\textbf{\textsc{English}}} &
      \multicolumn{3}{c}{\textbf{\textsc{Tier 4} (26)}} &
      \multicolumn{3}{c}{\textbf{\textsc{Tier 3} (32)}} &
      \multicolumn{3}{c}{\textbf{\textsc{Tier 2} (96)}} &
      \multicolumn{3}{c}{\textbf{\textsc{Tier 1} (51)}} \\
      \cmidrule(lr){2-4} \cmidrule(lr){5-7} \cmidrule(lr){8-10} \cmidrule(lr){11-13} \cmidrule(lr){14-16}
      \multicolumn{1}{c !{\vrule}}{\textbf{\textit{k} References}} &
      \textbf{1} &
      \textbf{3} &
      \textbf{5} &
      \textbf{1} &
      \textbf{3} &
      \textbf{5} &
      \textbf{1} &
      \textbf{3} &
      \textbf{5} &
      \textbf{1} &
      \textbf{3} &
      \textbf{5} &
      \textbf{1} &
      \textbf{3} &
      \textbf{5} \\
      \midrule \hline
      \rowcolor{gray!10} 
      \multicolumn{16}{l}{\rule{0pt}{2.5ex} \textbf{\textit{Images-To-Sentence:}} \textit{Select 1 of 4 sentences topically matching $k$ reference images}} \\ \hline

mSigLIP-base & 57.7 & 64.6 & 66.4 & 
53.3 & 58.6 & 59.7 & 
51.4 & 56.2 & 57.1 & 
38.9 & 41.2 & 41.7 & 
36.1 & 37.6 & 38.0 \\ \hline

Qwen2-VL 2B & 36.3 & 34.8 & 34.9 & 
35.5 & 35.3 & 34.1 & 
34.5 & 34.3 & 33.2 & 
31.0 & 30.6 & 30.0 & 
29.5 & 29.0 & 28.6 \\
Qwen2-VL 7B & 65.8 & 63.1 & 58.9 & 
57.7 & 56.5 & 51.7 & 
55.4 & 54.5 & 49.6 & 
44.3 & 44.4 & 40.5 & 
39.6 & 39.7 & 36.5 \\
InternVL 2.5 4B & 52.5 & 49.2 & 48.1 & 
50.3 & 46.6 & 47.7 & 
48.6 & 45.3 & 46.1 & 
38.7 & 37.1 & 37.3 & 
35.4 & 34.5 & 34.5 \\
InternVL 2.5 8B & \underline{67.7} & \underline{67.9} & \underline{68.7} & 
\underline{64.6} & \underline{64.9} & \underline{65.7} & 
\underline{61.2} & \underline{60.8} & \underline{61.6} & 
\underline{51.0} & \underline{51.4} & \underline{51.8} & 
\underline{46.1} & \underline{46.0} & \underline{46.3} \\
Centurio Qwen & 54.8 & 60.0 & 62.4 & 
54.2 & 59.2 & 60.6 & 
53.4 & 58.1 & 58.9 & 
46.6 & 48.9 & 49.2 & 
43.0 & 44.2 & 44.7 \\
GPT-4o-mini & \textbf{68.3} & \textbf{78.1} & \textbf{77.4} & 
\textbf{71.6} & \textbf{79.0} & \textbf{78.1} & 
\textbf{72.0} & \textbf{78.9} & \textbf{77.7} & 
\textbf{63.5} & \textbf{68.0} & \textbf{66.4} & 
\textbf{56.9} & \textbf{60.3} & \textbf{58.7} \\
      \midrule \hline
      \rowcolor{gray!10} 
      \multicolumn{16}{l}{\rule{0pt}{2.5ex} \textbf{\textit{Sentences-To-Image:}} \textit{Select 1 of 4 images topically matching $k$ reference sentences}} \\
      \hline

mSigLIP-base & 56.3 & 66.0 & \underline{69.6} & 
51.8 & 61.6 & \underline{64.0} & 
49.1 & 58.3 & 60.2 & 
36.0 & 40.4 & 41.2 & 
32.9 & 36.3 & 36.9 \\ \hline

Qwen2-VL 2B & 41.9 & 43.1 & 43.4 & 
41.6 & 42.5 & 42.7 & 
40.8 & 42.4 & 42.4 & 
33.7 & 35.6 & 35.5 & 
31.0 & 32.7 & 32.8 \\
Qwen2-VL 7B & \underline{71.7} & \underline{70.4} & 68.6 & 
\underline{65.5} & \underline{65.5} & 63.5 & 
\underline{64.4} & \underline{65.3} & \underline{64.1} & 
\underline{50.3} & \underline{52.5} & \underline{52.9} & 
\underline{43.5} & \underline{45.9} & \underline{46.6} \\
InternVL 2.5 4B & 47.7 & 44.5 & 43.0 & 
38.0 & 40.3 & 40.4 & 
36.7 & 39.6 & 40.3 & 
30.7 & 34.4 & 35.7 & 
28.8 & 32.1 & 33.7 \\
InternVL 2.5 8B & 66.2 & 69.0 & 68.7 & 
57.5 & 62.5 & 61.6 & 
52.9 & 58.5 & 58.1 & 
43.4 & 49.8 & 49.7 & 
39.7 & 45.9 & 46.2 \\
Centurio Qwen    & 35.3 & 36.1 & 35.6 & 
31.1 & 32.9 & 33.3 & 
31.0 & 32.8 & 33.1 & 
28.7 & 29.7 & 29.8 & 
28.1 & 28.7 & 28.7 \\
GPT-4o-mini & \textbf{77.5} & \textbf{86.4} & \textbf{89.1} & 
\textbf{77.2} & \textbf{86.5} & \textbf{88.6} & 
\textbf{77.1} & \textbf{86.1} & \textbf{88.4} & 
\textbf{68.4} & \textbf{79.8} & \textbf{82.7} & 
\textbf{61.7} & \textbf{74.0} & \textbf{77.2} \\
      \bottomrule
    \end{tabular}
  \end{adjustbox}
  \caption{\textbf{Cross-modal Topic Matching:} LVLMs must select the candidate sentence (image) from 4 choices that topically align with $k$ reference images (sentences). Prompts provided in \S\ref{appendix:prompts}. Languages are tiered by Wikipedia sizes (cf. \S\ref{sec:results}). Number of languages in parentheses. \textbf{Metric:} share of responses starting with correct option letter. Details in \S\ref{sec:experimental-setup}. In each column, the best model is emphasized in \textbf{bold}, the second-best model is \underline{underlined}.}
  \label{tab:main-results-image-text}
  \vspace{-0.2cm}
\end{table*}

\begin{table*}[ht]
  \centering
  \begin{adjustbox}{width=\textwidth,center}
    \setlength{\tabcolsep}{4pt}
    \begin{tabular}{
      @{\extracolsep{\fill}} 
      l  !{\vrule}
      c c c c  
      c c c c  
      c c c c  
      c c c c  
      c c c c  
    }
      \toprule
       \multicolumn{1}{c}{\textbf{Resourceness}} & \multicolumn{9}{r}{$\xleftarrow{\hspace{3cm}}$ \textbf{High} \rule[0.6ex]{4cm}{0.4pt}} & \multicolumn{2}{c}{\textbf{Mid}} & \multicolumn{9}{l}{\rule[0.6ex]{4cm}{0.4pt} \textbf{Low} $\xrightarrow{\hspace{3cm}}$} \\ \midrule
      \multicolumn{1}{c}{} &
      \multicolumn{4}{c}{\textbf{\textsc{English}}} &
      \multicolumn{4}{c}{\textbf{\textsc{Tier 4} (26)}} &
      \multicolumn{4}{c}{\textbf{\textsc{Tier 3} (32)}} &
      \multicolumn{4}{c}{\textbf{\textsc{Tier 2} (96)}} &
      \multicolumn{4}{c}{\textbf{\textsc{Tier 1} (51)}} \\
      \cmidrule(lr){2-5} \cmidrule(lr){6-9} \cmidrule(lr){9-13} \cmidrule(lr){14-17} \cmidrule(lr){18-21}
      \multirow{2}{*}{\textbf{Task}} &
      \textbf{Topic-To} &
      \multicolumn{3}{c}{\textbf{Sentences}} &
      \textbf{Topic-To} &
      \multicolumn{3}{c}{\textbf{Sentences}} &
      \textbf{Topic-To} &
      \multicolumn{3}{c}{\textbf{Sentences}} &
      \textbf{Topic-To} &
      \multicolumn{3}{c}{\textbf{Sentences}} &
      \textbf{Topic-To} &
      \multicolumn{3}{c}{\textbf{Sentences}} \\
        &
      \textbf{Sentence} &
      \multicolumn{3}{c}{\textbf{To-Topic}} &
      \textbf{Sentence} &
      \multicolumn{3}{c}{\textbf{      To-Topic}} &
      \textbf{Sentence} &
      \multicolumn{3}{c}{\textbf{      To-Topic}} &
      \textbf{Sentence} &
      \multicolumn{3}{c}{\textbf{      To-Topic}} &
      \textbf{Sentence} &
      \multicolumn{3}{c}{\textbf{      To-Topic}} \\
      
      \cmidrule(lr){2-2} \cmidrule(lr){3-5} 
      \cmidrule(lr){6-6} \cmidrule(lr){7-9} 
      \cmidrule(lr){10-10} \cmidrule(lr){11-13} 
      \cmidrule(lr){14-14} \cmidrule(lr){15-17}
      \cmidrule(lr){18-18} \cmidrule(lr){19-21}
      \textbf{\textit{k} References} &
      \textbf{1} & 
      \textbf{1} & 
      \textbf{3} &
      \textbf{5} & 
      \textbf{1} & 
      \textbf{1} & 
      \textbf{3} &
      \textbf{5} &
      \textbf{1} & 
      \textbf{1} & 
      \textbf{3} &
      \textbf{5} & 
      \textbf{1} &
      \textbf{1} &
      \textbf{3} & 
      \textbf{5} &
      \textbf{1} &
      \textbf{1} &
      \textbf{3} &
      \textbf{5} \\
      \midrule

    Qwen2-VL 2B &
                56.7 & 86.1 & 96.5 & 98.2 & 
                49.1 & 78.4 & 92.0 & 95.2 & 
                45.2 & 73.7 & 89.6 & 93.6 & 
                36.4 & 53.7 & 69.3 & 76.4 & 
                33.0 & 46.4 & 60.7 & 68.4 \\
Qwen2-VL 7B & 
            85.7 & 89.1 & 95.3 & 97.5 & 
            81.8 & 84.1 & 93.3 & 96.1 & 
            80.5 & 84.1 & 93.5 & 96.6 & 
            63.7 & 67.8 & 81.3 & 86.7 & 
            55.3 & 59.1 & 73.7 & 79.8 \\
InternVL 2.5 4B &
            81.4 & 90.6 & \underline{98.0} & \textbf{99.1} & 
            72.7 & 85.2 & 95.8 & 97.9 & 
            68.2 & 83.7 & 95.5 & 97.7 & 
            50.2 & 67.8 & 83.4 & 87.9 & 
            44.6 & 60.7 & 77.4 & 83.1 \\
InternVL 2.5 8B &
            \underline{87.0} & \underline{91.7} & \textbf{98.1} & \underline{99.0} & 
            83.0 & 86.3 & \underline{96.3} & \underline{98.3} & 
            79.1 & 82.4 & 94.1 & 96.7 & 
            65.4 & 66.8 & 81.8 & 86.9 & 
            58.1 & 58.3 & 74.6 & 80.9 \\
Centurio Qwen &
            85.4 & 89.9 & 96.7 & 97.7 & 
            \underline{83.6} & \underline{88.0} & 95.8 & 97.7 & 
            \underline{82.6} & \underline{87.7} & \underline{96.0} & \underline{97.9} & 
            \underline{70.4} & \underline{73.1} & \underline{86.8} & \underline{90.7} & 
            \underline{64.0} & \underline{66.1} & \underline{81.2} & \underline{85.7} \\
GPT-4o-mini &
            \textbf{88.5} & \textbf{92.4} & \textbf{98.1} & \textbf{99.1} & 
            \textbf{89.3} & \textbf{91.6} & \textbf{98.4} & \textbf{99.3} & 
            \textbf{89.3} & \textbf{91.6} & \textbf{98.4} & \textbf{99.3} & 
            \textbf{80.9} & \textbf{82.1} & \textbf{93.3} & \textbf{95.7} & 
            \textbf{73.9} & \textbf{74.3} & \textbf{88.2} & \textbf{92.1} \\
      \bottomrule
    \end{tabular}
  \end{adjustbox}
  \caption{\textbf{Text-only Topic Matching:} LVLMs must select the candidate sentence (topic) of 4 choices that aligns topically with the reference topics ($k$ reference sentences). See Table \ref{tab:main-results-image-text} for further details.}
  \label{tab:main-results-text}
  \vspace{-0.2cm}
\end{table*}

We categorize the languages in MVL-SIB based on their `resourceness'. To do so, we reorganize the language groups from \citet{joshi-etal-2020-state} into four tiers. We first rank the tiers w.r.t. Wikipedia size and then merge (i) the two highest-resource tiers and (ii) the two lowest-resource tiers.\footnote{Sorting by Wikipedia size (in number of pages) swaps tiers 1 and 2; we then merge Tier 0 with the new Tier 1, as well as tiers 4 and 5.} This both better reflects current corpus availability for LLM pre-training \citep{xue2021mt5,kudugunta2023madlad400},
and aligns with downstream performance (cf. Appendix \ref{app:tier-perf-plot}). We isolate English from Tier 4, since it is the pivotal language in NLP. The full per-language results by task and model are provided in Appendix \ref{app:per-language-results}.

\subsection{Cross-modal Topic Matching}

\rparagraph{Images-To-Sentence (\its)} The upper segment of Table \ref{tab:main-results-image-text} displays the results for \its, in which the LVLMs must pick the candidate sentence that topically matches the $k$ reference images. 

\iparagraph{English} The performance on \its with English candidate sentences scales well with model size. The small Qwen2-VL 2B performs only slightly better than chance (25\% vs. ca. 35\%).  The comparably sized Qwen2-VL 7B, InternVL 2.5 8B, and Centurio-Qwen 8B peak around 62\% to 68\% at various $k$. These models nevertheless non-negligibly trail GPT-4o-mini (78.1\%). Among LVLMs, only InternVL 2.5 8B, Centurio-Qwen, and GPT-4o-mini benefit from multiple reference images. When the number of references $k$ increases from 3 to 5, GPT-4o-mini declines slightly in performance, while InternVL and Centurio-Qwen continue to improve marginally (ca. +1\%) and more notably (ca. +3-6\%), respectively. All other LVLMs deteriorate materially with more reference images (ca. -3{-}4\%).
mSigLIP indeed is a strong baseline, trailing only GPT-4o-mini and InternVL 2.5 8B at $k{=}5$. The ViT yields large gains with 4 more images (+9.7\%).

\iparagraph{Tiers} The performance gap of other languages to English correlates well with their resource levels by language tier. When presented with candidate sentences in non-English high-resource languages (cf. Tier 4), GPT-4o-mini even performs better slightly better. For very low-resource languages in Tier 1, such as N'Koo or Tamazight, all models fail to perform better than chance (cf. Appendix \ref{app:images-to-sentences}). Among LVLMs, only GPT-4o-mini remains overall robust for topically matching sentences of low-resource languages to images, whereas other models drop severely in performance (ca. 15-20\%). While mSigLIP still performs well, it declines more notably than LVLMs on lower-resource languages.


\rparagraph{Sentences-To-Image (\sti)} The lower part of Table \ref{tab:main-results-image-text} presents the results for \sti. Here, the models select the candidate image among four options that topically fits the $k$ reference sentences.

\iparagraph{English} Performance again correlates well with model capacity. However, in \sti, only GPT-4o-mini significantly seizes on additional references (+13\%) to excel with 89\%, while models like Qwen2-VL 7B and InternVL 2.5 8B exhibit peak performance at $k{=}1$ and $k{=}3$, respectively, that taper slightly with more sentence references. Centurio-Qwen performs only slightly better than random (25\% vs. ca. 35\%). In \sti, mSigLIP is again very strong, second only to GPT-4o-mini at $k{=}5$. The encoder once more seizes sizable gains from additional references (+13.3\%).

\iparagraph{Tiers} In non-English evaluations, the overall trend remains similar, though absolute performance is lower. The gap between high- and low-resource language tiers is evident, as all models yield higher scores across Tiers 4 and 3. GPT-4o-mini maintains robust performance even in the most challenging Tier 1 when provided multiple references (ca.75\%).


\vspace{0.2cm}
\noindent \textbf{In sum}, both the training protocol and the model size collectively determine whether models favor \its or \sti. For instance, InternVL 2.5 8B outperforms Qwen2-VL 7B on \its across the board, while trailing on \sti.  Moreover, only GPT-4o-mini consistently seizes on additional references and remains largely robust to the lowest-resource languages on both tasks. This likely stems from insufficient training to enable open-weight LVLMs to perform higher-level VL reasoning with diverse texts and multiple images successfully. For instance, Centurio Qwen was mostly trained on data that prefixes images to text, resulting in low performance when images are postfixed to the context.

\subsection{Text-only Topic Matching}
\label{subsec:text-only-topic-matching}

Table \ref{tab:main-results-text} lists the results for text-only topic matching, in which the images are exchanged with their topic label. These tasks denominate `upper-bounds' for their cross-modal counterparts to enable ablations of language support and VL support in LVLMs.

\rparagraph{Topic-to-Sentence (\tts)}  In this task, LVLMs choose the sentence that best suits the reference topic. Barring Qwen2-VL 2B, all models perform well on the task. Notably, 5 images should capture the underlying topic well for all models (cf. Appendix \ref{app:images-to-topics}). Despite that, the gap between text-only and vision-language tasks (cf. Table \ref{tab:main-results-image-text}) is sizable across all open-weights models for `English' (ca. 20\%+). In English, GPT-4o-mini and InternVL 2.5 8B achieve the highest accuracies, indicating strong topic comprehension. For non-English languages, while the overall scores are reduced, high-resource languages benefit from richer training signals compared to their low-resource counterparts -- models like Qwen2-VL 2B and Centurio-Qwen 8B show a more pronounced drop in the latter, underscoring the impact of language resources.

\rparagraph{Sentences-to-Topic (\stt)} In \stt, where models choose the topic that best aligns with $k$ reference sentences, performance scales both with model size and the number of references. The gains from additional context ($3{-}5~k$) are particularly notable for larger LVLMs. In English, GPT-4o-mini improves markedly from 92.4\% at $k{=}1$ to 98.1\% at $k{=}3$. In non-English languages, similar patterns emerge: high-resource languages consistently yield higher accuracies than low-resource ones, with GPT-4o-mini and InternVL 2.5 8B exhibiting the most stable improvements across varying $k$. This reinforces the role of model capacity and training data diversity in effective cross-lingual  topic matching.


Comparing the results of cross-modal and text-only topic matching sheds further light on the VL interactions for lower-resource languages in LVLMs. The performance gap between matching sentences to reference images and matching them to textual topics tends to narrow regardless of the number of references, likely reflecting their limited textual support in LVLMs. In contrast, the discrepancy between selecting an image versus a topic for reference sentences becomes much more pronounced, especially at $k{=}5$. These findings suggest that VL support degrades more sharply than textual support for lower-resource languages in LVLMs.

\begin{figure*}[htb!]
  \centering
  \begin{adjustbox}{width=\textwidth,center}
    \begin{tabular}{cc}
      \begin{minipage}[t]{0.48\textwidth}
        \centering
        \includegraphics[width=\textwidth,trim={0 0 0 0.6cm},clip]{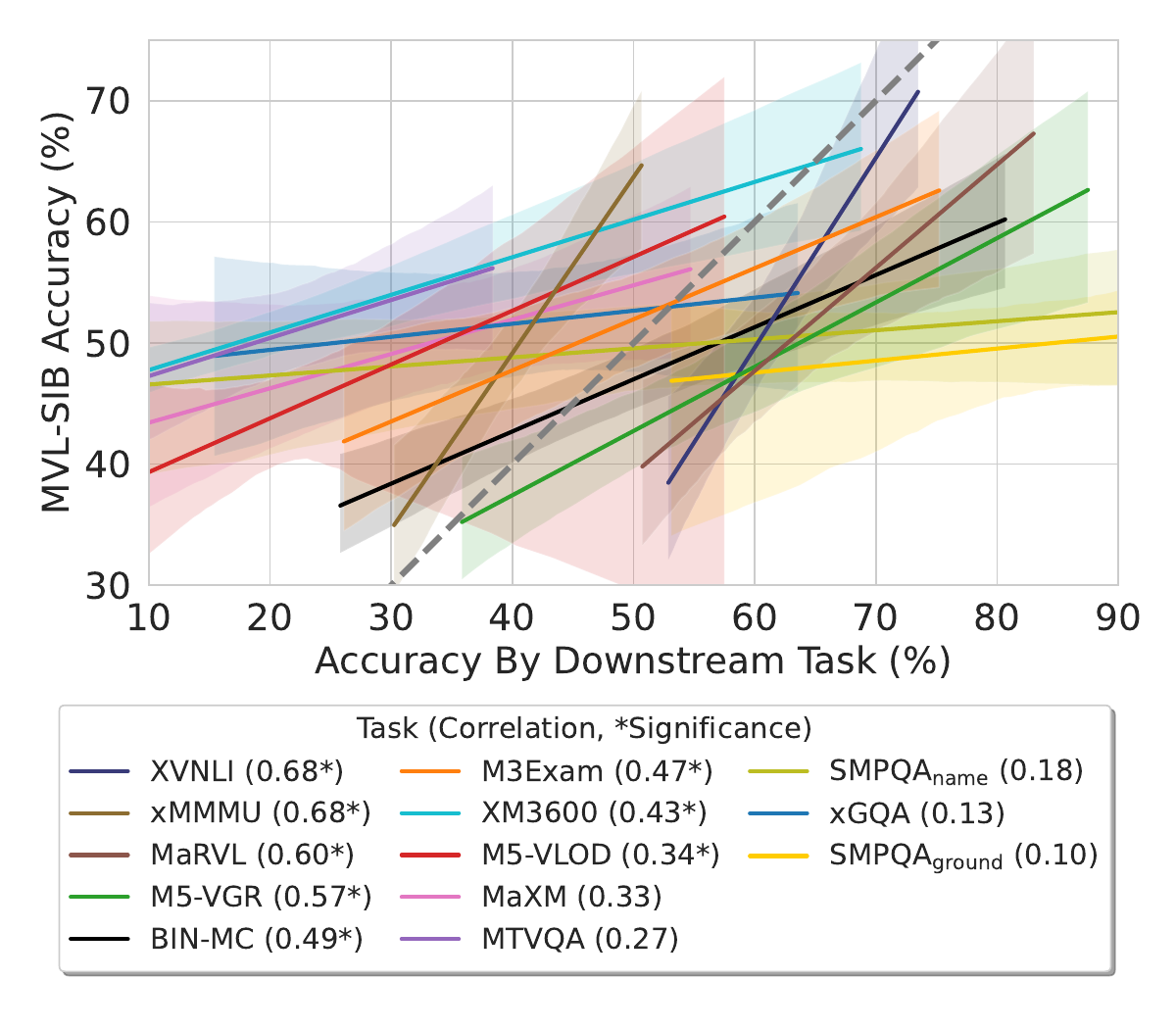}
        \vspace{-1cm}
        \captionof{figure}{\its with $k{=}3$.}
        \label{fig:img2sent}
      \end{minipage} &
      \hspace{1cm}
      \begin{minipage}[t]{0.48\textwidth}
        \centering
        \includegraphics[width=\textwidth,trim={0 0 0 0.6cm},clip]{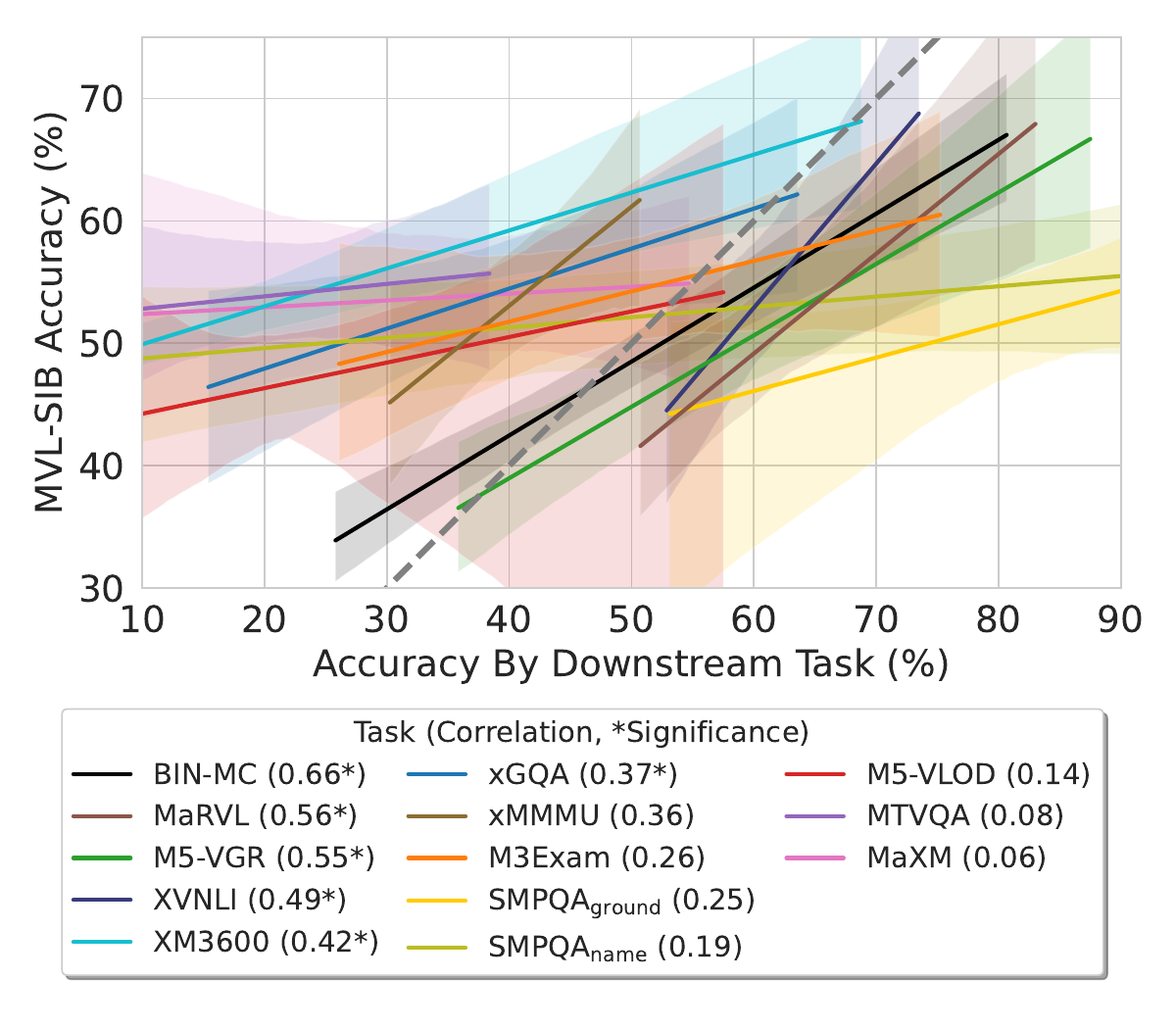}
        \vspace{-1cm}
        \captionof{figure}{\sti with $k{=}3$.}
        \label{fig:sent2img}
      \end{minipage} \\
    \end{tabular}
  \end{adjustbox}
   \vspace{-0.1cm}
   \caption*{\textbf{Correlations Between MVL-SIB \& Multilingual VL Benchmarks.} Pearson correlation coefficients obtained by regressing MVL-SIB performance against performance on multilingual VL tasks on languages common to both datasets, respectively. An asterisk (*) indicates whether the coefficient is statistically significant at $p \leq 0.05$.}
   \vspace{-0.4cm}
\end{figure*}

\subsection{Further Analyses}
\label{subsec:further-analyses}

\rparagraph{Task Correlation} To compare MVL-SIB with other tasks, we access the results for Qwen2-VL and InternVL 2.5 models on several established multilingual VL benchmarks from \citet{geigle2025centurio}.\footnote{We omit Centurio-Qwen, since it degrades on the \sti task. We further provide details on all the multilingual vision-language benchmarks we correlate MVL-SIB with in Appendix \ref{app:summary-mvl-benchmarks}}
Next, we align the results across languages for the MVL-SIB tasks and the other benchmarks. Finally, we plot the linear regressions of performance on \its and \sti with $k{=}3$ against the performance on the VL benchmarks, respectively, pooled over models, in Figures \ref{fig:img2sent} and \ref{fig:sent2img}.\footnote{Note that we include a constant in our regression model to bridge task-specific scales of results.}

MVL-SIB positively correlates with all tasks in both the \its and \sti evaluations. However, both the magnitude and statistical significance of these linear relationships vary across VL benchmarks. Both \its and \sti exhibit the strongest connections with XVNLI (visual inference),  BIN-MC (multiple-choice image classification), MaRVL, and M5-VGR (both visually-grounded boolean reasoning). Since these tasks restrict valid answers to a small set of fixed options (i.e., choice letters or `yes/no/maybe'), LVLMs must engage in higher‐level vision-language disambiguation rather than relying solely on lower‐level visual cues to solve these tasks.
In addition, xMMMU, M3Exam, and M5-VLOD are significantly related only to \its, whereas xGQA is significantly aligned solely with \sti. The former tasks are structurally similar to \its, typically presenting one or more images that LVLMs are given as visual context to answer multiple-choice questions.
We hypothesize that only \sti is significantly correlated with xGQA, since \sti is more analogous to object-centric benchmarks. In \sti, LVLMs likely leverage targeted semantic cues (e.g., keywords or phrases) from the reference sentences to better disambiguate candidate images by topic. This behavior aligns with lower-level VL tasks such as xGQA (cross-lingual short-form QA) or BIN-MC (multiple-choice object classification), where texts and images are more deliberately connected. 
In contrast, MTVQA, SMPQA‐name, and SMPQA‐ground show only weak or statistically insignificant correlations with the MVL‐SIB tasks. Since these tasks require LVLMs to comprehend text embedded in images, a low‐level, fine‐grained VL task, they differ substantially from the higher‐level VL reasoning evaluated by MVL‐SIB.

Overall, the regression analysis indicates that \its and \sti capture distinct yet complementary aspects of VL understanding, collectively exhibiting a strong relationship with a broad set of VL tasks. This aligns with our main results (cf. Table \ref{tab:main-results-image-text}), which show that different LVLMs may favor one MVL-SIB task over the other. This renders MVL-SIB as a suitable benchmark for evaluating \textit{universal} VL understanding (across 205 languages). It also enables to ablate performance across the key axes of analysis, the task formulation (\its vs. \sti), the language vs. vision-language support for 205 languages, and the number of images in context.\footnote{In \sti, candidates could comprise more than a single image.}

\begin{figure*}[ht]
  \begin{adjustbox}{width=\textwidth,center}
        \includegraphics[width=\textwidth, trim={0 1.65cm 0 0.2cm},clip]{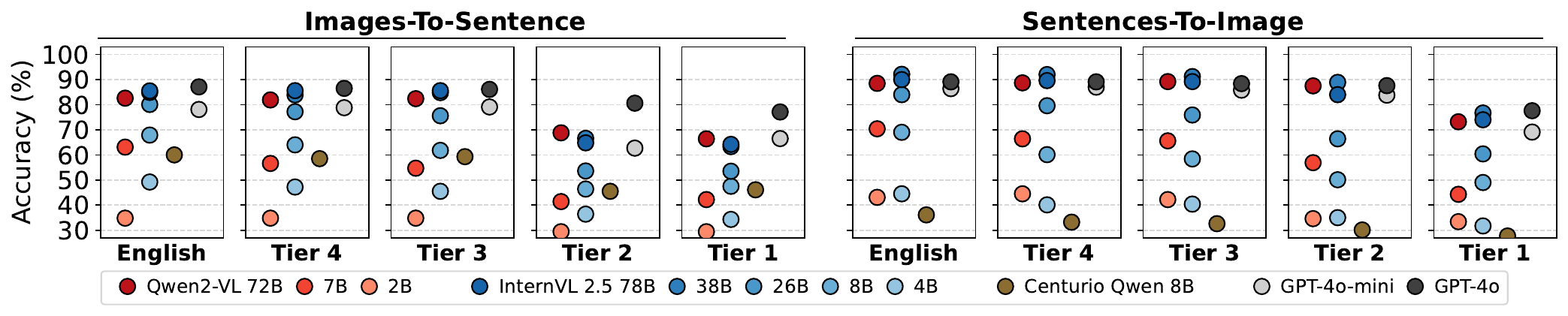}
  \end{adjustbox}
  \begin{adjustbox}{width=\textwidth,center}
        \includegraphics[width=\textwidth, trim={0 0 0 0.2cm},clip]{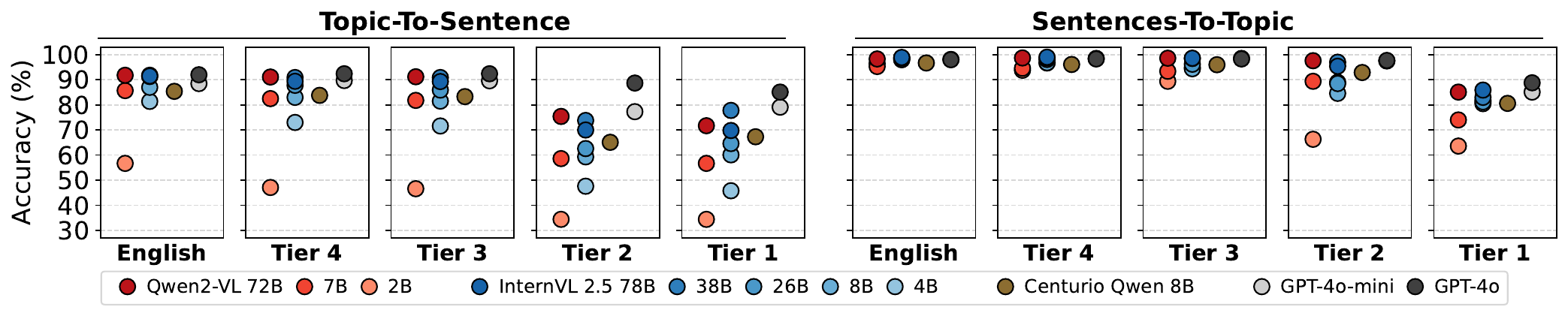}
  \end{adjustbox}
  \vspace{-0.75cm}
  \caption{\textbf{Larger LVLMs on Subsampled Tiers.} We extract 3 languages per tier that mimic avg. performance full language groups (cf. \S\ref{subsec:further-analyses}) and evaluate LVLMs across all model sizes on \{\its,\sti,\tts,\stt\} @~$k{=}3$ (cf. \S\ref{sec:tasks}).}
  \label{fig:perf-subgroups}
  \vspace{-0.4cm}
\end{figure*}

\rparagraph{Larger LVLMs} To evaluate larger LVLMs on MVL-SIB, we construct language-tier subsets that reliably estimate performance while mitigating excessive computational overhead (cf. \S\ref{sec:results}). For both \its and \sti, we identify the three languages in each tier that best replicate the average performance of the tier. First, we compute the average performance per language tier for both \its and \sti with $k{=}3$, pooling results across models (cf. \S\ref{sec:results}). Then, for each tier, we select the three languages whose performance deviates least from the tier mean. The languages chosen for each tier by task are detailed in Appendix \ref{app:subsets-full-results}. Finally, we test GPT-4o, InternVL 2.5 \{26,32,72\}B, and Qwen 2.5 VL \{32,70\}B on these subsets to assess how well larger models perform across language tiers.

Figure \ref{fig:perf-subgroups} displays the results for both \its and \sti with $k{=}3$.\footnote{Qwen2-VL 72B frequently responded with the first letter of the correct topic. If that letter uniquely identifies the correct choice, the answer is considered correct.} We observe that larger models catch up to GPT-4o on \its and even outperform it on \sti for higher-resource languages. Since open-weight LVLMs have more limited language support for low-resource languages, GPT-4 and GPT-4o-mini outperform all other models on \its for languages in tier 1 and 2. Increasing model capacity yields the largest gains on \sti, for which the models exceed GPT-4o up to the lowest-resource Tier 1. Moreover, unlike smaller models, all larger LVLMs (+26B) more effectively leverage multiple image references to improve performance (cf. Appendix \ref{app:subsets-full-results}). They reap the largest benefit when the number of references increases from 1 to 3 (ca. +3\%), with further improvements at $k=5$ (ca. +1.5\%). These results suggest that larger LVLMs have fundamentally better VL support irrespective of the evaluated language (cf. \S\ref{subsec:text-only-topic-matching}). The results on text-only topic matching further underscore this notion (cf. lower segment of Figure \ref{fig:perf-subgroups}). Larger LVLMs near perfectly match both the reference topic to the correct sentence (ca. 90\%) and references sentences to the right topic (ca. 98\%). Notably, as model size increases, the performance gap between cross-modal and text-only matching narrows.  Collectively, these results further indicate that VL support relative to text-only support improves with increasing model capacity in LVLMs.

\section{Conclusion}
\label{sec:conclusion}

We present the massively multilingual vision-language benchmark MVL-SIB for cross-modal (and text-only topic matching) in 205 languages that offers key advantages over prior multilingual VL benchmarks. Notably, it covers over 100 additional languages without relying on machine translation. MVL-SIB allows for a clear separation between textual language support and vision-language support in LVLMs by comparing performance on mirrored cross-modal and text-only tasks. Moreover, it allows us to study how LVLMs handle single-image versus multi-image formulations of cross-modal topic matching by varying the number of images provided.
In our comparative evaluation of state-of-the-art LVLMs on MVL-SIB, we find that model performance is strongly correlated with both model size and the volume of available pre-training data for each language. However, all LVLMs experience a dramatic performance drop on the lowest-resource languages. Our analysis further reveals that vision-language support deteriorates disproportionately relative to language support, highlighting the need to incorporate low-resource languages into VL training. Moreover, providing multiple images does not benefit open-weight LVLMs in cross-modal topic matching, suggesting that LVLMs are not yet fully effective in multi-image tasks. Lastly, we validate that MVL-SIB correlates well with existing multilingual VL benchmarks, underscoring its reliability as a source of evaluation data for 205 languages.
\section{Limitations}
\label{sec:limitations}
%

Our work faces three primary limitations. First, although a vast number of LVLMs exist, we selected a representative subset based on key criteria. Specifically, the LVLMs in our study (Qwen2-VL and InternVL, with the exception of Centurio) span a range of parameter counts typical of LLMs. Additionally, we include GPT-4o-mini in the full evaluation and GPT-4o on the subsampled language tiers. Evaluating MVL-SIB across all four tasks \its, \sti, \tts, and \stt (cf. \S\ref{sec:tasks}) at various $k \in \{1,3,5\}$   over 205 languages (i.e., evaluations per model and task, or $2050$, in sum per model) becomes computationally intractable.  This accumulates to $3 \times 205 = 615$ evaluations per model ($205$ for \tts as only $k{=}1$ reference topic exists) or $3 \times 205 + 205 = 2050$ evaluations in total. We therefore both provide subsets of the language tiers to evaluate on and demonstrate that evaluation only requires 1K instances to reliably estimate task performance.
Second, while we strove to choose a diverse set of images to capture the full semantic range of each topic, further diversification is possible by sourcing additional images. However, due to the limited availability of openly licensed images, some topics (e.g., \texttt{politics} and \texttt{entertainment}) are represented predominantly by images that embody the topic in a more Western-centric cultural context. Hand-selecting images by topic for each language or, more broadly, cultural groups would not scale to 205 languages and would hinder the comparability of results. Our results furthermore confirm that models just as well perform on a broad range of languages spanning diverse cultural backgrounds as on English (cf. Figure \ref{fig:intro-img2sent}). At the same time, LVLMs perform best on Western-centric images, mitigating any variation that would originate from using more culture-specific images.
Finally, for the topic \texttt{geography}, we manually selected images that are representative within the context of SIB, as the broader definition of geography is too diffuse to capture visually.
\section*{Acknowledgements}
\label{sec:ack}

We used AI assistance (chatGPT o3-mini) to polish the writing and the tables of the manuscript as well as to refine the code for our visualizations.
%

\bibliography{custom}

\appendix

\section{Appendix}
\label{sec:appendix}

\begin{table*}[ht]
    \subsection{Images Per Topic}
    \label{app:images-per-topic}
    \centering
    \begin{tabular}{c}
       \toprule 
            \textbf{Entertainment}  \\ \cmidrule(lr){1-1}
                    \begin{minipage}{0.08\textwidth}
                        \centering
                        \includegraphics[width=\linewidth]{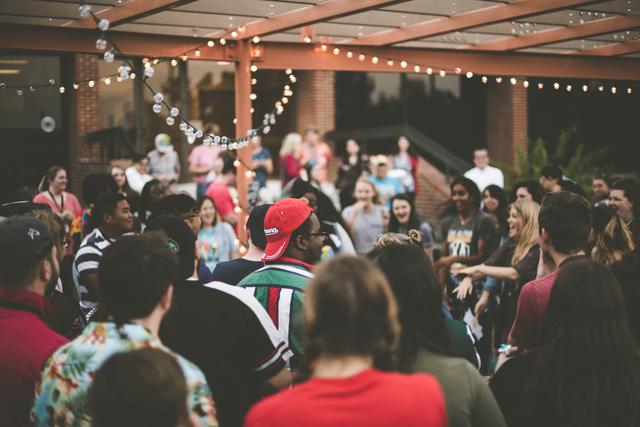}
                    \end{minipage}
                    \hfill
                    \begin{minipage}{0.08\textwidth}
                        \centering
                        \includegraphics[width=\linewidth]{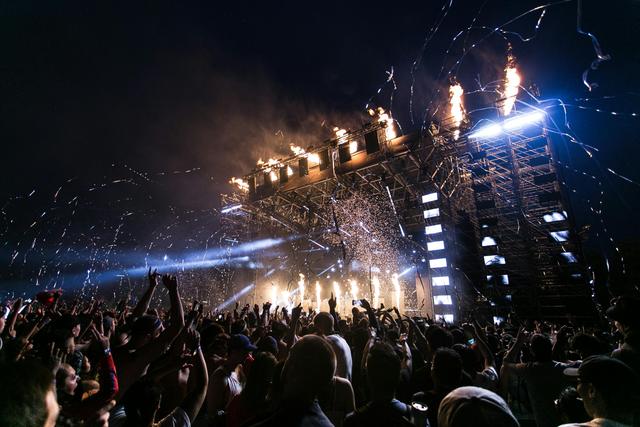}
                    \end{minipage}
                    \hfill
                    \begin{minipage}{0.08\textwidth}
                        \centering
                        \includegraphics[width=\linewidth]{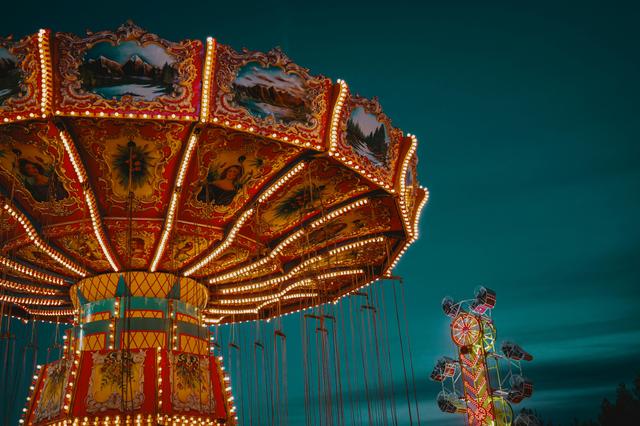}
                    \end{minipage}
                    \hfill
                    \begin{minipage}{0.08\textwidth}
                        \centering
                        \includegraphics[width=\linewidth]{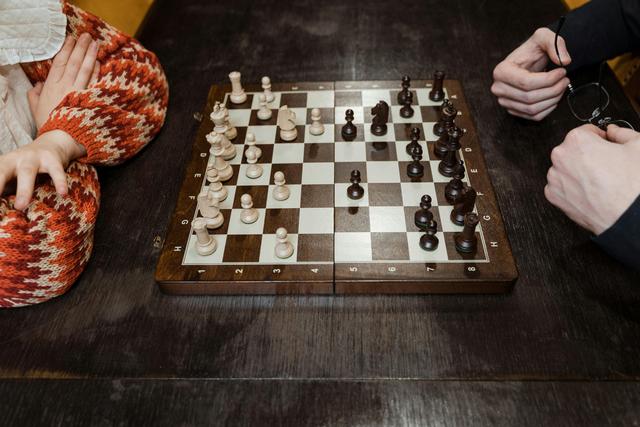}
                    \end{minipage}
                    \hfill
                    \begin{minipage}{0.08\textwidth}
                        \centering
                        \includegraphics[width=\linewidth]{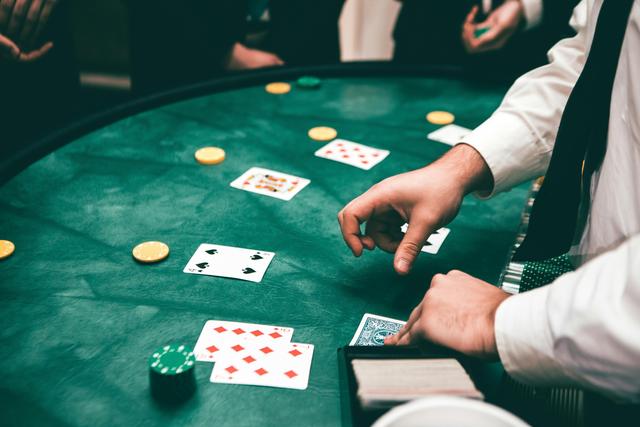}
                    \end{minipage}
                    \hfill
                    \begin{minipage}{0.08\textwidth}
                        \centering
                        \includegraphics[width=\linewidth]{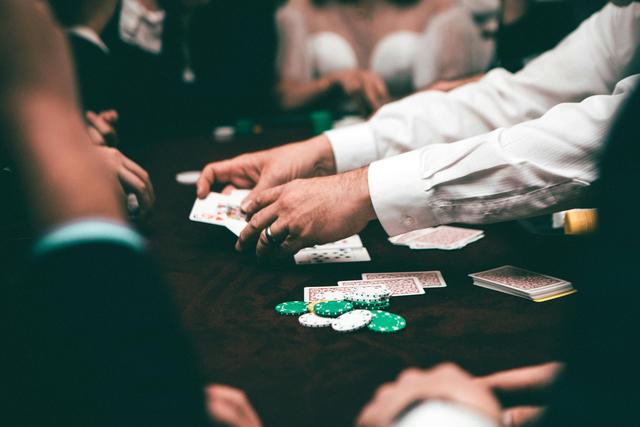}
                    \end{minipage}
                    \hfill
                    \begin{minipage}{0.08\textwidth}
                        \centering
                        \includegraphics[width=\linewidth]{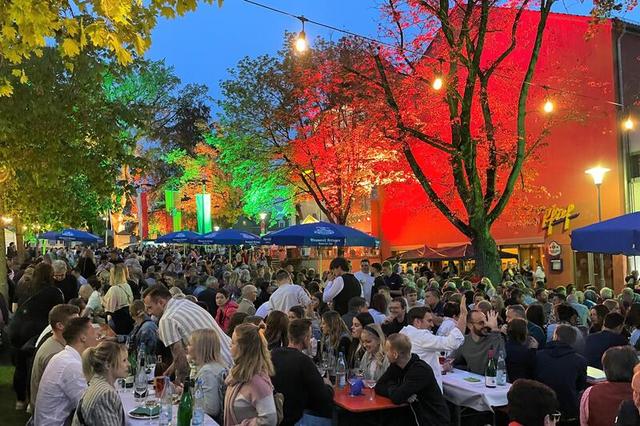}
                    \end{minipage}
                    \hfill
                    \begin{minipage}{0.08\textwidth}
                        \centering
                        \includegraphics[width=\linewidth]{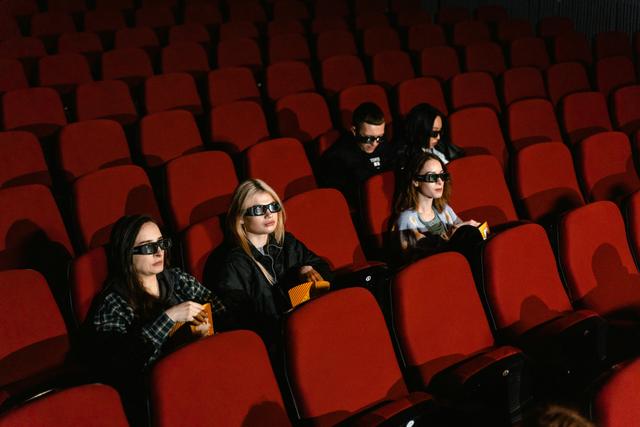}
                    \end{minipage}
                    \begin{minipage}{0.08\textwidth}
                        \centering
                        \includegraphics[width=\linewidth]{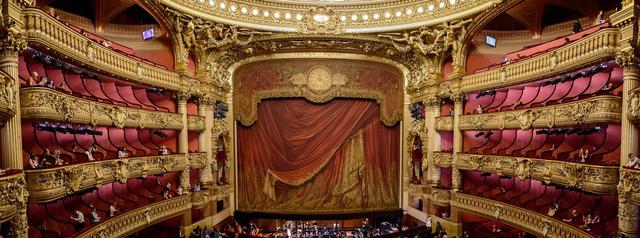}
                    \end{minipage}
                    \begin{minipage}{0.08\textwidth}
                        \centering
                        \includegraphics[width=\linewidth]{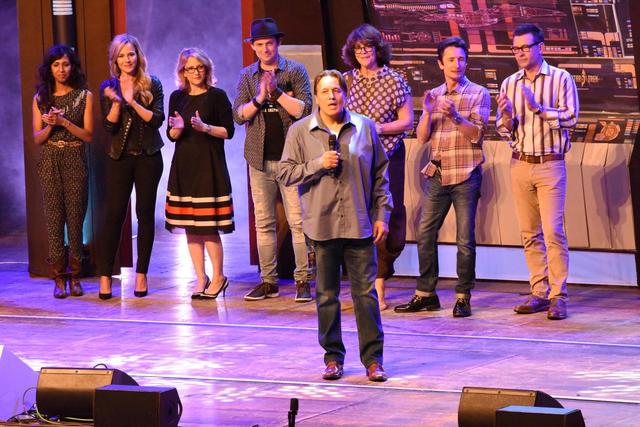}
                    \end{minipage}
                \\
    
            \textbf{Geography}  \\ \cmidrule(lr){1-1}
                    \begin{minipage}{0.08\textwidth}
                        \centering
                        \includegraphics[width=\linewidth]{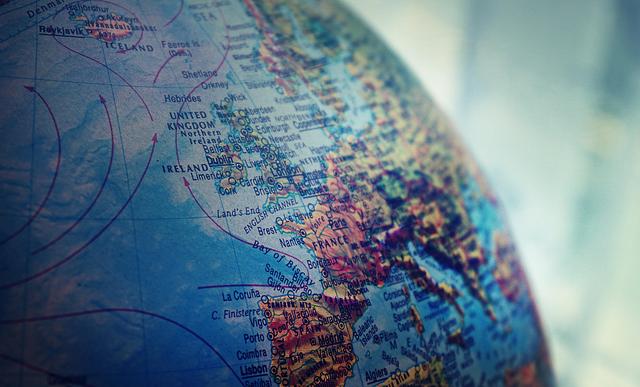}
                    \end{minipage}
                    \hfill
                    \begin{minipage}{0.08\textwidth}
                        \centering
                        \includegraphics[width=\linewidth]{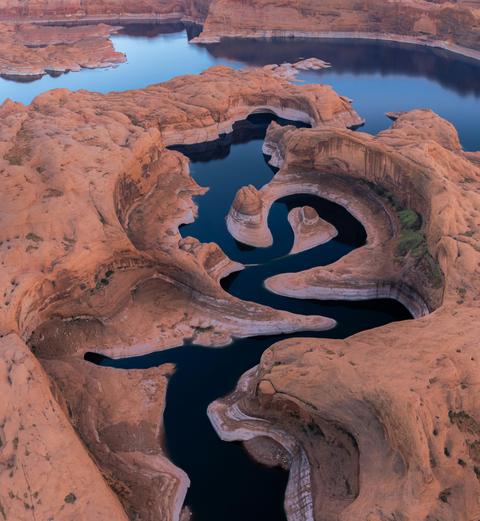}
                    \end{minipage}
                    \hfill
                    \begin{minipage}{0.08\textwidth}
                        \centering
                        \includegraphics[width=\linewidth]{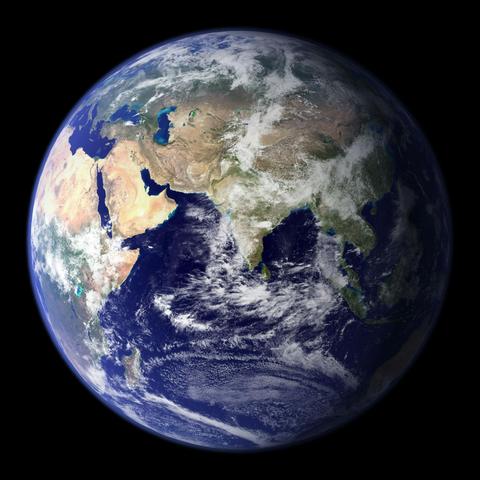}
                    \end{minipage}
                    \hfill
                    \begin{minipage}{0.08\textwidth}
                        \centering
                        \includegraphics[width=\linewidth]{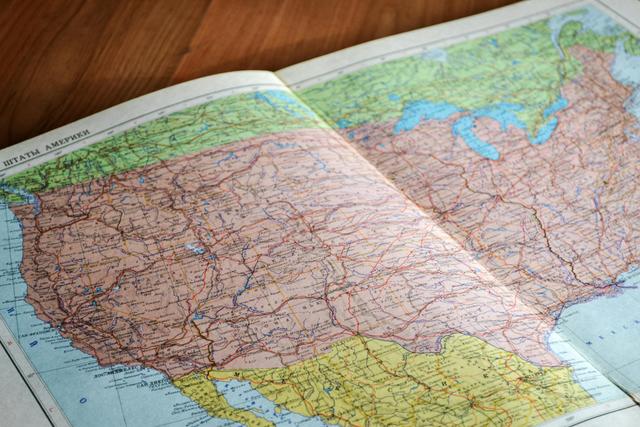}
                    \end{minipage}
                    \hfill
                    \begin{minipage}{0.08\textwidth}
                        \centering
                        \includegraphics[width=\linewidth]{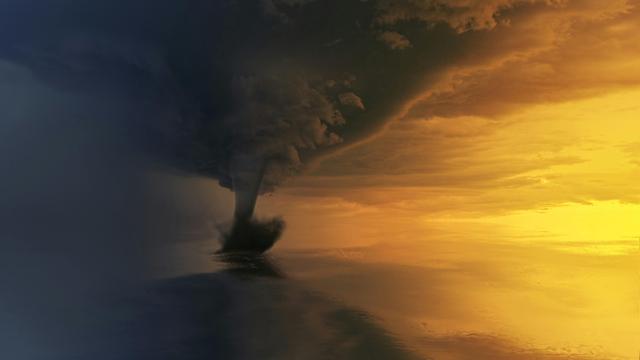}
                    \end{minipage}
                    \hfill
                    \begin{minipage}{0.08\textwidth}
                        \centering
                        \includegraphics[width=\linewidth]{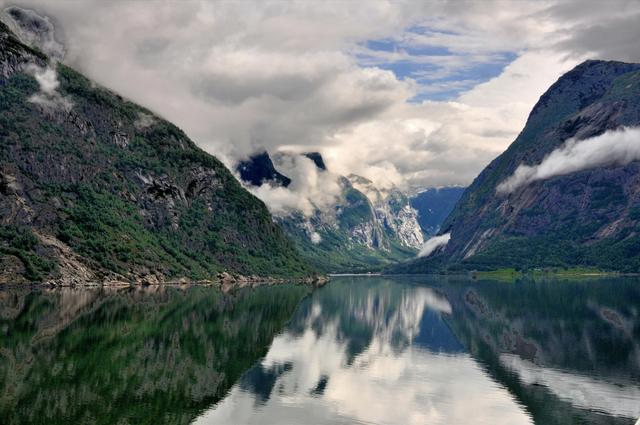}
                    \end{minipage}
                    \hfill
                    \begin{minipage}{0.08\textwidth}
                        \centering
                        \includegraphics[width=\linewidth]{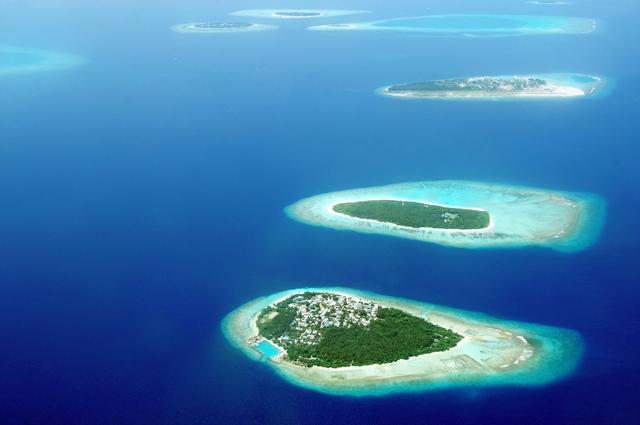}
                    \end{minipage}
                    \hfill
                    \begin{minipage}{0.08\textwidth}
                        \centering
                        \includegraphics[width=\linewidth]{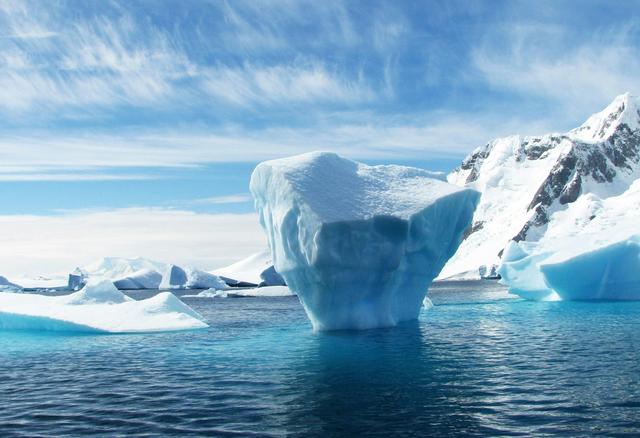}
                    \end{minipage}
                    \begin{minipage}{0.08\textwidth}
                        \centering
                        \includegraphics[width=\linewidth]{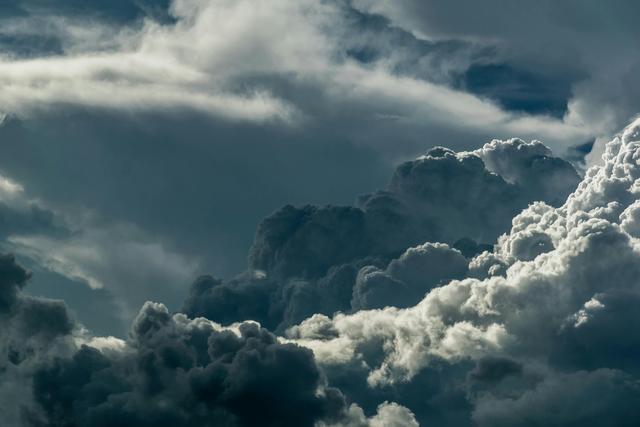}
                    \end{minipage}
                    \begin{minipage}{0.08\textwidth}
                        \centering
                        \includegraphics[width=\linewidth]{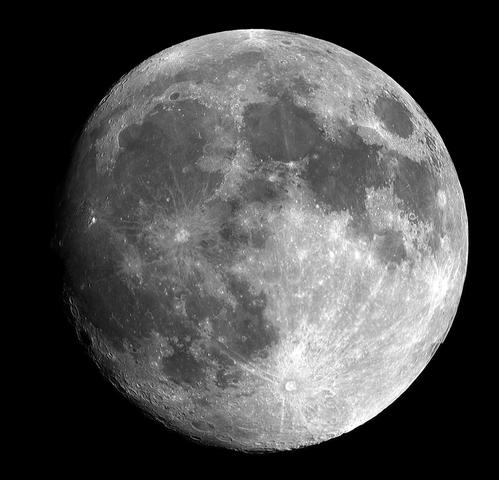}
                    \end{minipage}
                \\
    
            \textbf{Health}  \\ \cmidrule(lr){1-1}
                    \begin{minipage}{0.08\textwidth}
                        \centering
                        \includegraphics[width=\linewidth]{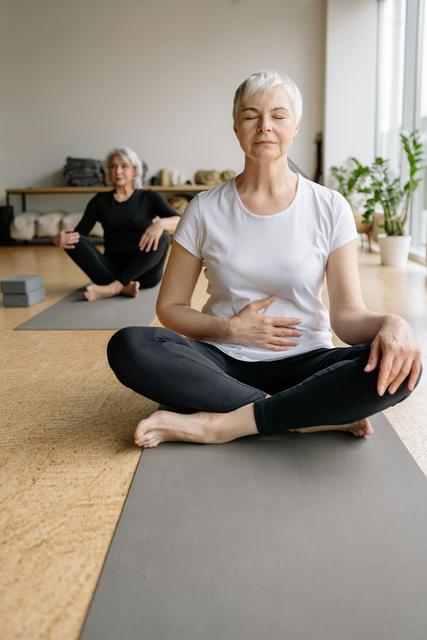}
                    \end{minipage}
                    \hfill
                    \begin{minipage}{0.08\textwidth}
                        \centering
                        \includegraphics[width=\linewidth]{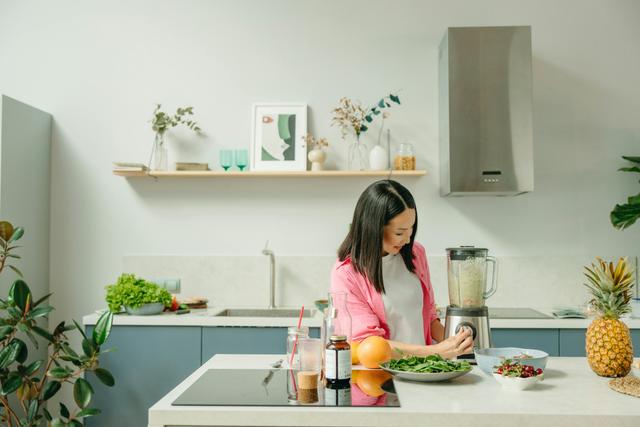}
                    \end{minipage}
                    \hfill
                    \begin{minipage}{0.08\textwidth}
                        \centering
                        \includegraphics[width=\linewidth]{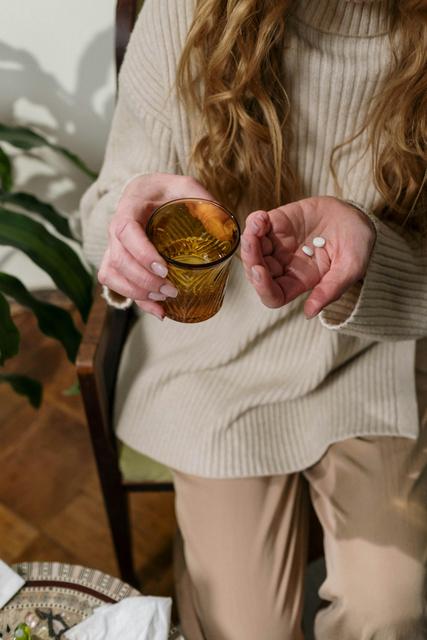}
                    \end{minipage}
                    \hfill
                    \begin{minipage}{0.08\textwidth}
                        \centering
                        \includegraphics[width=\linewidth]{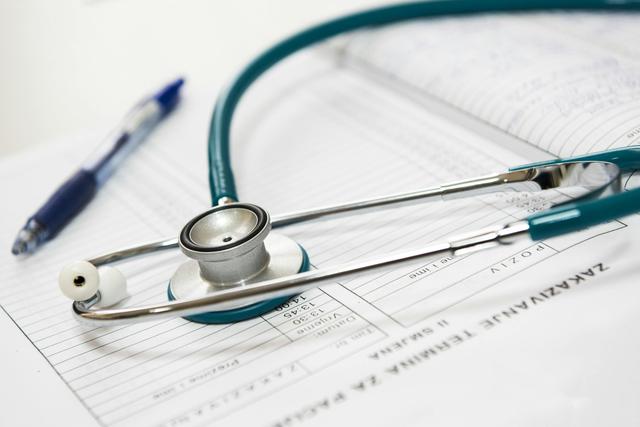}
                    \end{minipage}
                    \hfill
                    \begin{minipage}{0.08\textwidth}
                        \centering
                        \includegraphics[width=\linewidth]{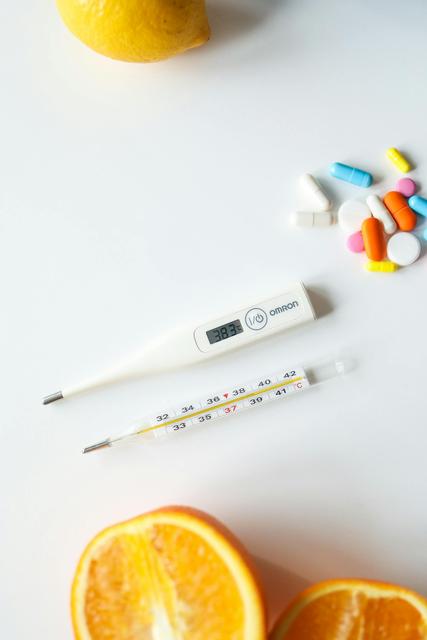}
                    \end{minipage}
                    \hfill
                    \begin{minipage}{0.08\textwidth}
                        \centering
                        \includegraphics[width=\linewidth]{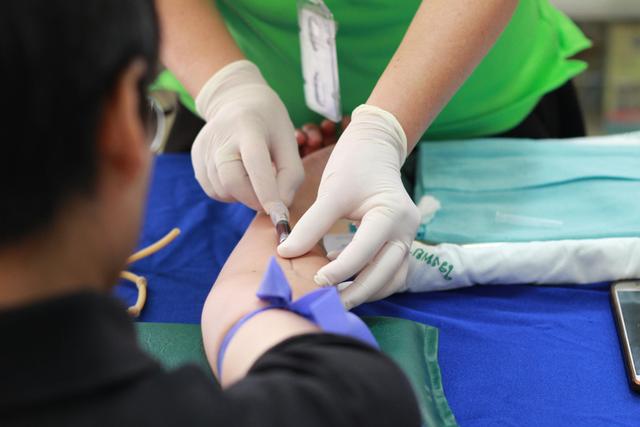}
                    \end{minipage}
                    \hfill
                    \begin{minipage}{0.08\textwidth}
                        \centering
                        \includegraphics[width=\linewidth]{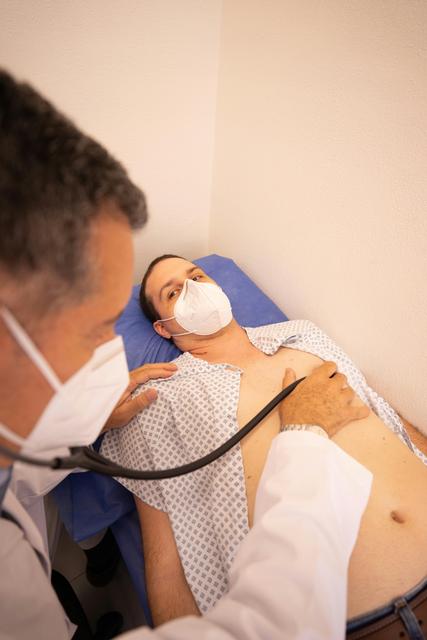}
                    \end{minipage}
                    \hfill
                    \begin{minipage}{0.08\textwidth}
                        \centering
                        \includegraphics[width=\linewidth]{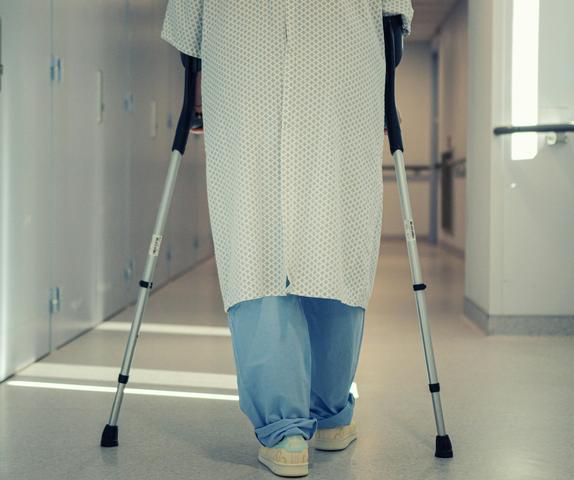}
                    \end{minipage}
                    \begin{minipage}{0.08\textwidth}
                        \centering
                        \includegraphics[width=\linewidth]{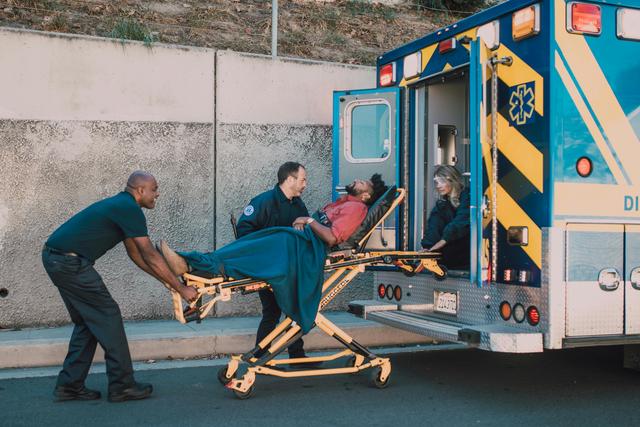}
                    \end{minipage}
                    \begin{minipage}{0.08\textwidth}
                        \centering
                        \includegraphics[width=\linewidth]{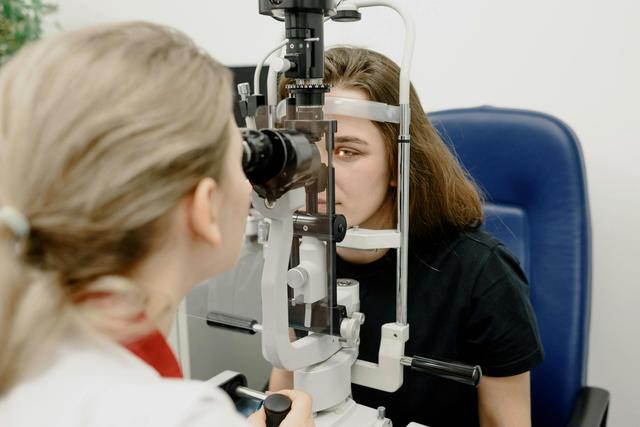}
                    \end{minipage}
                \\
                \textbf{Politics}  \\ \cmidrule(lr){1-1}
                    \begin{minipage}{0.08\textwidth}
                        \centering
                        \includegraphics[width=\linewidth]{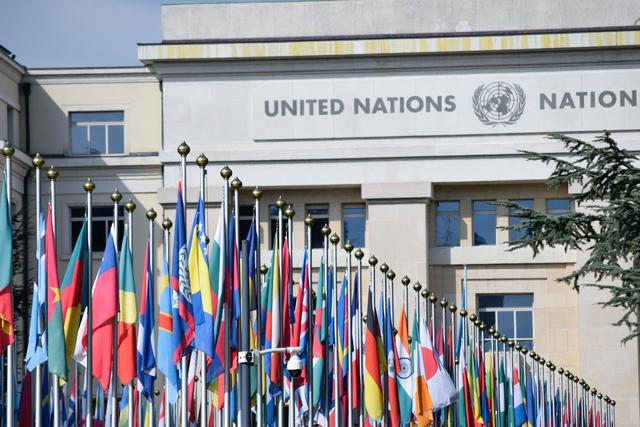}
                    \end{minipage}
                    \hfill
                    \begin{minipage}{0.08\textwidth}
                        \centering
                        \includegraphics[width=\linewidth]{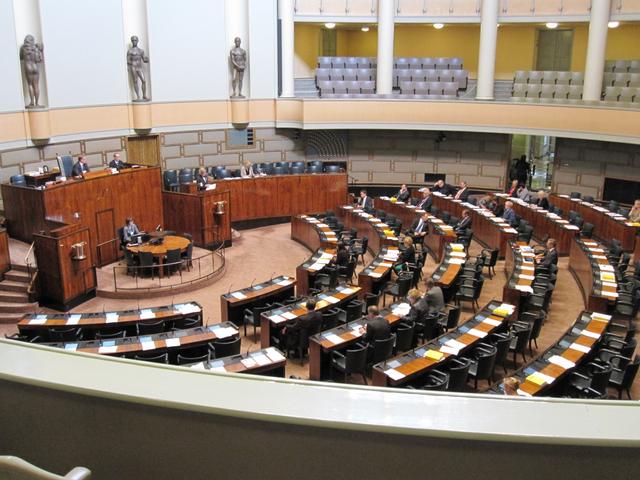}
                    \end{minipage}
                    \hfill
                    \begin{minipage}{0.08\textwidth}
                        \centering
                        \includegraphics[width=\linewidth]{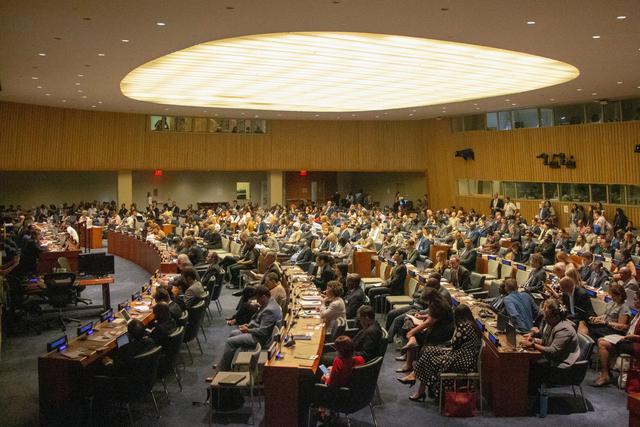}
                    \end{minipage}
                    \hfill
                    \begin{minipage}{0.08\textwidth}
                        \centering
                        \includegraphics[width=\linewidth]{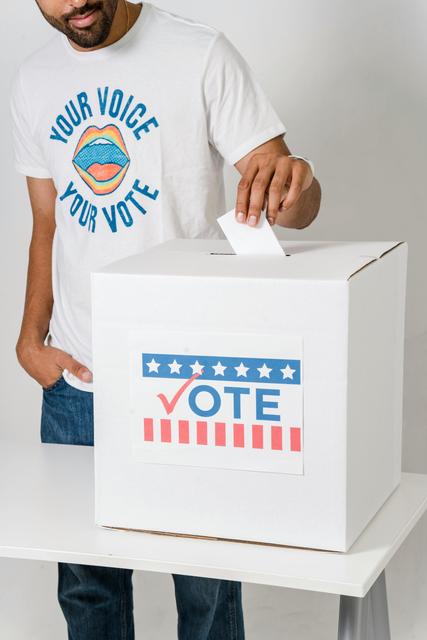}
                    \end{minipage}
                    \hfill
                    \begin{minipage}{0.08\textwidth}
                        \centering
                        \includegraphics[width=\linewidth]{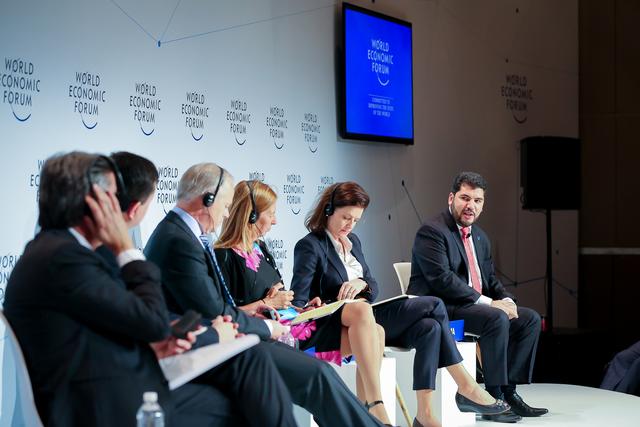}
                    \end{minipage}
                    \hfill
                    \begin{minipage}{0.08\textwidth}
                        \centering
                        \includegraphics[width=\linewidth]{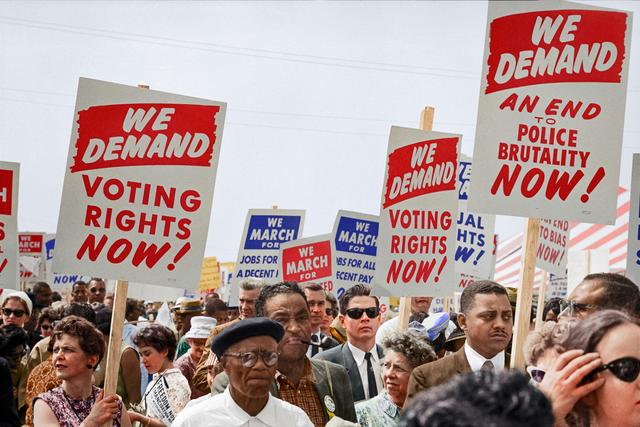}
                    \end{minipage}
                    \hfill
                    \begin{minipage}{0.08\textwidth}
                        \centering
                        \includegraphics[width=\linewidth]{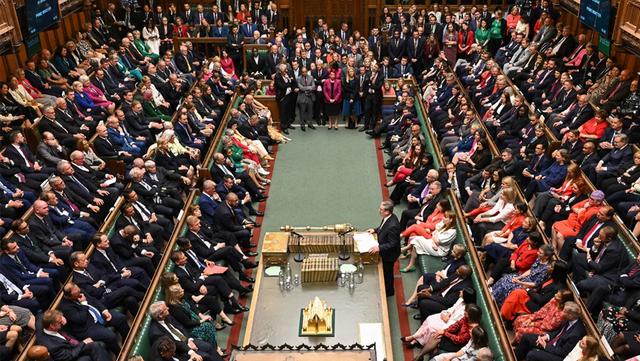}
                    \end{minipage}
                    \hfill
                    \begin{minipage}{0.08\textwidth}
                        \centering
                        \includegraphics[width=\linewidth]{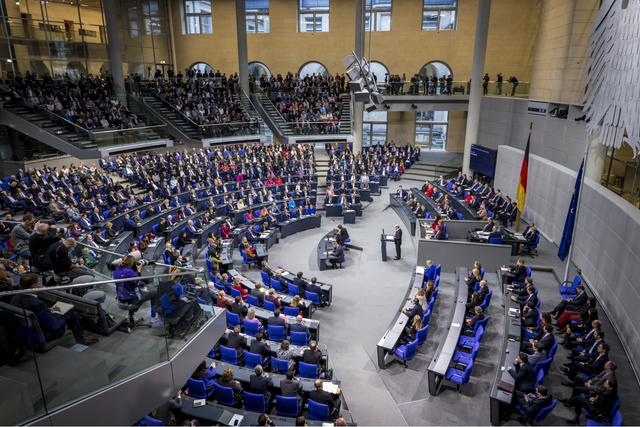}
                    \end{minipage}
                    \begin{minipage}{0.08\textwidth}
                        \centering
                        \includegraphics[width=\linewidth]{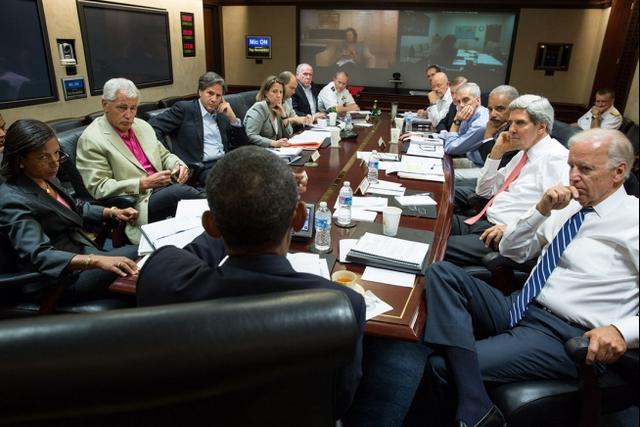}
                    \end{minipage}
                    \begin{minipage}{0.08\textwidth}
                        \centering
                        \includegraphics[width=\linewidth]{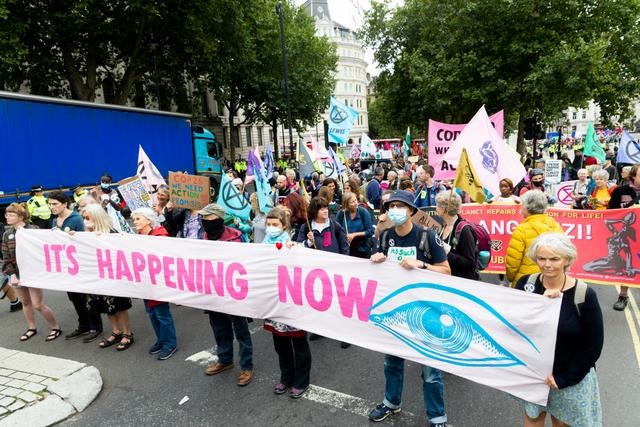}
                    \end{minipage}
                \\

            \textbf{Science \& Technology}  \\ \cmidrule(lr){1-1}
                    \begin{minipage}{0.08\textwidth}
                        \centering
                        \includegraphics[width=\linewidth]{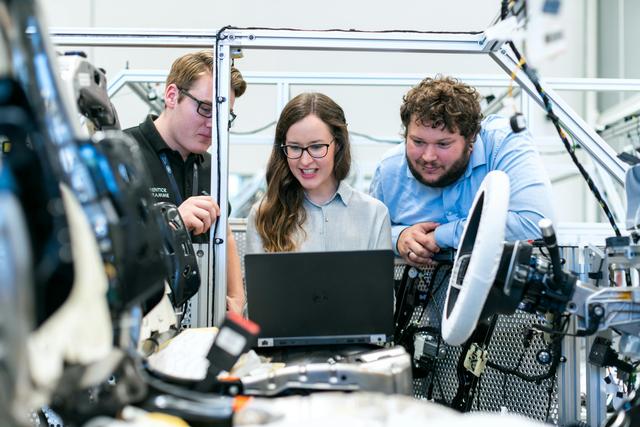}
                    \end{minipage}
                    \hfill
                    \begin{minipage}{0.08\textwidth}
                        \centering
                        \includegraphics[width=\linewidth]{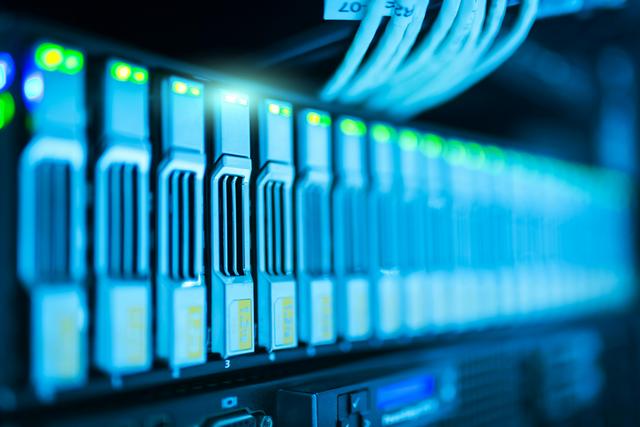}
                    \end{minipage}
                    \hfill
                    \begin{minipage}{0.08\textwidth}
                        \centering
                        \includegraphics[width=\linewidth]{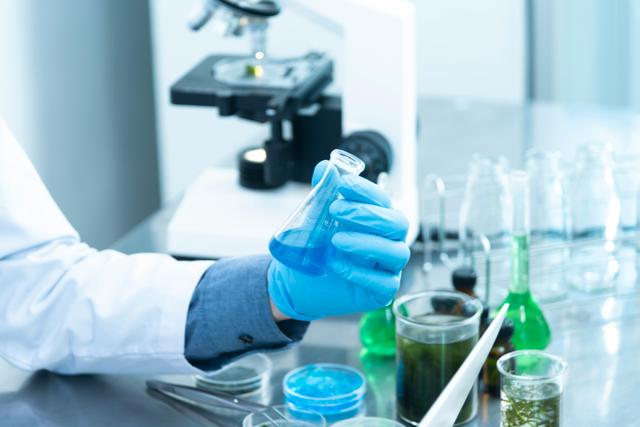}
                    \end{minipage}
                    \hfill
                    \begin{minipage}{0.08\textwidth}
                        \centering
                        \includegraphics[width=\linewidth]{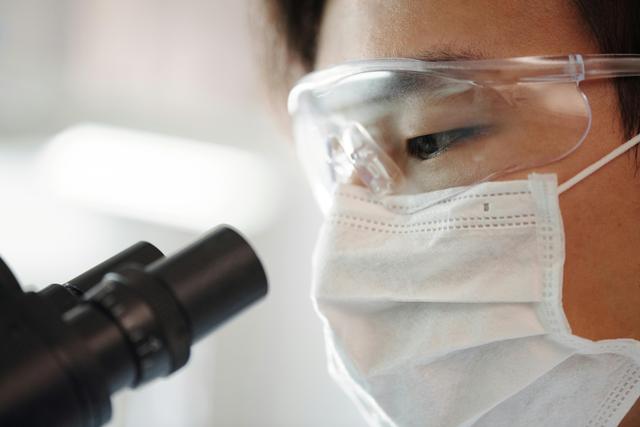}
                    \end{minipage}
                    \hfill
                    \begin{minipage}{0.08\textwidth}
                        \centering
                        \includegraphics[width=\linewidth]{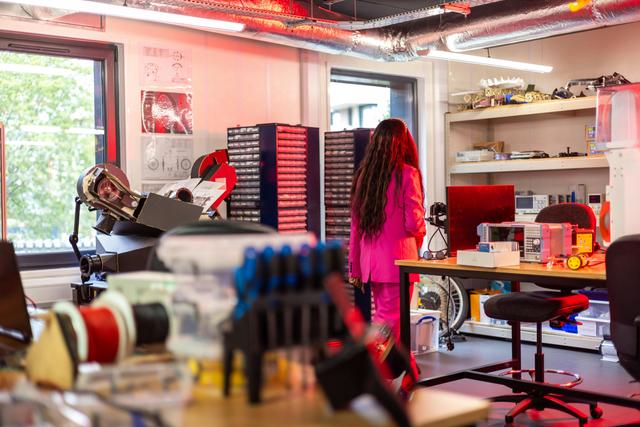}
                    \end{minipage}
                    \hfill
                    \begin{minipage}{0.08\textwidth}
                        \centering
                        \includegraphics[width=\linewidth]{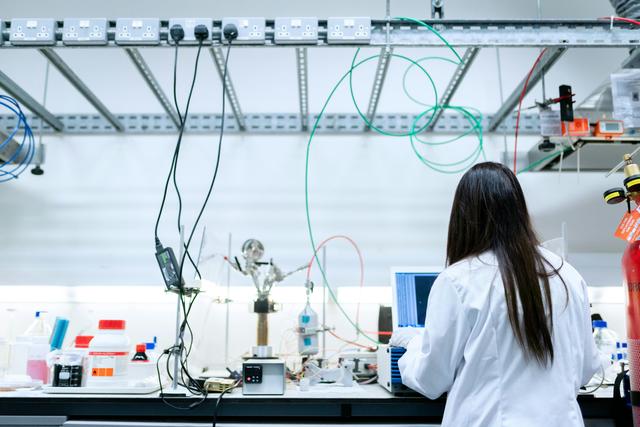}
                    \end{minipage}
                    \hfill
                    \begin{minipage}{0.08\textwidth}
                        \centering
                        \includegraphics[width=\linewidth]{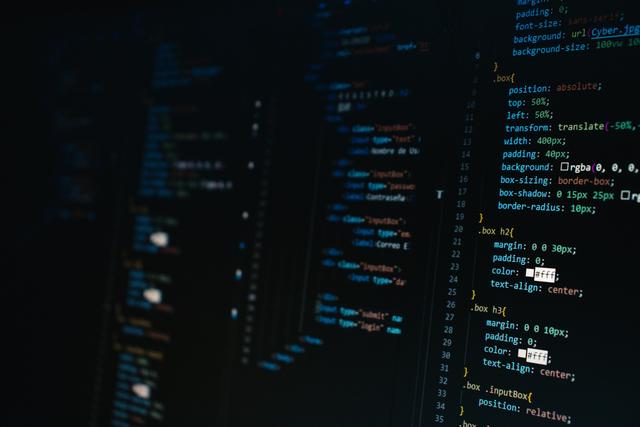}
                    \end{minipage}
                    \hfill
                    \begin{minipage}{0.08\textwidth}
                        \centering
                        \includegraphics[width=\linewidth]{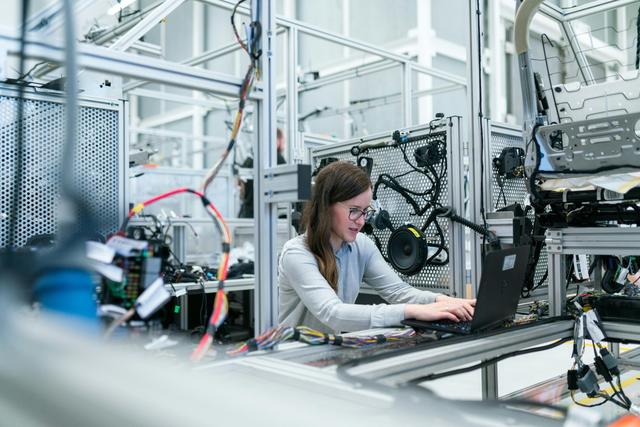}
                    \end{minipage}
                    \begin{minipage}{0.08\textwidth}
                        \centering
                        \includegraphics[width=\linewidth]{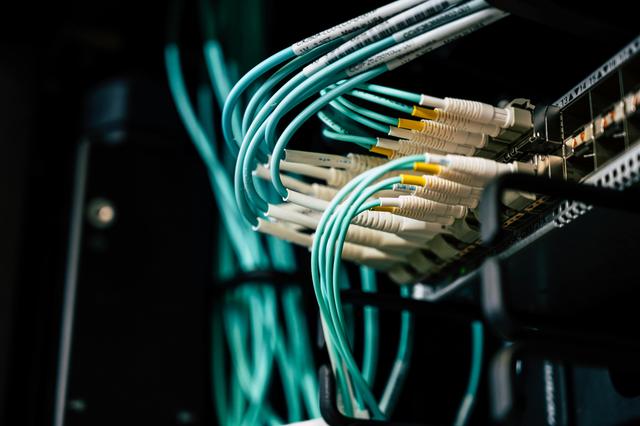}
                    \end{minipage}
                    \begin{minipage}{0.08\textwidth}
                        \centering
                        \includegraphics[width=\linewidth]{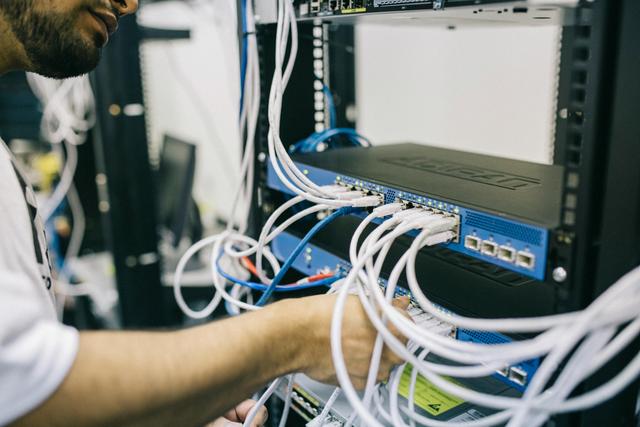}
                    \end{minipage}
                \\
    
            \textbf{Sports}  \\ \cmidrule(lr){1-1}
                    \begin{minipage}{0.08\textwidth}
                        \centering
                        \includegraphics[width=\linewidth]{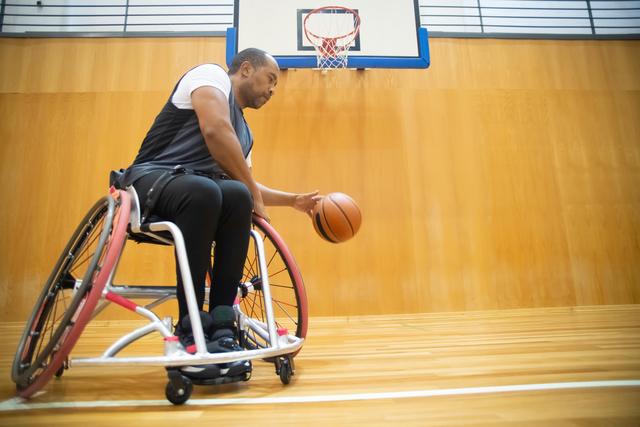}
                    \end{minipage}
                    \hfill
                    \begin{minipage}{0.08\textwidth}
                        \centering
                        \includegraphics[width=\linewidth]{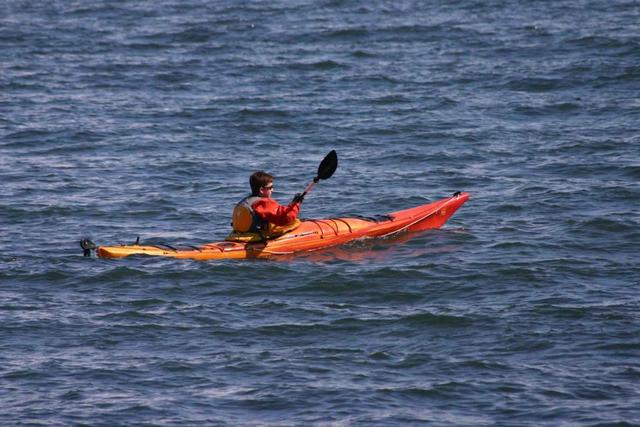}
                    \end{minipage}
                    \hfill
                    \begin{minipage}{0.08\textwidth}
                        \centering
                        \includegraphics[width=\linewidth]{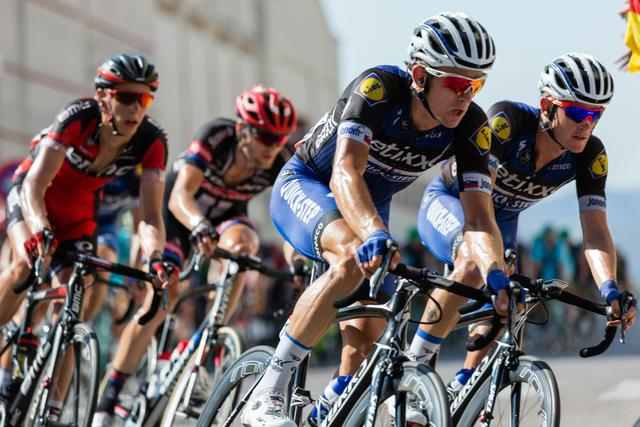}
                    \end{minipage}
                    \hfill
                    \begin{minipage}{0.08\textwidth}
                        \centering
                        \includegraphics[width=\linewidth]{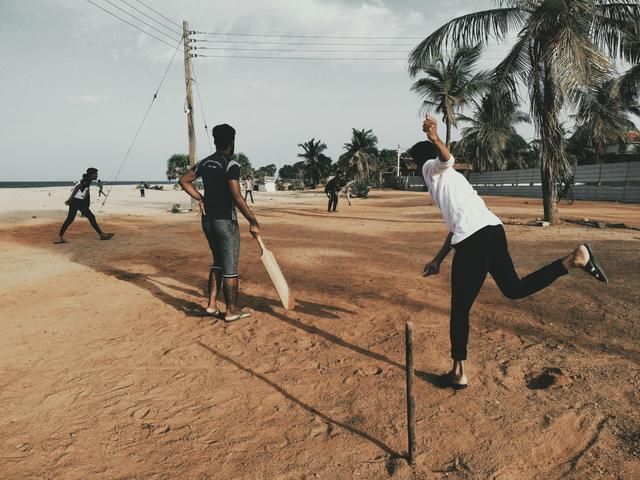}
                    \end{minipage}
                    \hfill
                    \begin{minipage}{0.08\textwidth}
                        \centering
                        \includegraphics[width=\linewidth]{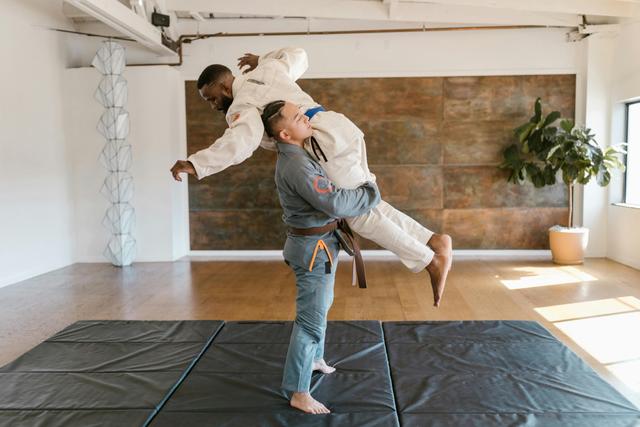}
                    \end{minipage}
                    \hfill
                    \begin{minipage}{0.08\textwidth}
                        \centering
                        \includegraphics[width=\linewidth]{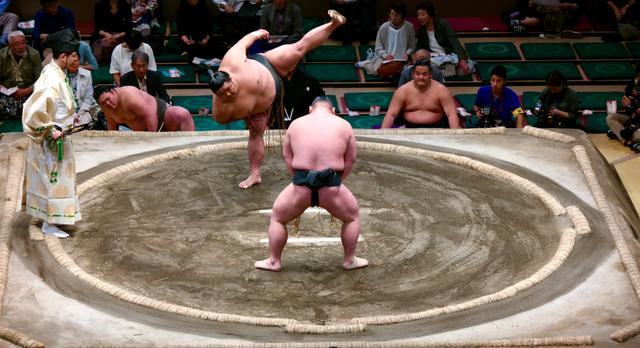}
                    \end{minipage}
                    \hfill
                    \begin{minipage}{0.08\textwidth}
                        \centering
                        \includegraphics[width=\linewidth]{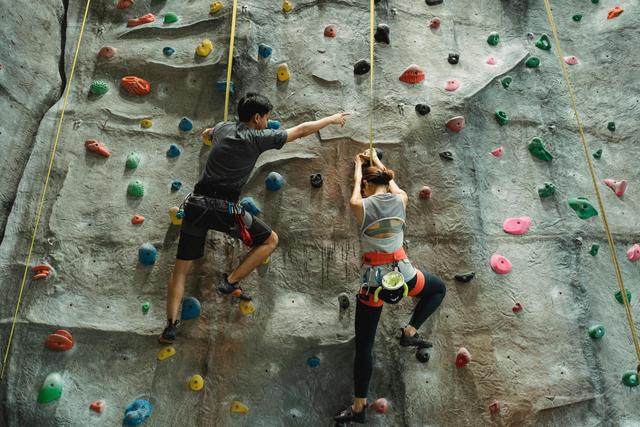}
                    \end{minipage}
                    \hfill
                    \begin{minipage}{0.08\textwidth}
                        \centering
                        \includegraphics[width=\linewidth]{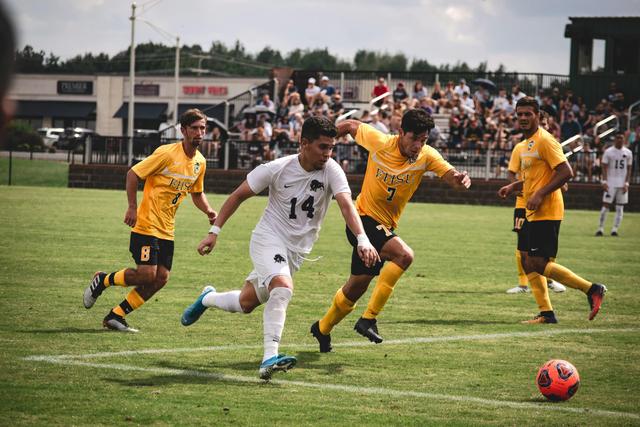}
                    \end{minipage}
                    \begin{minipage}{0.08\textwidth}
                        \centering
                        \includegraphics[width=\linewidth]{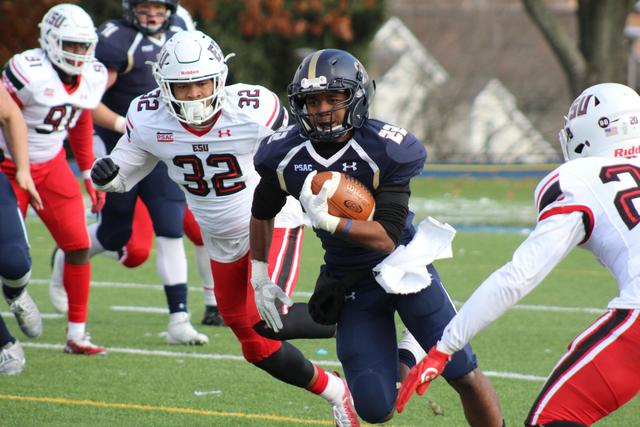}
                    \end{minipage}
                    \begin{minipage}{0.08\textwidth}
                        \centering
                        \includegraphics[width=\linewidth]{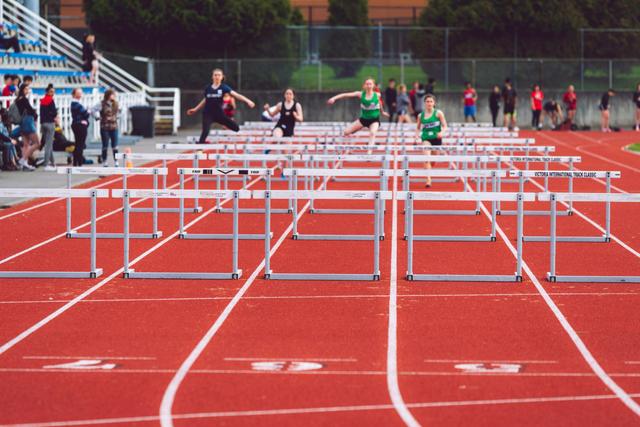}
                    \end{minipage}
                \\
    
            \textbf{Travel}  \\ \cmidrule(lr){1-1}
                    \begin{minipage}{0.08\textwidth}
                        \centering
                        \includegraphics[width=\linewidth]{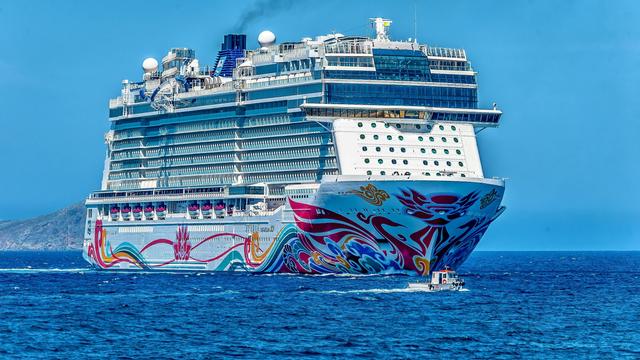}
                    \end{minipage}
                    \hfill
                    \begin{minipage}{0.08\textwidth}
                        \centering
                        \includegraphics[width=\linewidth]{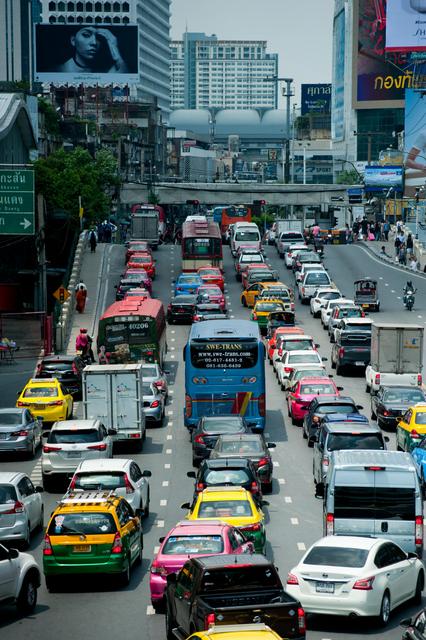}
                    \end{minipage}
                    \hfill
                    \begin{minipage}{0.08\textwidth}
                        \centering
                        \includegraphics[width=\linewidth]{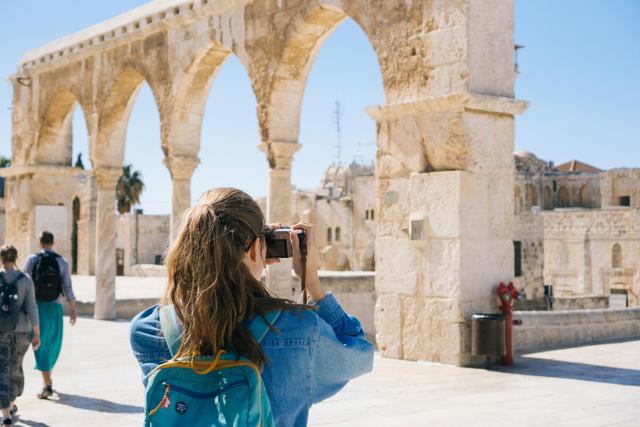}
                    \end{minipage}
                    \hfill
                    \begin{minipage}{0.08\textwidth}
                        \centering
                        \includegraphics[width=\linewidth]{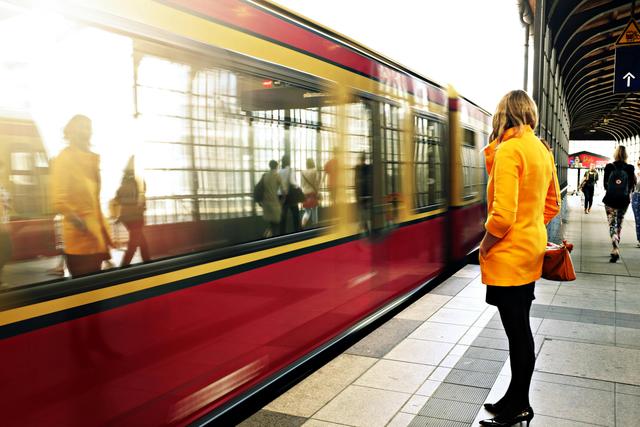}
                    \end{minipage}
                    \hfill
                    \begin{minipage}{0.08\textwidth}
                        \centering
                        \includegraphics[width=\linewidth]{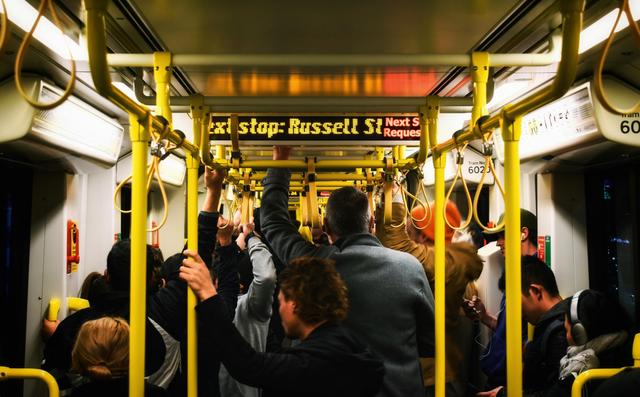}
                    \end{minipage}
                    \hfill
                    \begin{minipage}{0.08\textwidth}
                        \centering
                        \includegraphics[width=\linewidth]{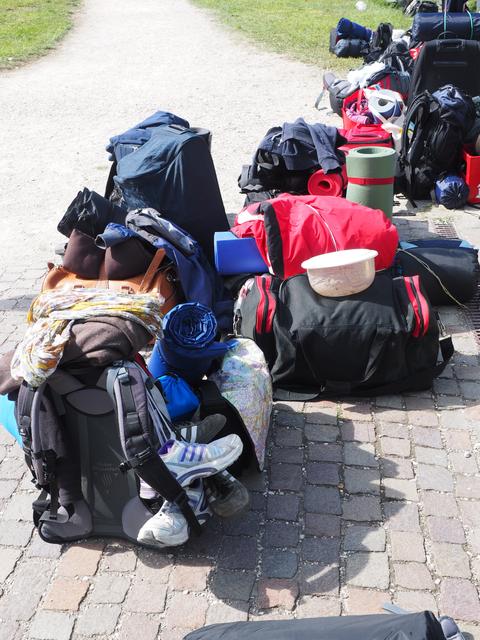}
                    \end{minipage}
                    \hfill
                    \begin{minipage}{0.08\textwidth}
                        \centering
                        \includegraphics[width=\linewidth]{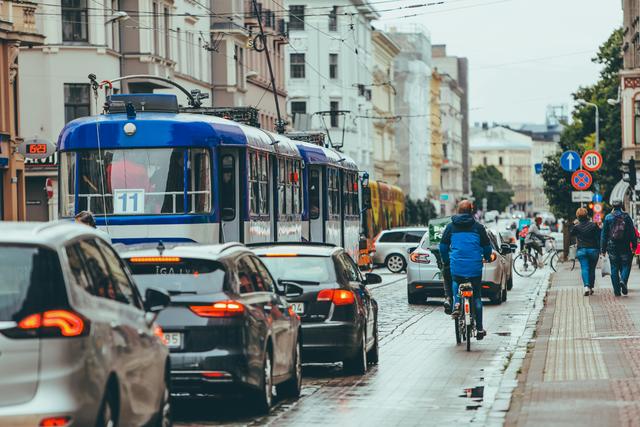}
                    \end{minipage}
                    \hfill
                    \begin{minipage}{0.08\textwidth}
                        \centering
                        \includegraphics[width=\linewidth]{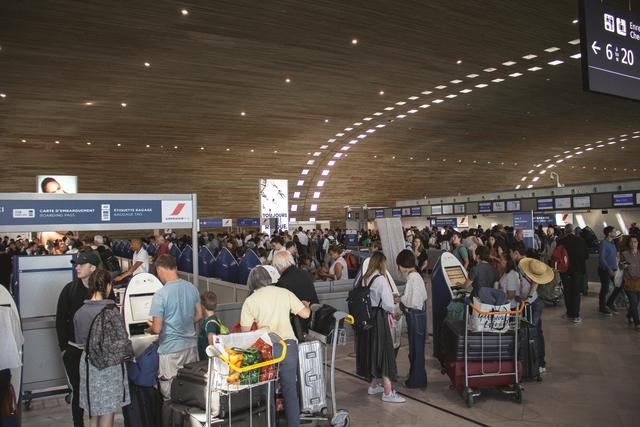}
                    \end{minipage}
                    \begin{minipage}{0.08\textwidth}
                        \centering
                        \includegraphics[width=\linewidth]{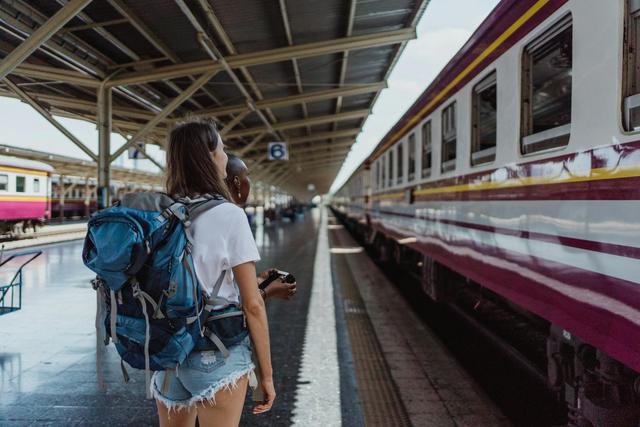}
                    \end{minipage}
                    \begin{minipage}{0.08\textwidth}
                        \centering
                        \includegraphics[width=\linewidth]{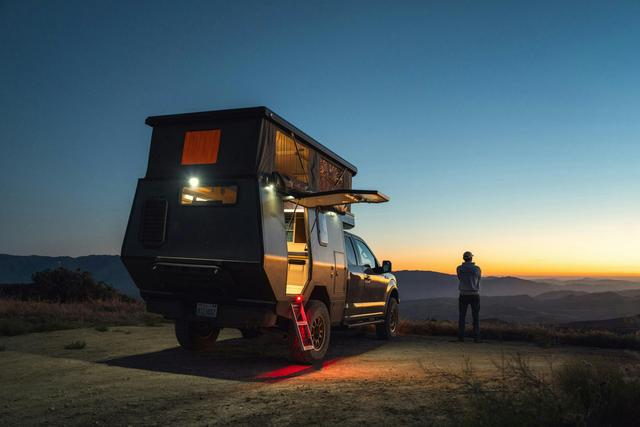}
                    \end{minipage}
                \\
         \bottomrule   
    \end{tabular}
\end{table*}
\clearpage

\begin{figure*}[ht]
\subsection{Prompts}
\label{appendix:prompts}

\begin{tcolorbox}[
    colback=white!95!black,    
    colframe=black,            
    title=Images-To-Sentences (\its), 
    fonttitle=\bfseries,       
    boxrule=0.5pt,             
    arc=4pt,                   
    outer arc=4pt,
    width=\textwidth,          
    enlarge left by=0mm,
    enlarge right by=0mm,
    before skip=1em,           
    after skip=1em,            
]

Which sentence best matches the topic of the images? The images and the sentences each belong
to one of the following topics: "entertainment", "geography", "health", "politics", "science and technology", "sports", or "travel". Choose one sentence from A, B, C, or D. Output only 
a single letter!

\medskip

\# Images

\begin{center}
    \begin{minipage}{0.18\textwidth}
        \centering
        \includegraphics[width=\linewidth]{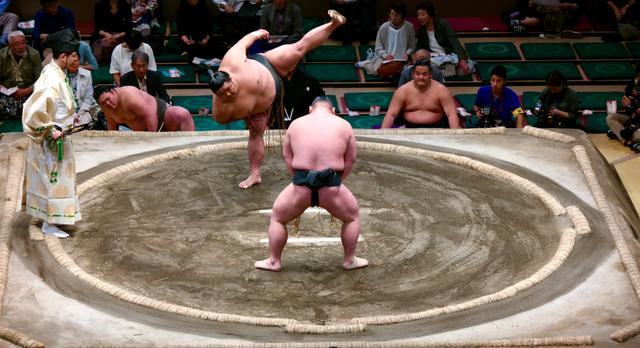}
    \end{minipage}
    \hfill
    \begin{minipage}{0.18\textwidth}
        \centering
        \includegraphics[width=\linewidth]{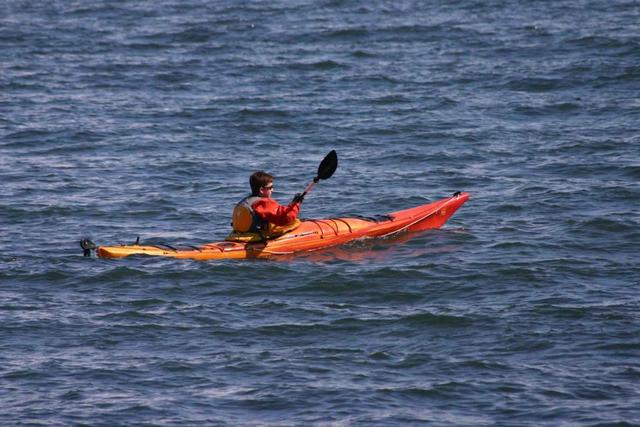}
    \end{minipage}
    \hfill
    \begin{minipage}{0.18\textwidth}
        \centering
        \includegraphics[width=\linewidth]{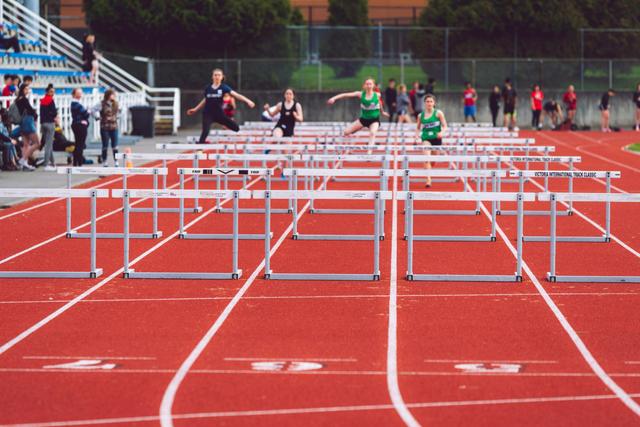}
    \end{minipage}
    \hfill
    \begin{minipage}{0.18\textwidth}
        \centering
        \includegraphics[width=\linewidth]{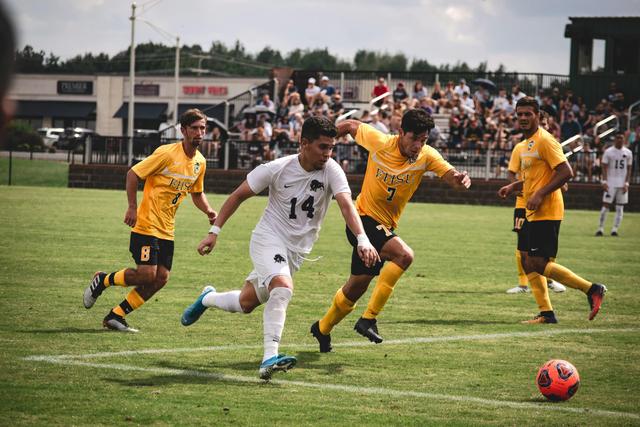}
    \end{minipage}
    \hfill
    \begin{minipage}{0.18\textwidth}
        \centering
        \includegraphics[width=\linewidth]{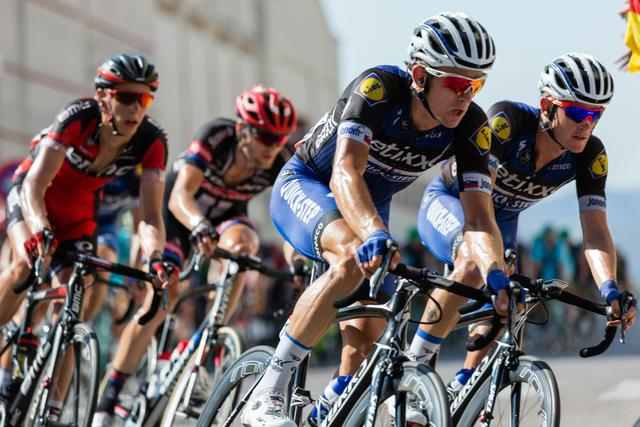}
    \end{minipage}
\end{center}

\medskip

\# Sentences

\begin{enumerate}[label=\Alph*., itemsep=0pt, topsep=0pt]
    \item \textasciigrave\textasciigrave\textasciigrave Maroochydore führte am Ende die Rangfolge an, mit sechs Punkten Vorsprung vor Noosa als Zweitem.\textasciigrave\textasciigrave\textasciigrave
    \item \textasciigrave\textasciigrave\textasciigrave 
    Es wurden keine schwere Verletzungen gemeldet, jedoch mussten mindestens fünf der zur Zeit der Explosion Anwesenden aufgrund von Schocksymptomen behandelt werden.\textasciigrave\textasciigrave\textasciigrave
    \item \textasciigrave\textasciigrave\textasciigrave 
    Finnland ist ein großartiges Reiseziel für Bootstouren. Das „Land der tausend Seen“ hat auch Tausende von Inseln – in den Seen und in den Küstenarchipelen.\textasciigrave\textasciigrave\textasciigrave
    \item \textasciigrave\textasciigrave\textasciigrave 
    Es ist auch nicht erforderlich, dass Sie eine lokale Nummer von der Gemeinde erhalten, in der Sie leben. Sie können eine Internetverbindung über Satellit in der Wildnis v on Chicken in Alaska erhalten und eine Nummer auswählen, die vorgibt, dass Sie im sonnigen Arizona sind.\textasciigrave\textasciigrave\textasciigrave
\end{enumerate}
Your answer letter: 
\end{tcolorbox}
\end{figure*}

\clearpage

\begin{figure*}[ht]
\begin{tcolorbox}[
    colback=white!95!black,    
    colframe=black,            
    title=Sentences-To-Images (\sti), 
    fonttitle=\bfseries,       
    boxrule=0.5pt,             
    arc=4pt,                   
    outer arc=4pt,
    width=\textwidth,          
    before skip=1em,           
    after skip=1em,            
]

Which image best matches the topic of the sentences?
The sentences and the images each belong to one of the following topics: 
"entertainment", "geography", "health", "politics", "science and technology", "sports", or "travel". 
Choose one image from A, B, C, or D. Output only a single letter!

\medskip

\# Sentences

\begin{itemize}[itemsep=0pt,topsep=2pt]
    \item \textasciigrave\textasciigrave\textasciigrave Maroochydore führte am Ende die Rangfolge an, mit sechs Punkten Vorsprung vor Noosa
 als Zweitem.\textasciigrave\textasciigrave\textasciigrave
    \item \textasciigrave\textasciigrave\textasciigrave 
    Die Schlagmänner der mittleren Reihe, Sachin Tendulkar und Rahul Dravid, zeigten gute Leistungen und erzielten eine Partnerschaft mit 100 Runs.\textasciigrave\textasciigrave\textasciigrave 
    \item \textasciigrave\textasciigrave\textasciigrave 
    Da pro Tag nur achtzehn Medaillen zur Verfügung stehen, hat es ein Anzahl an Ländern nicht auf das Podium geschafft.\textasciigrave\textasciigrave\textasciigrave 
    \item \textasciigrave\textasciigrave\textasciigrave 
    Wintersportarten sind in den nördlichen Regionen am beliebtesten und Italiener nehmen an internationalen Wettkämpfen und olympischen Spielen teil.\textasciigrave\textasciigrave\textasciigrave 
    \item \textasciigrave\textasciigrave\textasciigrave 
    Nach dem Rennen bleibt Keselowski mit 2.250 Punkten Spitzenreiter in der Fahrerwertung.
\end{itemize}

\medskip

\# Images

\begin{center}
    \begin{minipage}{0.23\textwidth}
        \centering
        A.
        \includegraphics[width=\linewidth]{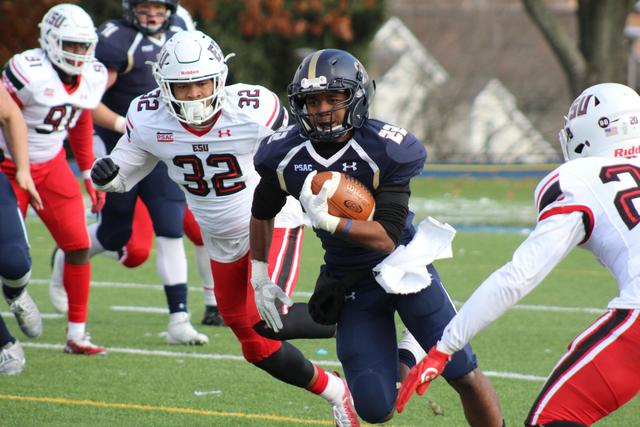}
    \end{minipage}
    \hfill
    \begin{minipage}{0.23\textwidth}
        \centering
        B. \vspace{0.1cm}
        \includegraphics[width=\linewidth]{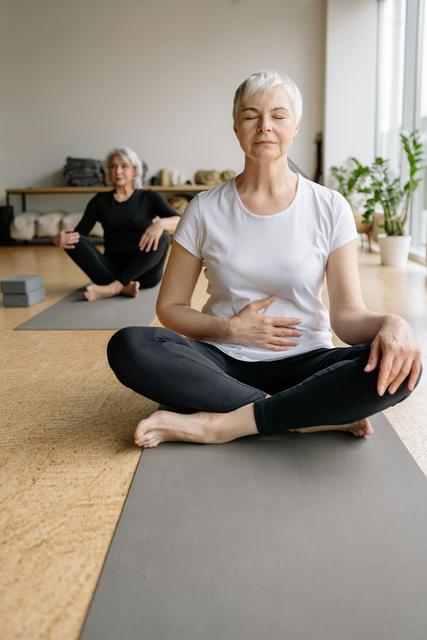}
    \end{minipage}
    \hfill
    \begin{minipage}{0.23\textwidth}
        \centering
        C. \vspace{0.1cm}
        \includegraphics[width=\linewidth]{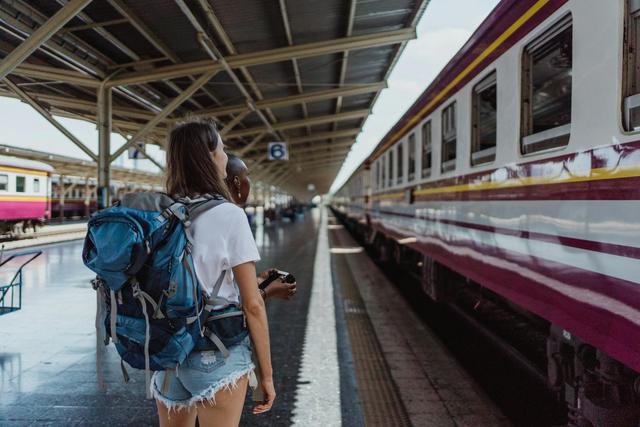}
    \end{minipage}
    \hfill
    \begin{minipage}{0.23\textwidth}
        \centering
        D. \vspace{0.1cm}
        \includegraphics[width=\linewidth]{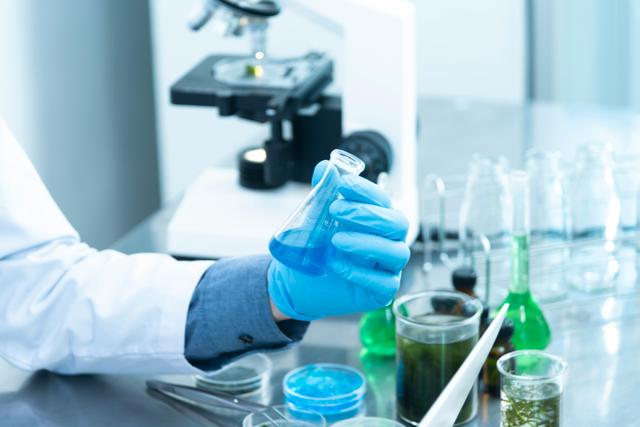}
    \end{minipage}
\end{center}

\medskip

Your answer letter: 

\end{tcolorbox}
\end{figure*}

\clearpage
\begin{figure*}[ht]
\begin{tcolorbox}[
    colback=white!95!black,    
    colframe=black,            
    title=Topic-To-Sentence (\tts), 
    fonttitle=\bfseries,       
    boxrule=0.5pt,             
    arc=4pt,                   
    outer arc=4pt,
    width=\textwidth,          
    before skip=1em,           
    after skip=1em,            
]

Which sentence best matches the topic "sports"? The sentences each belong to one of the following topics: "entertainment", "geography", "health", "politics", "science and technology", "sports", or "travel". Choose one sentence from A, B, C, or D. Output only
 a single letter!

\medskip

\# Sentences

\begin{enumerate}[label=\Alph*., itemsep=0pt, topsep=0pt]
    \item \textasciigrave\textasciigrave\textasciigrave Maroochydore führte am Ende die Rangfolge an, mit sechs Punkten Vorsprung vor Noosa als Zweitem.\textasciigrave\textasciigrave\textasciigrave
    \item \textasciigrave\textasciigrave\textasciigrave 
    Es wurden keine schwere Verletzungen gemeldet, jedoch mussten mindestens fünf der zur Zeit der Explosion Anwesenden aufgrund von Schocksymptomen behandelt werden.\textasciigrave\textasciigrave\textasciigrave
    \item \textasciigrave\textasciigrave\textasciigrave 
    Finnland ist ein großartiges Reiseziel für Bootstouren. Das „Land der tausend Seen“ hat auch Tausende von Inseln – in den Seen und in den Küstenarchipelen.\textasciigrave\textasciigrave\textasciigrave
    \item \textasciigrave\textasciigrave\textasciigrave 
    Es ist auch nicht erforderlich, dass Sie eine lokale Nummer von der Gemeinde erhalten, in der Sie leben. Sie können eine Internetverbindung über Satellit in der Wildnis v on Chicken in Alaska erhalten und eine Nummer auswählen, die vorgibt, dass Sie im sonnigen Arizona sind.\textasciigrave\textasciigrave\textasciigrave
\end{enumerate}

\medskip

Your answer letter: 

\end{tcolorbox}
\end{figure*}

\clearpage
\begin{figure*}[ht]
\begin{tcolorbox}[
    colback=white!95!black,    
    colframe=black,            
    title=Sentences-To-Topics (\stt), 
    fonttitle=\bfseries,       
    boxrule=0.5pt,             
    arc=4pt,                   
    outer arc=4pt,
    width=\textwidth,          
    before skip=1em,           
    after skip=1em,            
]

Which topic best matches the sentences? The sentences belong to one of the following topics: "entertainment", "geography", "health", "politics", "science and technology", "sports", or "travel". Choose one topic from A, B, C, or D. Output only a single letter!

\medskip

\# Sentences

\begin{itemize}[itemsep=0pt,topsep=2pt]
    \item \textasciigrave\textasciigrave\textasciigrave Maroochydore führte am Ende die Rangfolge an, mit sechs Punkten Vorsprung vor Noosa
 als Zweitem.\textasciigrave\textasciigrave\textasciigrave
    \item \textasciigrave\textasciigrave\textasciigrave 
    Die Schlagmänner der mittleren Reihe, Sachin Tendulkar und Rahul Dravid, zeigten gute Leistungen und erzielten eine Partnerschaft mit 100 Runs.
    \item \textasciigrave\textasciigrave\textasciigrave 
    Da pro Tag nur achtzehn Medaillen zur Verfügung stehen, hat es ein Anzahl an Ländern nicht auf das Podium geschafft.\textasciigrave\textasciigrave\textasciigrave 
    \item \textasciigrave\textasciigrave\textasciigrave 
    Wintersportarten sind in den nördlichen Regionen am beliebtesten und Italiener nehmen an internationalen Wettkämpfen und olympischen Spielen teil.\textasciigrave\textasciigrave\textasciigrave 
    \item \textasciigrave\textasciigrave\textasciigrave 
    Nach dem Rennen bleibt Keselowski mit 2.250 Punkten Spitzenreiter in der Fahrerwertung.\textasciigrave\textasciigrave\textasciigrave 
\end{itemize}

\medskip

\# Topics

\begin{enumerate}[label=\Alph*., itemsep=0pt, topsep=0pt]
    \item sports
    \item health
    \item travel
    \item science and technology
\end{enumerate}

\medskip

Your answer letter: 

\end{tcolorbox}
\end{figure*}
\clearpage

\clearpage
\subsection{Further Details}
\label{app:subsec:exp-details}

\rparagraph{Models} We test state-of-the-art instruction fine-tuned LVLMs across various sizes. Smaller models are evaluated on all languages, while larger LVLMs (26B+) are tested on subsets of the MVL-SIB languages (cf. \S\ref{subsec:further-analyses}).

\iparagraph{GPT-4o} We evaluate MVL-SIB on GPT-4o-mini-\textit{2024-07-18} and GPT-4o-\textit{2024-08-06}. We set the image detail in the API to `low', since our tasks require high-level reasoning that does not depend on finer image details.\footnote{We use GPT-4o-mini because evaluating GPT-4o would be too expensive.} Prior works show that GPT-4o is the best-performing multilingual LVLM \cite{schneider-sitaram-2024-m5, vayani2024alm}.

\iparagraph{Qwen2-VL} Qwen2-VL ties a 675M parameter vision-transformer (ViT) into Qwen2 LLMs \cite{wang2024qwen2vl}. An MLP compresses adjacent 2×2 visual tokens embedded by the ViT into one token representation, which is then input to the LLM. 

\iparagraph{InternVL 2.5} Depending on the model size, InternVL uses Qwen2.5 or InternLM as its LLM backbone  \cite{chen2024internvl}. The model embeds images either by a 6B or by a distilled 300M ViT pretrained with CLIP \cite{radford2021clip}. The resulting image patch encodings are downsampled by factor 4 and fed through an MLP to the LLM.

\iparagraph{Centurio} Centurio is the latest massively multilingual LVLM trained on 100 languages \cite{geigle2025centurio}, outperforming alternatives like Parrot \cite{sun2024parrot} or Pangea \cite{yue2024pangea}. It employs Qwen2.5 as its LLM \cite{yang2024qwen2technicalreport} and \texttt{SigLIP SO400/384} as its ViT \cite{zhai2023sigmoid}. The model mixes resolutions by stacking the encodings of the full image and those of 2×2 tiles along the features. The combined embedding is then projected via an MLP to the LLM's input space.

Besides architectures, the LVLMs crucially differ in dataset mixtures on which they were trained. Centurio translates image-caption, VQA, OCR, and a few multi-image datasets to 100 languages with NLLB \cite{nllbteam2022language} to mix 50:50 with the original English data. Qwen2-VL and InternVL, however, were trained on much larger, more diverse datasets that comprise sizable multi-image comparison and video understanding datasets. Moreover, assuming that the LLMs of Qwen2-VL, InternVL, and GPT-4o were pretrained on Flores, their performance would be overly optimistic.

\begin{figure*}[ht]
  \subsubsection{Performance by Model over Languages grouped by Language Tier}
  \label{app:tier-perf-plot}
  \centering
  \adjustbox{max width=\textwidth}{%
    \includegraphics{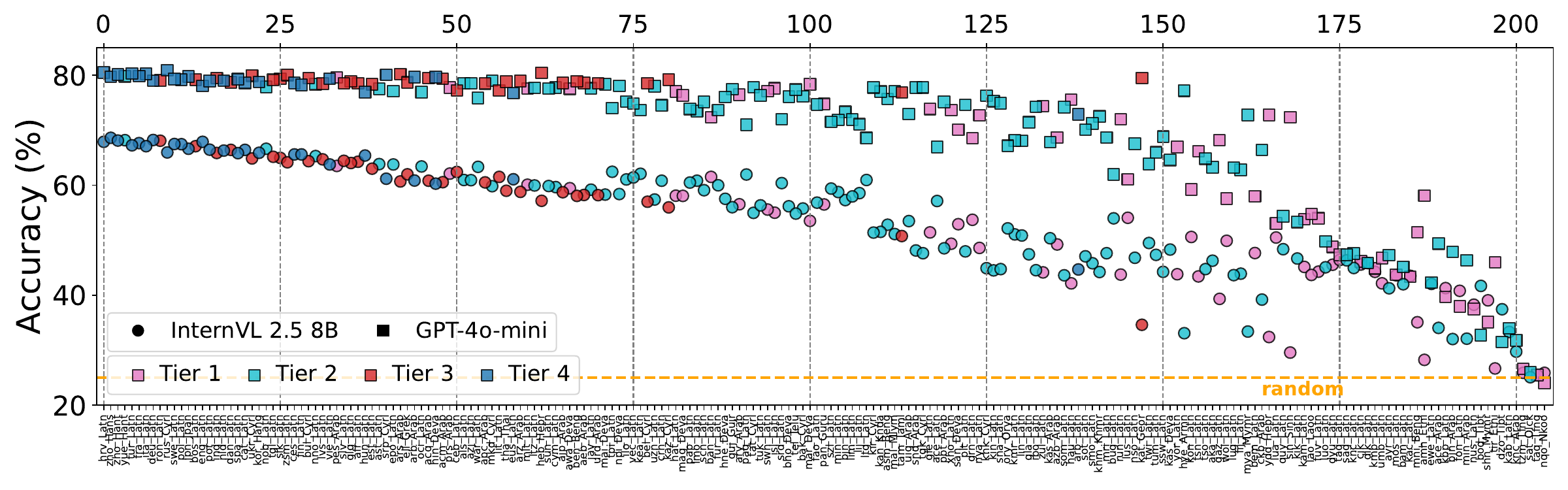}%
  }
  \vspace{-0.8cm}
  \caption{\textbf{Images-To-Sentences~@~$k{=}3$.} The English prompt describes the cross-modal  topic matching task, lists all topics, and provides both $k{=}3$ reference images and 4 sentences in the corresponding language \{\texttt{eng\_Latn}, $\dots$, \texttt{nqo\_Nkoo}\}. LVLMs must select the sentence of 4 options that topically fits $k{=}3$ reference images. The sentences spanning 205 languages and 7 topics are drawn from SIB-200 \cite{adelani-etal-2024-sib}, while images for the topics were hand-selected (cf. Appendix \ref{app:images-per-topic}). An example prompt is shown in Appendix \ref{app:images-to-sentences}; further details are in \S\ref{sec:experimental-setup}. \\ \textbf{Plot.} The x-axis orders the languages of the candidate sentences \{\texttt{eng\_Latn}, $\dots$, \texttt{nqo\_Nkoo}\}, respectively, by descending performance (y-axis). The top x-axis indicates the running index of each language $L_i$ ($i \in \{1, \dots, 205\}$). \\
  \textbf{Tiers.} The languages are grouped by tiers derived from \citet{joshi-etal-2020-state} (cf. \S\ref{sec:results}).}
  \vspace{-0.1cm}
  \label{fig:tier-img2sent}
\end{figure*}

\begin{figure*}[ht]
 \subsubsection{Calibration Analysis of Cross-Modal Topic Matching}
 \label{app:fig:calibration}
 \centering
 \adjustbox{max width=\textwidth}{%
   \includegraphics{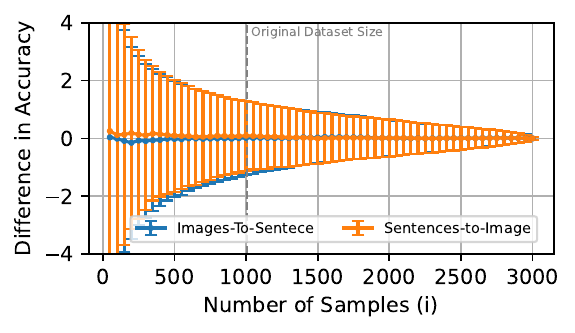}%
 }
 \caption{ \textbf{Calibration Analysis of Cross-Modal Topic Matching for InternVL 2.5 8B.} \textbf{Analysis:} To assess the reliability of cross-modal topic matching with fewer samples than our full dataset (1004 samples per language), we randomly select 500 trajectories. We then compute performance metrics on cumulative subsets, incrementing by 50 examples at each step. The difference in performance between the full dataset and each subset is calculated to quantify the deviation at each sample size. \textbf{Plot:} The plot displays the average absolute spread in performance (averaged over all languages) along with the standard deviation for InternVL 2.5 8B. We restrict ourselves to a single open-weight LVLM, since the analysis yields identical results across all combinations of LVLMs and language tiers. \textbf{Insights:} Our analysis shows that performance stabilizes rapidly, with deviations of only about 1\% observed at 1,004 instances -- same size as the subset from which the dataset was created. This indicates that reliable evaluation of cross-modal topic matching can be achieved with far fewer than 3,012 samples. }
 \vspace{-0.1cm}
\end{figure*}

\clearpage

\subsection{Overview of Multilingual Vision-Language Benchmarks}
\label{app:summary-mvl-benchmarks}

\begin{table*}[ht]
    \centering
\adjustbox{max width=\textwidth}{%
\large
\begin{tabular}{lcccccc}
\toprule
\textbf{Task} & \textbf{Dataset} & \textbf{Visual Input} & \textbf{Textual Input} & \textbf{Target Output} & \textbf{Metric} & \textbf{\#Lang.} \\
\midrule
\multirow{2}{*}{Captioning} 
    & \multirow{2}{*}{XM3600} & \multirow{2}{*}{Single-Image} & \multirow{2}{*}{Prompt (English)} & \multirow{2}{4cm}{\centering Caption\\(Target Language)} & CIDEr & $36$ \\
   & & & & & &  \\ 
\midrule
\multirow{2}{4cm}{Multiple-Choice Visual Question Answering} 
    & \multirow{2}{*}{BabelImageNet-MC} 
    & \multirow{2}{*}{Single-Image} 
    & \multirow{2}{*}{Question (Target Language)} 
    & \multirow{2}{*}{Letter of the correct Choice} 
    & \multirow{2}{*}{Relaxed Accuracy} 
    & \multirow{2}{*}{$20$} \\
   & & & & & &  \\ 
\midrule
\multirow{3}{4cm}{Text-Heavy Multiple-Choice Visual Question Answering} 
    & M3Exam 
    & \multirow{3}{*}{Single or Multi-Image} 
    & Question (Target Language) 
    & \multirow{3}{*}{Letter of the correct Choice} 
    & \multirow{3}{*}{Relaxed Accuracy} 
    & $7$ \\
    & MMMU     
    &  
    & Question (English)        
    &  
    &  
    & $1$ \\
    & xMMMU    
    &  
    & Question (Target Language) 
    &  
    &  
    & $7$ \\
\midrule
\multirow{2}{4cm}{Text-Heavy Visual\\Question Answering} 
    & MTVQA       
    & \multirow{2}{*}{Single-Image} 
    & \multirow{2}{*}{Question (Target Language)} 
    & \multirow{2}{*}{Word or Phrase (Target Language)} 
    & \multirow{2}{*}{Exact Accuracy} 
    & $9$ \\
    & SMPQA - Name 
    &  
    &  
    &  
    &  
    & $11$ \\
\midrule
\multirow{2}{4cm}{Text-Heavy Visually\\ Grounded Reasoning}  & \multirow{2}{*}{SMPQA - Ground} & \multirow{2}{*}{Single-Image} & \multirow{2}{*}{Question (Target Language)} & \multirow{2}{*}{'yes' / 'no'} & \multirow{2}{*}{Exact Accuracy} & \multirow{2}{*}{$11$} \\
& & & & & &  \\ 
\midrule
\multirow{2}{4cm}{Visio-Linguistic \\ Outlier Detection} 
    & \multirow{2}{*}{M5B-VLOD} & \multirow{2}{*}{Multi-Image} & \multirow{2}{*}{Hypothesis (Target Language)} & \multirow{2}{*}{Letter of the correct Choice} & \multirow{2}{*}{Relaxed Accuracy} & \multirow{2}{*}{$12$} \\
& & & & & &  \\ 
\midrule
\multirow{2}{4cm}{Visual Natural\\ Language Inference} 
    & \multirow{2}{*}{XVNLI} & \multirow{2}{*}{Single-Image} & \multirow{2}{*}{Hypothesis (Target Language)} & \multirow{2}{*}{'yes' / 'no' / 'maybe'} & \multirow{2}{*}{Exact Accuracy} & \multirow{2}{*}{$5$} \\
   & & & & & &  \\ 
\midrule
\multirow{2}{*}{Visual Question Answering} 
    & MaXM 
    & \multirow{2}{*}{Single-Image} 
    & \multirow{2}{*}{Question (Target Language)} 
    & Word or Phrase (Target Language) 
    & \multirow{2}{*}{Exact Accuracy} 
    & $6$ \\
    & xGQA 
    &  
    &  
    & Word or Phrase (English) 
    &  
    & $8$ \\ \midrule
    \multirow{2}{*}{Visually Grounded Reasoning} 
    & M5B-VGR
    & \multirow{2}{*}{Multi-Image} 
    & \multirow{2}{*}{Hypothesis (Target Language)} 
    & \multirow{2}{*}{'yes' / 'no'} 
    & \multirow{2}{*}{Exact Accuracy} 
    & $12$ \\
    & MaRVL   
    &  
    &  
    &  
    &  
    & $6$ \\

\bottomrule
\end{tabular}    }
    \caption{Summary of multilingual vision-language benchmarks we correlate MVL-SIB against. Relaxed 
 denotes responses that start with the correct option letter (cf. \ref{sec:experimental-setup}).}
    \label{appendix:tab:eval_suite_datasets}
\end{table*}

\rparagraph{xGQA} The xGQA dataset~\citep{pfeiffer-etal-2022-xgqa} is a cross-lingual visual question-answering resource. It extends the well-known English-only GQA dataset~\citep{hudson_gqa_2019} by providing manual translations of the questions in the balanced \textit{test-dev} set. The dataset contains $9666$ questions available in eight languages across five scripts, while the answers remain in English. In addition, it features $300$ unique images from Visual Genome~\citep{krishna_visual_2017}.

%
\rparagraph{MaXM}
MaXM, introduced by~\citet{changpinyo-etal-2023-maxm}, is a VQA dataset covering seven languages written in five scripts. In this dataset, both the questions and their corresponding answers are presented in the same language. The images are drawn from a subset of the XM3600~\citep{thapliyal_xm3600_2022} dataset and are selected to correspond to regions where the question-answer pair’s language is spoken, ensuring both linguistic and cultural diversity.

%
\rparagraph{XVNLI}
The XVNLI dataset~\citep{bugliarello-etal-2022-iglue} introduces the task of Cross-lingual Visual Natural Language Inference, where a model must determine if a textual hypothesis \textit{entails}, \textit{contradicts}, or is \textit{neutral} with respect to a visual premise. This dataset spans five languages across three scripts and includes $357$ unique images from Visual Genome. It is built upon a combination of the text-only SNLI~\citep{bowman_large_2015} dataset and its cross-lingual~\citep{agic_cli_2018} and cross-modal~\citep{xie_visual_entailment_2019} counterparts.

%
\rparagraph{MaRVL}
The MaRVL dataset~\citep{liu-etal-2021-visually} is designed to benchmark models on Multicultural Reasoning over Vision and Language. Each sample consists of two images, a textual statement, and a binary (true/false) answer grounded in the images. Covering five languages across three scripts, MaRVL includes $4914$ culturally diverse images that align with the respective languages. The images in each sample are selected to reflect the culture of the annotator who composed the textual statement in their native language.

%
\rparagraph{XM3600}
The XM3600 dataset~\citep{thapliyal_xm3600_2022} is an extensive multilingual image captioning resource encompassing 36 languages. It contains $261375$ captions across 13 scripts, with 100 unique images per language. The images are chosen to reflect the cultural background of the language, ensuring both cultural and linguistic diversity. All captions were manually produced by professional native speakers rather than being automatically generated. Due to the dataset's large size, we evaluate XM3600 using a randomly selected subset of 500 images per language.

%
\rparagraph{Babel-ImageNet (multiple-choice) (BIN-MC)}
Babel-ImageNet~\cite{geigle-etal-2024-babel} translates ImageNet’s~\cite{deng_imagenet_2009} labels into nearly 300 languages, allowing us to assess whether models can recognize and correctly link diverse ImageNet objects to their labels in the target language. Given the computational cost, we focus on languages that appear in only one or two other datasets, in addition to English and a select few high-resource languages, and we use 10 images per class instead of 50. We follow \citet{geigle-etal-2024-african} and frame the task as a multiple-choice problem by mining hard negative options from the complete label pool. This approach avoids the ambiguity inherent in traditional open-ended VQA formats. Negatives are selected based on the English labels, filtering out candidates not translated by Babel-ImageNet into the target language, and ultimately choosing the three most similar negative labels available.

%
\rparagraph{SMPQA}
\citet{geigle2025centurio} introduce SMPQA (Synthetic Multilingual Plot QA) as a test dataset for evaluating multilingual OCR capabilities for bar plots and pie charts, covering 11 languages and various scripts and resource levels.

%
\rparagraph{M5B-VGR}
The M5B-VGR dataset, presented by~\cite{schneider-sitaram-2024-m5}, is a visually grounded reasoning benchmark akin to MaRVL. Each sample comprises two images, a textual statement, and a binary (true/false) answer based on the images. It spans 12 languages across 7 scripts and features culturally diverse photos from regions where the respective languages are spoken. The images are sampled from the Dollar Street~\cite{gaviria2022dollar} dataset, with 120 samples provided per language.

%
\rparagraph{M5B-VLOD}
The M5B-VLOD (Visio-Linguistic Outlier Detection) dataset, also introduced by~\cite{schneider-sitaram-2024-m5}, consists of samples containing five images paired with a textual statement that is true for all but one image. The task is to identify the outlier image that does not match the statement. This dataset covers the same 12 languages as M5B-VGR, with images sampled in a similar manner from the same source, and provides 120 samples per language.

%
\rparagraph{MTVQA}
The MTVQA dataset, introduced by~\cite{tang2024mtvqa}, features text-heavy visual question answering tasks. It includes expert human annotations in 9 diverse languages, comprising 6778 question-answer pairs across 2116 images. The images predominantly contain text in the corresponding language, with questions and answers closely tied to that text. These images are sourced from various publicly available datasets.

%
\rparagraph{CVQA}
The CVQA dataset, introduced by~\cite{romero2024cvqa}, is a multilingual and culturally nuanced VQA benchmark that includes a broad array of languages, many of which are underrepresented in NLP. It consists of 10000 questions spanning 30 countries and 31 languages, forming 39 distinct country-language pairs (for instance, Spanish appears in 7 different splits corresponding to 7 Spanish-speaking countries). The images were manually collected by human annotators to accurately depict the culture associated with each country-language pair. Each sample includes one image and a question in the respective language. Although the test set is not publicly available, the authors permit up to 5 daily leaderboard submissions for evaluation.

%
\rparagraph{M3Exam}
The M3Exam dataset, presented by~\cite{zhang_m3exam_2023}, contains real-world exam questions in 9 languages, available as either text-only or multimodal samples. For our evaluation, we only include samples that require at least one image. Due to the limited number of samples for Swhalili and Javanese, we focus on the remaining 7 languages. The selected samples consist of multiple-choice questions in the target language, accompanied by up to 8 images that may appear in both the question and the answer options, with the number of choices ranging from 4 to 8 per sample.

%
\rparagraph{xMMMU}
xMMMU, introduced by~\cite{yue2024pangea}, comprises college-level multiple-choice VQA samples in seven languages. It was automatically translated using GPT4o from a randomly selected subset of 300 questions from the MMMU~\cite{yue_mmmu_2023} validation split.

\subsection{Prompts}
\label{appendix:sec:evaluation:prompts}
We list the prompts for each dataset in our test suite used for all models in Figure~\ref{appendix:fig:evaluation:prompts}.
\begin{figure*}[ht!]\tiny
    \centering
    \begin{promptbox}{SMPQA}
    <IMG>\{QUESTION\}\textbackslash{n}Answer the question using a single word or phrase.
    \end{promptbox}
    \begin{promptbox}{CVQA}
    <IMG>\{QUESTION\}\textbackslash{n}There are several options:\textbackslash{n}A. \{OPTION A\}\textbackslash{n}B. \{OPTION B\}\textbackslash{n}C. \{OPTION C\}\textbackslash{n}D. \{OPTION D\}\textbackslash{n}Answer with the option's letter from the given choices directly.
    \end{promptbox}
    \begin{promptbox}{xMMMU}
    \{QUESTION\}\textbackslash{n}There are several options:\textbackslash{n}A. \{OPTION A\}\textbackslash{n}B. \{OPTION B\}\textbackslash{n}C. \{OPTION C\}\textbackslash{n}D. \{OPTION D\}\textbackslash{n}Answer with the option's letter from the given choices directly.
    \end{promptbox}
    \begin{promptbox}{MTVQA}
    <IMG>\{QUESTION\}\textbackslash{n}Answer the question using a single word or phrase.\textbackslash{n}Answer in \{LANGUAGE\}.
    \end{promptbox}
    \begin{promptbox}{M3Exam}
    \{QUESTION\}\textbackslash{n}Options:\textbackslash{n}A. \{OPTION A\}\textbackslash{n}B. \{OPTION B\}\textbackslash{n}C. \{OPTION C\}\textbackslash{n}D. \{OPTION D\}\textbackslash{n} Answer with the option's letter from the given choices directly.
    \end{promptbox}
    \begin{promptbox}{BIN-MC}
    <IMG>Which of these choices (in English) is shown in the image?\textbackslash{n} Choices:\textbackslash{n}A. \{CHOICE A\}\textbackslash{n}B. \{CHOICE B\}\textbackslash{n}C. \{CHOICE C\}\textbackslash{n}D. \{CHOICE D\}\textbackslash{n} Answer with the letter from the given choices directly.
    \end{promptbox}
    \begin{promptbox}{xGQA}
    <IMG>\{QUESTION\}?\textbackslash{n}Answer the question using a single word or phrase.\textbackslash{n}Answer in English.
    \end{promptbox}
    \begin{promptbox}{MaXM}
    <IMG>\{QUESTION\}?\textbackslash{n}Answer the question using a single word or phrase.\textbackslash{n}Answer in \{LANGUAGE\}.
    \end{promptbox}
    \begin{promptbox}{MaRVL}
    <IMG>Given the two images <IMG><IMG>, is it correct to say ``\{HYPOTHESIS\}''? Answer yes or no.'
    \end{promptbox}
    \begin{promptbox}{XVNLI}
    <IMG>Is it guaranteed true that ``\{HYPOTHESIS\}''? Yes, no, or maybe? Answer in English.
    \end{promptbox}
    \begin{promptbox}{M5-VGR}
    Given the two images <IMG><IMG>, is it correct to say ``\{HYPOTHESIS\}''? Answer yes or no.'
    \end{promptbox}
    \begin{promptbox}{M5-VLOD}
    Based on the 5 images <IMG><IMG><IMG><IMG><IMG> ordered from top-left to bottom-right, which image does not match the hypothesis ``\{HYPOTHESIS\}''? Choose one from [A, B, C, D, E] and only output a single letter:
    \end{promptbox}
    \begin{promptbox}{XM3600}
    Briefly describe the image in \{LANGUAGE\} in one sentence.
    \end{promptbox}
    \caption{Prompts used for the different datasets of our test suite. For M3Exam and xMMMU, the questions contain images at individual positions, and also the options can consist of images. In total, a sample of M3Exam can contain up to 8 images and 8 options, and a sample of xMMMU can contain up to 4 images and 4 options.}
    \label{appendix:fig:evaluation:prompts}
\end{figure*}
\begin{table*}[ht]
\subsection{Full Results For Subsets by Task, Model, and Language}
\label{app:subsets-full-results}
    \rowcolors{2}{gray!15}{white} 
    \begin{adjustbox}{max width=\textwidth}
    \centering
    \small
\begin{tabular}{l c | c c c | c c c c c | c | c c}
    \toprule
    & & \multicolumn{3}{c|}{\textbf{Qwen2-VL}} & \multicolumn{5}{c|}{\textbf{InternVL 2.5}} & \multicolumn{1}{c|}{\textbf{Centurio Qwen}} & \multicolumn{2}{c}{\textbf{GPT-4o}} \\
    \cmidrule(lr){3-5} \cmidrule(lr){6-10}  \cmidrule(lr){11-11} \cmidrule(lr){12-13}
    \textbf{Lang} & \textbf{Tier} & Qwen2-VL 2B & Qwen2-VL 7B & Qwen2-VL 72B & InternVL 2.5 4B & InternVL 2.5 8B & InternVL 2.5 26B & InternVL 2.5 38B & InternVL 2.5 78B & Centurio Qwen 8B & GPT-4o-mini & GPT-4o \\
    \midrule
\texttt{acm\_Arab} & 3 & 34.9 & 55.4 & 80.8 & 50.0 & 61.0 & 74.1 & 80.2 & 82.2 & 54.4 & 72.0 & N/A \\
\texttt{aka\_Latn} & 2 & 28.7 & 38.0 & 57.2 & 34.6 & 46.3 & 51.3 & 58.3 & 57.0 & 42.4 & 61.4 & N/A \\
\texttt{apc\_Arab} & 3 & 34.9 & 54.8 & 80.0 & 49.1 & 61.3 & 74.2 & 78.8 & 82.1 & 54.4 & 71.5 & N/A \\
\texttt{arb\_Arab} & 4 & 35.2 & 57.1 & 81.4 & 51.7 & 62.2 & 75.8 & 80.8 & 82.8 & 55.2 & 73.2 & N/A \\
\texttt{azj\_Latn} & 2 & 34.6 & 53.8 & 80.5 & 49.5 & 58.6 & 70.9 & 78.3 & 80.0 & 51.8 & 72.8 & N/A \\
\texttt{bul\_Cyrl} & 3 & 35.0 & 56.6 & 79.3 & 50.3 & 65.3 & 77.1 & 79.8 & 81.6 & 53.9 & 71.5 & N/A \\
\texttt{ces\_Latn} & 4 & 36.2 & 58.2 & 79.5 & 52.0 & 65.6 & 78.5 & 79.5 & 81.5 & 53.2 & 72.7 & N/A \\
\texttt{eng\_Latn} & 4 & 36.3 & 65.8 & 79.8 & 52.5 & 67.7 & N/A & 79.7 & 81.7 & 54.8 & 68.3 & N/A \\
\texttt{fin\_Latn} & 4 & 33.0 & 57.1 & 79.1 & 49.8 & 64.1 & 76.5 & 79.2 & 81.1 & 54.1 & 70.2 & N/A \\
\texttt{hat\_Latn} & 1 & 32.1 & 50.4 & 77.8 & 43.0 & 54.7 & 66.1 & 74.1 & 74.2 & 54.9 & 69.3 & N/A \\
\texttt{hau\_Latn} & 1 & 29.1 & 39.2 & 56.3 & 32.4 & 41.8 & 45.2 & 58.0 & 57.5 & 43.9 & 70.2 & N/A \\
\texttt{min\_Arab} & 2 & 27.0 & 30.4 & 58.4 & 27.8 & 33.6 & 36.3 & 53.4 & 47.8 & 37.1 & 44.8 & N/A \\
\texttt{umb\_Latn} & 1 & 29.1 & 34.8 & 52.5 & 31.1 & 42.5 & 45.2 & 50.8 & 52.0 & 35.4 & 47.4 & N/A \\
    \bottomrule
    \end{tabular}
    \end{adjustbox}
    \caption{\textbf{Subsets of language tiers for \its @ $k{=}1$}.}
    \label{tab:subsets-full-results-i2s-k1}
\end{table*}

\begin{table*}[ht]
    \rowcolors{2}{gray!15}{white} 
    \begin{adjustbox}{max width=\textwidth}
    \centering
    \small
    \begin{tabular}{l c | c c c | c c c c c | c | c c}
    \toprule
    & & \multicolumn{3}{c|}{\textbf{Qwen2-VL}} & \multicolumn{5}{c|}{\textbf{InternVL 2.5}} & \multicolumn{1}{c|}{\textbf{Centurio Qwen}} & \multicolumn{2}{c}{\textbf{GPT-4o}} \\
    \cmidrule(lr){3-5} \cmidrule(lr){6-10}  \cmidrule(lr){11-11} \cmidrule(lr){12-13}
    \textbf{Lang} & \textbf{Tier} & Qwen2-VL 2B & Qwen2-VL 7B & Qwen2-VL 72B & InternVL 2.5 4B & InternVL 2.5 8B & InternVL 2.5 26B & InternVL 2.5 38B & InternVL 2.5 78B & Centurio Qwen 8B & GPT-4o-mini & GPT-4o \\
    \midrule
    \texttt{acm\_Arab} & 3 & 35.2 & 55.1 & 82.4 & 45.6 & 60.6 & 74.0 & 85.5 & 86.7 & 59.7 & 79.3 & 86.0 \\
    \texttt{aka\_Latn} & 2 & 27.9 & 38.3 & 63.2 & 34.2 & 46.3 & 52.9 & 61.1 & 60.6 & 42.6 & 63.3 & 79.3 \\
    \texttt{apc\_Arab} & 3 & 34.7 & 54.3 & 82.2 & 45.1 & 60.5 & 73.9 & 84.1 & 85.7 & 59.4 & 78.5 & 85.8 \\
    \texttt{arb\_Arab} & 4 & 35.2 & 56.4 & 81.8 & 46.9 & 60.9 & 76.0 & 85.0 & 87.1 & 58.7 & 79.7 & 86.7 \\
    \texttt{azj\_Latn} & 2 & 33.1 & 53.4 & 82.3 & 46.2 & 60.9 & 71.4 & 82.7 & 83.5 & 57.9 & 78.6 & 86.5 \\
    \texttt{bul\_Cyrl} & 3 & 34.7 & 54.6 & 82.7 & 45.9 & 64.4 & 78.8 & 84.8 & 84.5 & 58.8 & 79.5 & 86.4 \\
    \texttt{ces\_Latn} & 4 & 35.6 & 56.8 & 82.2 & 48.9 & 65.6 & 78.4 & 83.6 & 85.3 & 57.3 & 78.6 & 86.3 \\
    \texttt{eng\_Latn} & 4 & 34.8 & 63.1 & 82.6 & 49.2 & 67.9 & 80.1 & 84.9 & 85.5 & 60.0 & 78.1 & 87.1 \\
    \texttt{fin\_Latn} & 4 & 33.5 & 56.4 & 81.9 & 45.8 & 65.6 & 77.2 & 83.2 & 84.4 & 59.7 & 78.3 & 86.5 \\
    \texttt{hat\_Latn} & 1 & 30.7 & 52.7 & 78.9 & 42.1 & 58.1 & 67.9 & 77.6 & 79.3 & 58.0 & 77.1 & 85.1 \\
    \texttt{hau\_Latn} & 1 & 28.7 & 39.0 & 61.6 & 30.2 & 42.2 & 46.8 & 59.8 & 59.8 & 44.5 & 75.6 & 85.1 \\
    \texttt{min\_Arab} & 2 & 27.4 & 32.6 & 61.0 & 28.9 & 32.1 & 36.4 & 55.9 & 50.2 & 35.9 & 46.3 & 76.1 \\
    \texttt{umb\_Latn} & 1 & 28.9 & 34.8 & 58.7 & 30.5 & 42.2 & 45.8 & 52.6 & 53.5 & 35.8 & 46.8 & 61.1 \\
    \bottomrule
    \end{tabular}
    \end{adjustbox}
    \caption{\textbf{Subsets of language tiers for \its @ $k{=}3$}.}
    \label{tab:subsets-full-results-i2s-k3}
\end{table*}

\begin{table*}[ht]
    \rowcolors{2}{gray!15}{white} 
    \begin{adjustbox}{max width=\textwidth}
    \centering
    \small
    \begin{tabular}{l c | c c c | c c c c c | c | c c}
    \toprule
    & & \multicolumn{3}{c|}{\textbf{Qwen2-VL}} & \multicolumn{5}{c|}{\textbf{InternVL 2.5}} & \multicolumn{1}{c|}{\textbf{Centurio Qwen}} & \multicolumn{2}{c}{\textbf{GPT-4o}} \\
    \cmidrule(lr){3-5} \cmidrule(lr){6-10}  \cmidrule(lr){11-11} \cmidrule(lr){12-13}
    \textbf{Lang} & \textbf{Tier} & Qwen2-VL 2B & Qwen2-VL 7B & Qwen2-VL 72B & InternVL 2.5 4B & InternVL 2.5 8B & InternVL 2.5 26B & InternVL 2.5 38B & InternVL 2.5 78B & Centurio Qwen 8B & GPT-4o-mini & GPT-4o \\
    \midrule
    \texttt{acm\_Arab} & 3 & 33.7 & 48.7 & 83.2 & 47.6 & 60.6 & 73.9 & 85.8 & 87.1 & 61.1 & 77.3 & N/A \\
    \texttt{aka\_Latn} & 2 & 27.0 & 34.0 & 64.8 & 34.5 & 47.0 & 52.6 & 60.6 & 60.3 & 43.2 & 60.7 & N/A \\
    \texttt{apc\_Arab} & 3 & 32.7 & 48.9 & 83.9 & 46.2 & 61.2 & 73.6 & 84.7 & 87.3 & 60.3 & 77.4 & N/A \\
    \texttt{arb\_Arab} & 4 & 33.3 & 50.6 & 83.2 & 49.3 & 62.3 & 75.6 & 85.9 & 87.7 & 60.2 & 78.2 & N/A \\
    \texttt{azj\_Latn} & 2 & 32.2 & 47.9 & 82.2 & 45.5 & 63.0 & 72.8 & 83.6 & 85.3 & 59.4 & 78.1 & N/A \\
    \texttt{bul\_Cyrl} & 3 & 33.2 & 50.1 & 83.1 & 48.5 & 66.0 & 77.6 & 85.2 & 86.5 & 60.2 & 78.3 & N/A \\
    \texttt{ces\_Latn} & 4 & 34.9 & 52.2 & 82.2 & 48.6 & 66.0 & 78.8 & 84.7 & 86.4 & 58.9 & 79.3 & N/A \\
    \texttt{eng\_Latn} & 4 & 34.9 & 58.9 & 83.9 & 48.1 & 68.7 & N/A & 84.5 & 87.0 & 62.4 & 77.4 & N/A \\
    \texttt{fin\_Latn} & 4 & 32.6 & 51.1 & 82.2 & 46.8 & 66.6 & 76.3 & 83.7 & 86.3 & 60.7 & 77.0 & N/A \\
    \texttt{hat\_Latn} & 1 & 31.1 & 49.5 & 80.0 & 41.3 & 59.4 & 68.7 & 77.0 & 80.4 & 58.5 & 75.6 & N/A \\
    \texttt{hau\_Latn} & 1 & 28.4 & 35.9 & 63.3 & 31.4 & 42.8 & 46.2 & 60.5 & 60.4 & 44.7 & 73.9 & N/A \\
    \texttt{min\_Arab} & 2 & 27.4 & 30.1 & 62.0 & 28.5 & 32.3 & 36.3 & 55.7 & 49.8 & 36.4 & 43.9 & N/A \\
    \texttt{umb\_Latn} & 1 & 28.0 & 31.7 & 59.1 & 31.6 & 42.7 & 46.6 & 52.3 & 53.3 & 35.7 & 45.4 & N/A \\

    \bottomrule
    \end{tabular}
    \end{adjustbox}
    \caption{\textbf{Subsets of language tiers for \its @ $k{=}5$}.}
    \label{tab:subsets-full-results-i2s-k5}
\end{table*}

\begin{table*}[ht]
    \rowcolors{2}{gray!15}{white} 
    \begin{adjustbox}{max width=\textwidth}
    \centering
    \small
    \begin{tabular}{l c | c c c | c c c c c | c | c c}
    \toprule
    & & \multicolumn{3}{c|}{\textbf{Qwen2-VL}} & \multicolumn{5}{c|}{\textbf{InternVL 2.5}} & \multicolumn{1}{c|}{\textbf{Centurio Qwen}} & \multicolumn{2}{c}{\textbf{GPT-4o}} \\
    \cmidrule(lr){3-5} \cmidrule(lr){6-10}  \cmidrule(lr){11-11} \cmidrule(lr){12-13}
    \textbf{Lang} & \textbf{Tier} & Qwen2-VL 2B & Qwen2-VL 7B & Qwen2-VL 72B & InternVL 2.5 4B & InternVL 2.5 8B & InternVL 2.5 26B & InternVL 2.5 38B & InternVL 2.5 78B & Centurio Qwen 8B & GPT-4o-mini & GPT-4o \\
    \midrule
    \texttt{acm\_Arab} & 3 & 46.6 & 81.1 & 91.1 & 71.5 & 80.5 & 84.9 & 91.3 & 89.6 & 83.2 & 89.9 & 92.3 \\
    \texttt{aka\_Latn} & 2 & 32.3 & 53.2 & 68.1 & 44.4 & 59.6 & 64.4 & 68.6 & 66.5 & 63.5 & 79.9 & 87.7 \\
    \texttt{apc\_Arab} & 3 & 46.0 & 80.9 & 91.3 & 70.0 & 80.4 & 84.6 & 90.9 & 89.1 & 82.7 & 89.4 & 92.5 \\
    \texttt{arb\_Arab} & 4 & 48.1 & 82.7 & 91.4 & 73.1 & 81.3 & 86.5 & 91.7 & 90.2 & 84.9 & 89.9 & 92.8 \\
    \texttt{azj\_Latn} & 2 & 42.6 & 80.2 & 90.2 & 65.5 & 76.9 & 82.0 & 89.1 & 87.3 & 82.0 & 89.1 & 91.9 \\
    \texttt{bul\_Cyrl} & 3 & 47.1 & 83.3 & 91.2 & 73.4 & 83.5 & 88.2 & 90.6 & 89.2 & 83.9 & 89.3 & 92.5 \\
    \texttt{ces\_Latn} & 4 & 50.8 & 82.8 & 90.6 & 75.3 & 85.3 & 89.8 & 90.8 & 90.0 & 84.6 & 90.1 & 92.2 \\
    \texttt{eng\_Latn} & 4 & 56.7 & 85.7 & 91.8 & 81.4 & 87.0 & 91.3 & 91.8 & 91.5 & 85.4 & 88.5 & 92.0 \\
    \texttt{fin\_Latn} & 4 & 42.3 & 82.0 & 91.4 & 70.5 & 82.3 & 87.1 & 90.2 & 88.1 & 81.8 & 89.2 & 92.3 \\
    \texttt{hat\_Latn} & 1 & 40.2 & 70.8 & 87.3 & 57.9 & 74.1 & 79.2 & 84.9 & 83.7 & 82.0 & 87.3 & 91.6 \\
    \texttt{hau\_Latn} & 1 & 32.6 & 52.1 & 66.5 & 40.0 & 53.8 & 57.5 & 70.7 & 66.2 & 66.5 & 87.3 & 91.9 \\
    \texttt{min\_Arab} & 2 & 28.3 & 42.3 & 68.0 & 32.8 & 41.5 & 41.4 & 63.7 & 56.4 & 49.8 & 62.8 & 86.4 \\
    \texttt{umb\_Latn} & 1 & 30.3 & 47.1 & 61.4 & 39.3 & 52.4 & 57.2 & N/A & 59.6 & 53.4 & 62.9 & 71.7 \\
    \bottomrule
    \end{tabular}
    \end{adjustbox}
    \caption{\textbf{Subsets of language tiers for \tts @ $k{=}1$}.}
    \label{tab:subsets-full-results-t2s-k1}
\end{table*}

\begin{table*}[ht]
    \rowcolors{2}{gray!15}{white} 
    \begin{adjustbox}{max width=\textwidth}
    \centering
    \small
    \begin{tabular}{l c | c c c | c c c c c | c | c c}
    \toprule
    & & \multicolumn{3}{c|}{\textbf{Qwen2-VL}} & \multicolumn{5}{c|}{\textbf{InternVL 2.5}} & \multicolumn{1}{c|}{\textbf{Centurio Qwen}} & \multicolumn{2}{c}{\textbf{GPT-4o}} \\
    \cmidrule(lr){3-5} \cmidrule(lr){6-10}  \cmidrule(lr){11-11} \cmidrule(lr){12-13}
    \textbf{Lang} & \textbf{Tier} & Qwen2-VL 2B & Qwen2-VL 7B & Qwen2-VL 72B & InternVL 2.5 4B & InternVL 2.5 8B & InternVL 2.5 26B & InternVL 2.5 38B & InternVL 2.5 78B & Centurio Qwen 8B & GPT-4o-mini & GPT-4o \\
    \midrule
    \texttt{aeb\_Arab} & 3 & 43.1 & 62.9 & 80.1 & 36.8 & 53.2 & 68.7 & 79.3 & 77.6 & 29.8 & 76.3 & N/A \\
    \texttt{arb\_Arab} & 4 & 44.3 & 66.5 & 80.2 & 37.7 & 55.2 & 71.9 & 81.4 & 79.9 & 30.8 & 77.7 & N/A \\
    \texttt{ben\_Beng} & 3 & 39.8 & 64.4 & 81.2 & 41.2 & 50.9 & 69.0 & 79.1 & 77.2 & 33.0 & 76.4 & N/A \\
    \texttt{eng\_Latn} & 4 & 41.9 & 71.7 & N/A & 47.7 & 66.2 & N/A & 81.7 & 81.3 & 35.3 & 77.5 & N/A \\
    \texttt{fao\_Latn} & 2 & 33.6 & 52.2 & 75.1 & 30.9 & 43.9 & 56.8 & 73.6 & 69.0 & 29.2 & 75.8 & N/A \\
    \texttt{kac\_Latn} & 1 & 28.8 & 35.4 & 51.5 & 26.4 & 37.8 & 44.7 & 51.6 & 46.9 & 26.7 & 46.6 & N/A \\
    \texttt{kas\_Deva} & 2 & 33.3 & 47.1 & 72.4 & 30.2 & 42.4 & 53.0 & 69.0 & 59.3 & 27.7 & 66.5 & N/A \\
    \texttt{lit\_Latn} & 3 & 38.1 & 63.6 & 78.9 & 34.7 & 50.2 & 68.8 & 78.4 & 75.1 & 30.1 & 77.4 & N/A \\
    \texttt{lua\_Latn} & 1 & 31.0 & 43.8 & 57.6 & 29.3 & 41.7 & 48.8 & 57.4 & 55.9 & 27.7 & 54.9 & N/A \\
    \texttt{mal\_Mlym} & 2 & 32.1 & 64.3 & 80.1 & 34.1 & 47.0 & 62.3 & 77.0 & 73.2 & 31.0 & 76.2 & N/A \\
    \texttt{srp\_Cyrl} & 4 & 42.7 & 66.2 & 79.5 & 37.0 & 53.0 & 71.3 & 80.1 & 78.2 & 31.7 & 77.9 & N/A \\
    \texttt{tur\_Latn} & 4 & 42.7 & 66.6 & 78.5 & 38.7 & 56.4 & 74.4 & 80.5 & 77.6 & 31.0 & 77.4 & N/A \\
    \texttt{wol\_Latn} & 1 & 31.5 & 42.1 & 56.8 & 28.4 & 40.0 & 49.8 & 57.0 & 54.0 & 27.2 & 56.5 & N/A \\
    \bottomrule
    \end{tabular}
    \end{adjustbox}
    \caption{\textbf{Subsets of language tiers for \sti @ $k{=}1$}.}
    \label{tab:subsets-full-results-s2i-k1}
\end{table*}
\begin{table*}[ht]
    \rowcolors{2}{gray!15}{white} 
    \begin{adjustbox}{max width=\textwidth}
    \centering
    \small
    \begin{tabular}{l c | c c c | c c c c c | c | c c}
    \toprule
    & & \multicolumn{3}{c|}{\textbf{Qwen2-VL}} & \multicolumn{5}{c|}{\textbf{InternVL 2.5}} & \multicolumn{1}{c|}{\textbf{Centurio Qwen}} & \multicolumn{2}{c}{\textbf{GPT-4o}} \\
    \cmidrule(lr){3-5} \cmidrule(lr){6-10}  \cmidrule(lr){11-11} \cmidrule(lr){12-13}
    \textbf{Lang} & \textbf{Tier} & Qwen2-VL 2B & Qwen2-VL 7B & Qwen2-VL 72B & InternVL 2.5 4B & InternVL 2.5 8B & InternVL 2.5 26B & InternVL 2.5 38B & InternVL 2.5 78B & Centurio Qwen 8B & GPT-4o-mini & GPT-4o \\
    \midrule
    \texttt{aeb\_Arab} & 3 & 44.1 & 65.3 & 89.6 & 38.5 & 58.6 & 76.0 & 92.0 & 89.0 & 31.7 & 85.4 & 88.3 \\
    \texttt{arb\_Arab} & 4 & 44.8 & 67.6 & 89.1 & 39.2 & 59.8 & 78.9 & 92.4 & 90.1 & 32.8 & 86.8 & 89.1 \\
    \texttt{ben\_Beng} & 3 & 40.4 & 67.0 & 88.9 & 43.4 & 58.1 & 74.5 & 90.8 & 89.2 & 33.9 & 85.7 & 87.9 \\
    \texttt{eng\_Latn} & 4 & 43.1 & 70.4 & 88.5 & 44.5 & 69.0 & 84.0 & 92.1 & 90.0 & 36.1 & 86.4 & 89.1 \\
    \texttt{fao\_Latn} & 2 & 35.0 & 55.2 & 87.5 & 34.9 & 50.6 & 68.9 & 89.5 & 85.9 & 30.5 & 86.0 & 89.1 \\
    \texttt{kac\_Latn} & 1 & 31.6 & 41.0 & 69.7 & 29.2 & 46.3 & 57.5 & 72.3 & 68.0 & 27.4 & 64.2 & 70.5 \\
    \texttt{kas\_Deva} & 2 & 35.1 & 50.1 & 87.0 & 31.9 & 48.9 & 60.6 & 86.5 & 78.0 & 28.9 & 80.4 & 86.1 \\
    \texttt{lit\_Latn} & 3 & 42.1 & 64.7 & 89.1 & 39.2 & 58.5 & 77.4 & 91.0 & 89.4 & 32.2 & 86.2 & 89.0 \\
    \texttt{lua\_Latn} & 1 & 33.7 & 46.2 & 74.6 & 33.0 & 50.8 & 61.5 & 78.5 & 77.1 & 27.8 & 70.4 & 79.3 \\
    \texttt{mal\_Mlym} & 2 & 33.7 & 65.3 & 88.0 & 38.1 & 50.8 & 69.7 & 90.7 & 88.2 & 31.0 & 85.1 & 87.8 \\
    \texttt{srp\_Cyrl} & 4 & 44.1 & 65.1 & 88.7 & 39.7 & 59.4 & 78.3 & 91.8 & 89.8 & 32.6 & 87.1 & 89.5 \\
    \texttt{tur\_Latn} & 4 & 44.6 & 66.7 & 88.3 & 41.3 & 61.0 & 81.7 & 91.8 & 88.8 & 34.3 & 87.0 & 88.5 \\
    \texttt{wol\_Latn} & 1 & 34.8 & 45.8 & 75.2 & 32.9 & 49.7 & 62.3 & 79.2 & 77.0 & 28.1 & 72.7 & 83.1 \\
    \bottomrule
    \end{tabular}
    \end{adjustbox}
    \caption{\textbf{Subsets of language tiers for \sti @ $k{=}3$}.}
    \label{tab:subsets-full-results-s2i-k3}
\end{table*}

\begin{table*}[ht]
    \rowcolors{2}{gray!15}{white} 
    \begin{adjustbox}{max width=\textwidth}
    \centering
    \small
    \begin{tabular}{l c | c c c | c c c c c | c | c c}
    \toprule
    & & \multicolumn{3}{c|}{\textbf{Qwen2-VL}} & \multicolumn{5}{c|}{\textbf{InternVL 2.5}} & \multicolumn{1}{c|}{\textbf{Centurio Qwen}} & \multicolumn{2}{c}{\textbf{GPT-4o}} \\
    \cmidrule(lr){3-5} \cmidrule(lr){6-10}  \cmidrule(lr){11-11} \cmidrule(lr){12-13}
    \textbf{Lang} & \textbf{Tier} & Qwen2-VL 2B & Qwen2-VL 7B & Qwen2-VL 72B & InternVL 2.5 4B & InternVL 2.5 8B & InternVL 2.5 26B & InternVL 2.5 38B & InternVL 2.5 78B & Centurio Qwen 8B & GPT-4o-mini & GPT-4o \\
    \midrule
    \texttt{aeb\_Arab} & 3 & 44.0 & 65.4 & 91.6 & 39.0 & 57.6 & 77.0 & 93.6 & 91.8 & 32.8 & 88.9 & N/A \\
    \texttt{arb\_Arab} & 4 & 44.3 & 65.7 & 90.8 & 40.0 & 59.9 & 79.4 & 94.0 & 91.9 & 33.5 & 88.7 & N/A \\
    \texttt{ben\_Beng} & 3 & 41.1 & 63.2 & 89.3 & 45.9 & 59.4 & 76.4 & 93.0 & 91.5 & 34.8 & 88.4 & N/A \\
    \texttt{eng\_Latn} & 4 & 43.4 & 68.6 & 89.6 & 43.0 & 68.7 & N/A & 93.7 & 91.7 & 35.6 & 89.1 & N/A \\
    \texttt{fao\_Latn} & 2 & 35.3 & 56.6 & 90.4 & 36.0 & 50.8 & 71.8 & 92.8 & 89.5 & 30.7 & 87.8 & N/A \\
    \texttt{kac\_Latn} & 1 & 31.8 & 42.3 & 74.8 & 31.6 & 46.2 & 60.6 & 79.2 & 73.9 & 27.0 & 68.4 & N/A \\
    \texttt{kas\_Deva} & 2 & 35.3 & 48.8 & 89.0 & 34.1 & 50.0 & 64.7 & 91.3 & 82.9 & 29.4 & 83.2 & N/A \\
    \texttt{lit\_Latn} & 3 & 42.1 & 63.5 & 91.1 & 39.5 & 57.7 & 78.1 & 93.8 & 92.1 & 32.8 & 88.2 & N/A \\
    \texttt{lua\_Latn} & 1 & 34.6 & 48.4 & 80.2 & 34.6 & 49.7 & 63.1 & 85.1 & 82.8 & 28.1 & 75.9 & N/A \\
    \texttt{mal\_Mlym} & 2 & 31.6 & 62.7 & 88.9 & 39.0 & 51.8 & 69.3 & 93.2 & 91.7 & 31.5 & 87.1 & N/A \\
    \texttt{srp\_Cyrl} & 4 & 44.5 & 63.5 & 90.7 & 41.2 & 58.4 & 79.4 & 93.6 & 91.4 & 33.3 & 89.1 & N/A \\
    \texttt{tur\_Latn} & 4 & 45.6 & 64.8 & 90.3 & 40.9 & 59.9 & 81.4 & 93.3 & 91.3 & 34.8 & 88.5 & N/A \\
    \texttt{wol\_Latn} & 1 & 34.4 & 46.3 & 82.1 & 34.8 & 50.9 & 64.3 & 86.1 & 82.2 & 27.9 & 77.9 & N/A \\
    \bottomrule
    \end{tabular}
    \end{adjustbox}
    \caption{\textbf{Subsets of language tiers for \sti @ $k{=}5$}.}
    \label{tab:subsets-full-results-s2i-k5}
\end{table*}

\begin{table*}[ht]
    \rowcolors{2}{gray!15}{white} 
    \begin{adjustbox}{max width=\textwidth}
    \centering
    \small
    \begin{tabular}{l c | c c c | c c c c c | c | c c}
    \toprule
    & & \multicolumn{3}{c|}{\textbf{Qwen2-VL}} & \multicolumn{5}{c|}{\textbf{InternVL 2.5}} & \multicolumn{1}{c|}{\textbf{Centurio Qwen}} & \multicolumn{2}{c}{\textbf{GPT-4o}} \\
    \cmidrule(lr){3-5} \cmidrule(lr){6-10}  \cmidrule(lr){11-11} \cmidrule(lr){12-13}
    \textbf{Lang} & \textbf{Tier} & Qwen2-VL 2B & Qwen2-VL 7B & Qwen2-VL 72B & InternVL 2.5 4B & InternVL 2.5 8B & InternVL 2.5 26B & InternVL 2.5 38B & InternVL 2.5 78B & Centurio Qwen 8B & GPT-4o-mini & GPT-4o \\
    \midrule
    \texttt{aeb\_Arab} & 3 & 77.9 & 85.2 & 91.0 & 87.5 & 83.5 & 83.9 & 91.5 & 91.9 & 88.8 & 91.5 & N/A \\
    \texttt{arb\_Arab} & 4 & 81.5 & 86.5 & 91.7 & 88.9 & 86.1 & 88.0 & 92.8 & 93.4 & 89.8 & 92.7 & N/A \\
    \texttt{ben\_Beng} & 3 & 68.8 & 85.2 & 91.2 & 83.7 & 79.5 & 81.6 & 91.2 & 90.8 & 86.4 & 91.1 & N/A \\
    \texttt{eng\_Latn} & 4 & 86.1 & 89.1 & 91.0 & 90.6 & 91.7 & 92.0 & 92.4 & 93.5 & 89.9 & 92.4 & N/A \\
    \texttt{fao\_Latn} & 2 & 52.1 & 72.1 & 86.3 & 71.0 & 72.8 & 76.9 & 86.5 & 83.9 & 80.3 & 90.0 & N/A \\
    \texttt{kac\_Latn} & 1 & 41.3 & 51.5 & 58.8 & 55.2 & 54.6 & 55.7 & 57.4 & 58.0 & 54.7 & 58.2 & N/A \\
    \texttt{kas\_Deva} & 2 & 50.1 & 72.4 & 83.5 & 71.7 & 65.1 & 63.4 & 80.4 & 74.2 & 75.0 & 82.9 & N/A \\
    \texttt{lit\_Latn} & 3 & 67.2 & 81.7 & 90.1 & 76.6 & 77.7 & 84.6 & 89.6 & 89.6 & 84.2 & 91.1 & N/A \\
    \texttt{lua\_Latn} & 1 & 49.4 & 58.5 & 65.2 & 61.3 & 61.0 & 59.0 & 64.8 & 68.5 & 63.3 & 65.8 & N/A \\
    \texttt{mal\_Mlym} & 2 & 49.0 & 83.0 & 89.7 & 76.9 & 69.1 & 74.3 & 88.7 & 87.2 & 80.6 & 90.5 & N/A \\
    \texttt{srp\_Cyrl} & 4 & 77.3 & 86.1 & 91.0 & 84.8 & 85.6 & 87.0 & 91.5 & 92.3 & 88.2 & 91.7 & N/A \\
    \texttt{tur\_Latn} & 4 & 76.4 & 85.0 & 91.3 & 85.1 & 87.5 & 89.4 & 91.7 & 91.6 & 88.2 & 91.5 & N/A \\
    \texttt{wol\_Latn} & 1 & 50.4 & 59.1 & 66.8 & 63.3 & 64.0 & 62.0 & 66.6 & 67.6 & 65.5 & 69.5 & N/A \\
    \bottomrule
    \end{tabular}
    \end{adjustbox}
    \caption{\textbf{Subsets of language tiers for \stt @ $k{=}1$}.}
    \label{tab:subsets-full-results-s2t-k1}
\end{table*}

\begin{table*}[ht]
    \rowcolors{2}{gray!15}{white} 
    \begin{adjustbox}{max width=\textwidth}
    \centering
    \small
    \begin{tabular}{l c | c c c | c c c c c | c | c c}
    \toprule
    & & \multicolumn{3}{c|}{\textbf{Qwen2-VL}} & \multicolumn{5}{c|}{\textbf{InternVL 2.5}} & \multicolumn{1}{c|}{\textbf{Centurio Qwen}} & \multicolumn{2}{c}{\textbf{GPT-4o}} \\
    \cmidrule(lr){3-5} \cmidrule(lr){6-10}  \cmidrule(lr){11-11} \cmidrule(lr){12-13}
    \textbf{Lang} & \textbf{Tier} & Qwen2-VL 2B & Qwen2-VL 7B & Qwen2-VL 72B & InternVL 2.5 4B & InternVL 2.5 8B & InternVL 2.5 26B & InternVL 2.5 38B & InternVL 2.5 78B & Centurio Qwen 8B & GPT-4o-mini & GPT-4o \\
    \midrule
    \texttt{aeb\_Arab} & 3 & 94.1 & 93.9 & 98.7 & 97.2 & 95.3 & 96.3 & 98.8 & 99.0 & 96.6 & 98.4 & 98.4 \\
    \texttt{arb\_Arab} & 4 & 95.7 & 94.4 & 98.7 & 97.3 & 96.7 & 98.0 & 98.7 & 99.1 & 95.7 & 98.6 & 98.4 \\
    \texttt{ben\_Beng} & 3 & 86.9 & 93.9 & 98.7 & 95.3 & 93.0 & 94.7 & 98.6 & 98.4 & 96.5 & 98.5 & 98.3 \\
    \texttt{eng\_Latn} & 4 & 96.5 & 95.3 & 98.3 & 98.0 & 98.1 & 98.7 & 98.7 & 98.9 & 96.7 & 98.1 & 98.0 \\
    \texttt{fao\_Latn} & 2 & 69.4 & 86.6 & 97.6 & 88.7 & 87.1 & 92.6 & 97.2 & 97.2 & 94.1 & 98.2 & 98.3 \\
    \texttt{kac\_Latn} & 1 & 57.6 & 68.0 & 80.1 & 75.0 & 77.7 & 75.5 & 79.4 & 79.9 & 75.3 & 78.6 & 79.1 \\
    \texttt{kas\_Deva} & 2 & 68.6 & 88.4 & 96.5 & 86.9 & 81.2 & 82.3 & 95.4 & 91.4 & 90.5 & 95.9 & 96.7 \\
    \texttt{lit\_Latn} & 3 & 87.1 & 92.4 & 98.6 & 92.2 & 94.9 & 97.6 & 98.2 & 98.3 & 94.9 & 98.6 & 98.2 \\
    \texttt{lua\_Latn} & 1 & 66.7 & 76.9 & 86.2 & 82.1 & 83.0 & 82.1 & 86.5 & 88.3 & 81.4 & 87.4 & 92.2 \\
    \texttt{mal\_Mlym} & 2 & 61.1 & 93.3 & 98.7 & 91.2 & 85.5 & 90.7 & 98.3 & 97.7 & 94.0 & 98.3 & 98.1 \\
    \texttt{srp\_Cyrl} & 4 & 93.2 & 94.4 & 98.7 & 96.2 & 96.5 & 97.6 & 98.7 & 98.9 & 96.5 & 98.3 & 98.4 \\
    \texttt{tur\_Latn} & 4 & 92.5 & 94.9 & 98.8 & 96.6 & 97.2 & 98.6 & 98.6 & 99.0 & 96.0 & 98.6 & 98.1 \\
    \texttt{wol\_Latn} & 1 & 66.5 & 77.1 & 88.9 & 84.5 & 84.9 & 86.0 & 88.8 & 89.6 & 85.0 & 89.3 & 95.1 \\
    \bottomrule
    \end{tabular}
    \end{adjustbox}
    \caption{\textbf{Subsets of language tiers for \stt @ $k{=}3$}.}
    \label{tab:subsets-full-results-s2t-k3}
\end{table*}

\begin{table*}[ht]
    \rowcolors{2}{gray!15}{white} 
    \begin{adjustbox}{max width=\textwidth}
    \centering
    \small
    \begin{tabular}{l c | c c c | c c c c c | c | c c}
    \toprule
    & & \multicolumn{3}{c|}{\textbf{Qwen2-VL}} & \multicolumn{5}{c|}{\textbf{InternVL 2.5}} & \multicolumn{1}{c|}{\textbf{Centurio Qwen}} & \multicolumn{2}{c}{\textbf{GPT-4o}} \\
    \cmidrule(lr){3-5} \cmidrule(lr){6-10}  \cmidrule(lr){11-11} \cmidrule(lr){12-13}
    \textbf{Lang} & \textbf{Tier} & Qwen2-VL 2B & Qwen2-VL 7B & Qwen2-VL 72B & InternVL 2.5 4B & InternVL 2.5 8B & InternVL 2.5 26B & InternVL 2.5 38B & InternVL 2.5 78B & Centurio Qwen 8B & GPT-4o-mini & GPT-4o \\
    \midrule
    \texttt{aeb\_Arab} & 3 & 97.2 & 97.4 & 99.7 & 98.7 & 97.7 & 98.8 & 99.7 & 99.7 & 98.3 & 99.5 & N/A \\
    \texttt{arb\_Arab} & 4 & 97.6 & 97.4 & 99.6 & 98.6 & 98.5 & 99.3 & 99.7 & 99.7 & 97.6 & 99.3 & N/A \\
    \texttt{ben\_Beng} & 3 & 92.6 & 96.9 & 99.5 & 97.6 & 96.8 & 97.9 & 99.6 & 99.5 & 97.7 & 99.2 & N/A \\
    \texttt{eng\_Latn} & 4 & 98.2 & 97.5 & 99.2 & 99.1 & 99.0 & 99.5 & 99.4 & 99.6 & 97.7 & 99.1 & N/A \\
    \texttt{fao\_Latn} & 2 & 79.9 & 91.8 & 99.4 & 92.7 & 92.9 & 96.3 & 99.2 & 99.5 & 96.9 & 99.0 & N/A \\
    \texttt{kac\_Latn} & 1 & 68.0 & 75.0 & 87.4 & 81.5 & 86.0 & 83.9 & 87.1 & 87.3 & 80.9 & 86.3 & N/A \\
    \texttt{kas\_Deva} & 2 & 78.4 & 92.4 & 99.0 & 92.3 & 87.9 & 87.5 & 99.0 & 96.6 & 93.8 & 98.7 & N/A \\
    \texttt{lit\_Latn} & 3 & 93.8 & 96.0 & 99.4 & 95.9 & 98.2 & 99.3 & 99.4 & 99.7 & 97.2 & 99.4 & N/A \\
    \texttt{lua\_Latn} & 1 & 75.9 & 83.3 & 91.9 & 88.1 & 90.1 & 89.1 & 92.4 & 93.5 & 85.2 & 93.1 & N/A \\
    \texttt{mal\_Mlym} & 2 & 67.6 & 96.7 & 99.6 & 95.5 & 90.2 & 95.7 & 99.5 & 99.1 & 97.4 & 99.2 & N/A \\
    \texttt{srp\_Cyrl} & 4 & 97.2 & 97.4 & 99.4 & 98.0 & 99.0 & 99.4 & 99.6 & 99.6 & 98.2 & 99.4 & N/A \\
    \texttt{tur\_Latn} & 4 & 95.6 & 97.2 & 99.4 & 98.4 & 98.8 & 99.5 & 99.4 & 99.7 & 97.6 & 99.4 & N/A \\
    \texttt{wol\_Latn} & 1 & 77.5 & 84.3 & 94.5 & 90.4 & 92.7 & 92.9 & 95.2 & 95.7 & 91.1 & 94.9 & N/A \\

    \bottomrule
    \end{tabular}
    \end{adjustbox}
    \caption{\textbf{Subsets of language tiers for \stt @ $k{=}5$}.}
    \label{tab:subsets-full-results-s2t-k5}
\end{table*}

\begin{table*}[ht]
\subsection{Full Performance by Task, Model, and Language}
\label{app:per-language-results}
\subsubsection{Images-To-Topics}
\label{app:images-to-topics}
    \begin{adjustbox}{max width=\textwidth}

    \begin{tabular}{l *{7}{c}}

    & \multicolumn{7}{c}{Topics}\\
    \cmidrule{2-8}
Model & 
 Entertainment & 
 Geography & 
 Health &
 Politics &
 Science \& Tech. &
 Sports &
 Travel \\ \midrule
QwenVL-2.5-2B   &  99.4 & 83.1 & 100.0 &  99.9  & 100.0  & 100.0 &  95.7 \\
QwenVL-2.5-7B   & 100.0 & 92.9 & 100.0 & 100.0  &  99.8  & 100.0 & 100.0 \\
InternVL-2.5-4B &  99.5 & 84.5 & 100.0 & 100.0  &  99.7  & 100.0 & 100.0 \\
InternVL-2.5-8B & 100.0 & 90.0 & 100.0 & 100.0  &  97.5  & 100.0 & 100.0 \\
Centurio-Qwen   &  99.6 & 90.6 & 100.0 & 100.0  &  99.2  & 100.0 & 100.0 \\ \bottomrule
    \end{tabular} 
    \end{adjustbox}
    \caption{\textbf{Image-To-Topics.} For a reference image, the model must pick the correct topic out of 4 choices.}
    \label{tab:images-to-topics}
\end{table*}

\begin{table*}[ht]
    \subsubsection{Images-To-Sentences}
    \label{app:images-to-sentences}
    \begin{minipage}{0.48\textwidth}
        \centering
        \renewcommand{\arraystretch}{0.8} 
        \rowcolors{2}{gray!15}{white} 
        \begin{adjustbox}{max height=0.6\textheight, max width=\columnwidth}
            \begin{tabular}{l *{18}{c}}
                \toprule
                Lang.  & 
                \multicolumn{3}{c}{QwenVL-2 2B} & 
                \multicolumn{3}{c}{QwenVL-2 7B} & 
                \multicolumn{3}{c}{InternVL-2.5 4B} & 
                \multicolumn{3}{c}{InternVL-2.5 8B} & 
                \multicolumn{3}{c}{Centurio-Qwen} &
                \multicolumn{3}{c}{4o-mini} \\
                \cmidrule(lr){2-4} \cmidrule(lr){5-7} \cmidrule(lr){8-10} \cmidrule(lr){11-13} \cmidrule(lr){14-16} \cmidrule(lr){17-19}
                \textit{k} & 
                1 & 3 & 5 & 
                1 & 3 & 5 & 
                1 & 3 & 5 & 
                1 & 3 & 5 & 
                1 & 3 & 5 & 
                1 & 3 & 5 \\
                \midrule
                \texttt{Avg}       & 31.8 & 31.4 & 30.7 & 46.6 & 46.4 & 42.4 & 40.9 & 39.0 & 39.3 & 53.1 & 53.3 & 53.7 & 47.8 & 50.5 & 51.1 & 64.2 & 69.2 & 67.7 \\
                \texttt{eng\_Latn} & 36.3 & 34.8 & 34.9 & 65.8 & 63.1 & 58.9 & 52.5 & 49.2 & 48.1 & 67.7 & 67.9 & 68.7 & 54.8 & 60.0 & 62.4 & 68.3 & 78.1 & 77.4 \\
                \texttt{ace\_Arab} & 27.7 & 27.9 & 27.4 & 33.4 & 34.3 & 31.7 & 29.5 & 28.9 & 29.4 & 34.6 & 34.1 & 34.0 & 38.9 & 37.5 & 38.3 & 48.1 & 49.4 & 46.6 \\
                \texttt{ace\_Latn} & 32.4 & 33.1 & 32.0 & 45.5 & 45.7 & 42.0 & 39.9 & 38.7 & 38.4 & 55.7 & 57.1 & 58.0 & 46.1 & 51.7 & 51.5 & 62.4 & 67.0 & 65.3 \\
                \texttt{acm\_Arab} & 34.9 & 35.2 & 33.7 & 55.4 & 55.1 & 48.7 & 50.0 & 45.6 & 47.6 & 61.0 & 60.6 & 60.6 & 54.4 & 59.7 & 61.1 & 72.0 & 79.3 & 77.3 \\
                \texttt{acq\_Arab} & 35.2 & 35.0 & 32.8 & 56.4 & 55.2 & 50.0 & 50.8 & 45.8 & 48.3 & 61.8 & 60.8 & 61.0 & 54.1 & 59.1 & 60.9 & 72.6 & 79.5 & 78.0 \\
\bottomrule
            \end{tabular}
        \end{adjustbox}
    \end{minipage}
    \hfill
    \begin{minipage}{0.48\textwidth}
        \centering
        \renewcommand{\arraystretch}{0.8} 
        \rowcolors{2}{gray!15}{white} 
        \begin{adjustbox}{max height=0.6\textheight, max width=\columnwidth}
            \begin{tabular}{l *{18}{c}}
                \toprule
                Lang.  & 
                \multicolumn{3}{c}{QwenVL-2 2B} & 
                \multicolumn{3}{c}{QwenVL-2 7B} & 
                \multicolumn{3}{c}{InternVL-2.5 4B} & 
                \multicolumn{3}{c}{InternVL-2.5 8B} & 
                \multicolumn{3}{c}{Centurio-Qwen} &
                \multicolumn{3}{c}{4o-mini} \\
                \cmidrule(lr){2-4} \cmidrule(lr){5-7} \cmidrule(lr){8-10} \cmidrule(lr){11-13} \cmidrule(lr){14-16} \cmidrule(lr){17-19}
                \textit{k} & 
                1 & 3 & 5 & 
                1 & 3 & 5 & 
                1 & 3 & 5 & 
                1 & 3 & 5 & 
                1 & 3 & 5 & 
                1 & 3 & 5 \\
                \midrule
                \texttt{lao\_Laoo} & 26.6 & 27.0 & 26.0 & 36.8 & 36.4 & 34.0 & 39.6 & 38.0 & 37.5 & 46.5 & 43.7 & 41.1 & 46.8 & 46.2 & 47.0 & 53.4 & 54.8 & 50.6 \\
                \texttt{lij\_Latn} & 33.2 & 31.8 & 31.5 & 49.8 & 50.1 & 46.7 & 43.1 & 41.3 & 42.7 & 56.4 & 58.6 & 59.1 & 51.1 & 55.8 & 55.2 & 66.3 & 71.1 & 69.7 \\
                \texttt{lim\_Latn} & 33.1 & 31.9 & 31.7 & 50.3 & 50.7 & 47.0 & 42.0 & 38.8 & 39.7 & 56.6 & 58.0 & 60.1 & 51.5 & 55.9 & 57.9 & 66.8 & 71.9 & 69.6 \\
                \texttt{lin\_Latn} & 30.5 & 30.4 & 30.9 & 40.7 & 42.2 & 39.5 & 38.0 & 36.7 & 36.8 & 49.7 & 50.9 & 52.1 & 48.8 & 52.0 & 52.8 & 65.1 & 68.1 & 66.8 \\
                \texttt{lit\_Latn} & 34.9 & 34.4 & 33.5 & 56.9 & 56.6 & 52.3 & 49.9 & 45.1 & 46.6 & 61.2 & 61.5 & 62.8 & 52.2 & 57.5 & 57.9 & 72.0 & 77.2 & 77.6 \\
                \texttt{lmo\_Latn} & 33.9 & 32.7 & 31.3 & 52.2 & 52.7 & 48.1 & 43.2 & 41.6 & 41.4 & 57.0 & 60.8 & 62.0 & 51.7 & 56.9 & 57.3 & 68.2 & 73.5 & 70.9 \\
\bottomrule
            \end{tabular}
        \end{adjustbox}
    \end{minipage}
\end{table*}

\begin{table*}[ht]
    \subsubsection{Topic-To-Sentence}
    \label{app:topic-to-sentences}
    \begin{minipage}{0.48\textwidth}
        \centering
        \renewcommand{\arraystretch}{0.8} 
        \rowcolors{2}{gray!15}{white} 
        \begin{adjustbox}{max height=0.6\textheight, max width=\columnwidth}
            \begin{tabular}{l *{6}{c}}
                \toprule
                Lang.  & 
                \multicolumn{1}{c}{QwenVL-2 2B} & 
                \multicolumn{1}{c}{QwenVL-2 7B} & 
                \multicolumn{1}{c}{InternVL-2.5 4B} & 
                \multicolumn{1}{c}{InternVL-2.5 8B} & 
                \multicolumn{1}{c}{Centurio-Qwen} &
                \multicolumn{1}{c}{4o-mini} \\
                \midrule
\texttt{Avg}       & 38.6 & 66.5 & 54.5 & 68.0 & 72.4 & 81.5 \\
\texttt{eng\_Latn} & 56.7 & 85.7 & 81.4 & 87.0 & 85.4 & 88.5 \\
\texttt{ace\_Arab} & 29.9 & 46.4 & 34.9 & 42.1 & 54.1 & 66.3 \\
\texttt{ace\_Latn} & 39.6 & 68.7 & 52.0 & 71.8 & 72.9 & 80.6 \\
\texttt{acm\_Arab} & 46.6 & 81.1 & 71.5 & 80.5 & 83.2 & 89.9 \\
\texttt{acq\_Arab} & 47.5 & 81.6 & 71.3 & 80.6 & 84.0 & 90.0 \\
\texttt{aeb\_Arab} & 45.7 & 79.6 & 68.4 & 78.2 & 82.1 & 88.9 \\
\bottomrule
            \end{tabular}
        \end{adjustbox}
    \end{minipage}
    \hfill
    \begin{minipage}{0.48\textwidth}
        \centering
        \renewcommand{\arraystretch}{0.8} 
        \rowcolors{2}{gray!15}{white} 
        \begin{adjustbox}{max height=0.6\textheight, max width=\columnwidth}
            \begin{tabular}{l *{6}{c}}
                \toprule
                Lang.  & 
                \multicolumn{1}{c}{QwenVL-2 2B} & 
                \multicolumn{1}{c}{QwenVL-2 7B} & 
                \multicolumn{1}{c}{InternVL-2.5 4B} & 
                \multicolumn{1}{c}{InternVL-2.5 8B} & 
                \multicolumn{1}{c}{Centurio-Qwen} &
                \multicolumn{1}{c}{4o-mini} \\
                \midrule
\texttt{lao\_Laoo} & 28.2 & 53.1 & 47.6 & 57.8 & 74.6 & 80.1 \\
\texttt{lij\_Latn} & 42.2 & 75.8 & 60.3 & 76.7 & 79.2 & 86.4 \\
\texttt{lim\_Latn} & 42.5 & 77.8 & 61.9 & 75.7 & 80.0 & 85.1 \\
\texttt{lin\_Latn} & 34.6 & 59.4 & 46.5 & 62.5 & 75.6 & 81.2 \\
\texttt{lit\_Latn} & 44.6 & 81.3 & 63.7 & 77.3 & 80.3 & 89.6 \\
\texttt{lmo\_Latn} & 44.1 & 77.2 & 61.3 & 77.4 & 79.9 & 85.9 \\
\texttt{ltg\_Latn} & 38.2 & 71.0 & 54.6 & 72.6 & 75.4 & 83.4 \\
\bottomrule
            \end{tabular}
        \end{adjustbox}
    \end{minipage}
\end{table*}

\begin{table*}[ht]
    \subsubsection{Sentences-To-Images}
    \label{app:sentences-to-images}
    \begin{minipage}{0.48\textwidth}
        \centering
        \renewcommand{\arraystretch}{0.8} 
        \rowcolors{2}{gray!15}{white} 
        \begin{adjustbox}{max height=0.6\textheight, max width=\columnwidth}
            \begin{tabular}{l *{18}{c}}
                \toprule
                Lang.  & 
                \multicolumn{3}{c}{QwenVL-2 2B} & 
                \multicolumn{3}{c}{QwenVL-2 7B} & 
                \multicolumn{3}{c}{InternVL-2.5 4B} & 
                \multicolumn{3}{c}{InternVL-2.5 8B} & 
                \multicolumn{3}{c}{Centurio-Qwen} &
                \multicolumn{3}{c}{4o-mini} \\
                \cmidrule(lr){2-4} \cmidrule(lr){5-7} \cmidrule(lr){8-10} \cmidrule(lr){11-13} \cmidrule(lr){14-16} \cmidrule(lr){17-19}
                \textit{k} & 
                1 & 3 & 5 & 
                1 & 3 & 5 & 
                1 & 3 & 5 & 
                1 & 3 & 5 & 
                1 & 3 & 5 & 
                1 & 3 & 5 \\
                \midrule
                \texttt{Avg}       & 35.1 & 36.8 & 36.8 & 52.8 & 54.5 & 54.4 & 32.2 & 35.4 & 36.6 & 45.8 & 51.8 & 51.7 & 29.2 & 30.3 & 30.5 & 69.2 & 80.2 & 83.0 \\
                \texttt{eng\_Latn} & 41.9 & 43.1 & 43.4 & 71.7 & 70.4 & 68.6 & 47.7 & 44.5 & 43.0 & 66.2 & 69.0 & 68.7 & 35.3 & 36.1 & 35.6 & 77.5 & 86.4 & 89.1 \\
                \texttt{ace\_Arab} & 29.2 & 30.4 & 30.6 & 36.7 & 38.8 & 41.8 & 26.3 & 28.2 & 30.9 & 29.7 & 31.8 & 30.7 & 26.8 & 26.5 & 26.4 & 53.7 & 70.0 & 73.6 \\
                \texttt{ace\_Latn} & 36.5 & 38.0 & 38.0 & 52.5 & 55.8 & 56.9 & 31.2 & 36.9 & 38.2 & 45.5 & 54.1 & 54.6 & 28.5 & 29.4 & 29.4 & 66.5 & 80.9 & 84.9 \\
                \texttt{acm\_Arab} & 44.1 & 45.3 & 43.9 & 63.8 & 66.6 & 65.2 & 37.9 & 39.0 & 39.9 & 54.1 & 59.6 & 59.7 & 30.5 & 33.2 & 33.5 & 77.5 & 86.6 & 88.8 \\
                \texttt{acq\_Arab} & 44.6 & 45.6 & 44.4 & 64.7 & 66.1 & 65.0 & 37.6 & 38.7 & 39.9 & 54.7 & 60.4 & 59.9 & 30.1 & 32.6 & 33.5 & 77.2 & 86.6 & 88.6 \\
\bottomrule
            \end{tabular}
        \end{adjustbox}
    \end{minipage}
    \hfill
    \begin{minipage}{0.48\textwidth}
        \centering
        \renewcommand{\arraystretch}{0.8} 
        \rowcolors{2}{gray!15}{white} 
        \begin{adjustbox}{max height=0.6\textheight, max width=\columnwidth}
            \begin{tabular}{l *{18}{c}}
                \toprule
                Lang.  & 
                \multicolumn{3}{c}{InternVL-2.5-4B} & 
                \multicolumn{3}{c}{InternVL-2.5-4B} & 
                \multicolumn{3}{c}{QwenVL-2.5-2B} & 
                \multicolumn{3}{c}{QwenVL-2.5-2B} & 
                \multicolumn{3}{c}{Centurio-Qwen} &
                \multicolumn{3}{c}{4o-mini} \\
                \cmidrule(lr){2-4} \cmidrule(lr){5-7} \cmidrule(lr){8-10} \cmidrule(lr){11-13} \cmidrule(lr){14-16} \cmidrule(lr){17-19}
                \textit{k} & 
                1 & 3 & 5 & 
                1 & 3 & 5 & 
                1 & 3 & 5 & 
                1 & 3 & 5 & 
                1 & 3 & 5 & 
                1 & 3 & 5 \\
                \midrule
                \texttt{lao\_Laoo} & 28.1 & 28.7 & 28.0 & 40.8 & 42.6 & 40.3 & 30.2 & 33.4 & 34.3 & 39.4 & 44.1 & 44.6 & 29.9 & 28.9 & 29.2 & 67.7 & 77.8 & 79.5 \\
                \texttt{lij\_Latn} & 39.5 & 42.7 & 43.3 & 61.6 & 61.9 & 63.1 & 34.8 & 37.6 & 38.5 & 49.6 & 56.9 & 57.3 & 29.7 & 31.3 & 31.2 & 73.9 & 85.0 & 86.9 \\
                \texttt{lim\_Latn} & 39.1 & 42.3 & 41.4 & 60.5 & 62.6 & 62.8 & 33.5 & 38.6 & 38.7 & 51.0 & 59.0 & 58.4 & 29.4 & 31.0 & 31.0 & 73.3 & 84.7 & 87.3 \\
                \texttt{lin\_Latn} & 31.8 & 34.7 & 34.8 & 46.1 & 49.2 & 51.0 & 28.8 & 34.0 & 34.3 & 41.8 & 51.1 & 50.5 & 29.0 & 30.3 & 30.0 & 67.2 & 81.6 & 85.5 \\
                \texttt{lit\_Latn} & 38.1 & 42.1 & 42.1 & 63.6 & 64.7 & 63.5 & 34.7 & 39.2 & 39.5 & 50.2 & 58.5 & 57.7 & 30.1 & 32.2 & 32.8 & 77.4 & 86.2 & 88.2 \\
                \texttt{lmo\_Latn} & 39.8 & 41.4 & 41.4 & 61.1 & 62.9 & 63.2 & 34.2 & 38.4 & 39.6 & 50.4 & 58.6 & 58.0 & 29.4 & 31.3 & 31.8 & 74.2 & 84.8 & 87.9 \\
\bottomrule
            \end{tabular}
        \end{adjustbox}
    \end{minipage}
\end{table*}

\begin{table*}[ht]
    \subsubsection{Sentences-To-Topics}
    \label{app:sentences-to-topics}
    \begin{minipage}{0.48\textwidth}
        \centering
        \renewcommand{\arraystretch}{0.8} 
        \rowcolors{2}{gray!15}{white} 
        \begin{adjustbox}{max height=0.6\textheight, max width=\columnwidth}
            \begin{tabular}{l *{18}{c}}
                \toprule
                Lang.  & 
                \multicolumn{3}{c}{InternVL-2.5-4B} & 
                \multicolumn{3}{c}{InternVL-2.5-4B} & 
                \multicolumn{3}{c}{QwenVL-2.5-2B} & 
                \multicolumn{3}{c}{QwenVL-2.5-2B} & 
                \multicolumn{3}{c}{Centurio-Qwen} &
                \multicolumn{3}{c}{4o-mini} \\
                \cmidrule(lr){2-4} \cmidrule(lr){5-7} \cmidrule(lr){8-10} \cmidrule(lr){11-13} \cmidrule(lr){14-16} \cmidrule(lr){17-19}
                \textit{k} & 
                1 & 3 & 5 & 
                1 & 3 & 5 & 
                1 & 3 & 5 & 
                1 & 3 & 5 & 
                1 & 3 & 5 & 
                1 & 3 & 5 \\
                \midrule
                \texttt{Avg} & 58.2 & 73.2 & 79.5 & 70.3 & 82.8 & 87.7 & 70.8 & 85.4 & 89.5 & 69.6 & 83.8 & 88.4 & 75.5 & 88.0 & 91.5 & 82.9 & 93.5 & 95.8 \\
\texttt{eng\_Latn} & 86.1 & 96.5 & 98.2 & 89.1 & 95.3 & 97.5 & 90.6 & 98.0 & 99.1 & 91.7 & 98.1 & 99.0 & 89.9 & 96.7 & 97.7 & 92.4 & 98.1 & 99.1 \\
\texttt{ace\_Arab} & 36.9 & 48.1 & 54.6 & 55.2 & 70.0 & 77.9 & 52.5 & 69.0 & 73.6 & 42.3 & 50.9 & 54.3 & 58.1 & 75.5 & 82.1 & 66.8 & 85.4 & 91.6 \\
\texttt{ace\_Latn} & 59.8 & 79.6 & 87.2 & 71.8 & 86.6 & 92.6 & 74.6 & 91.4 & 96.2 & 74.4 & 92.6 & 96.9 & 76.3 & 91.5 & 95.7 & 80.5 & 94.9 & 98.2 \\
\texttt{acm\_Arab} & 80.8 & 95.2 & 97.4 & 85.7 & 94.2 & 97.4 & 88.0 & 97.1 & 98.6 & 84.9 & 95.9 & 98.5 & 90.6 & 96.2 & 98.0 & 92.2 & 98.5 & 99.4 \\
\texttt{acq\_Arab} & 81.3 & 95.3 & 97.4 & 86.2 & 94.0 & 96.9 & 88.6 & 97.2 & 98.7 & 85.5 & 96.5 & 98.5 & 89.9 & 95.9 & 97.8 & 92.2 & 98.7 & 99.4 \\
\bottomrule
            \end{tabular}
        \end{adjustbox}
    \end{minipage}
    \hfill
    \begin{minipage}{0.48\textwidth}
        \centering
        \renewcommand{\arraystretch}{0.8} 
        \rowcolors{2}{gray!15}{white} 
        \begin{adjustbox}{max height=0.6\textheight, max width=\columnwidth}
            \begin{tabular}{l *{18}{c}}
                \toprule
                Lang.  & 
                \multicolumn{3}{c}{InternVL-2.5-4B} & 
                \multicolumn{3}{c}{InternVL-2.5-4B} & 
                \multicolumn{3}{c}{QwenVL-2.5-2B} & 
                \multicolumn{3}{c}{QwenVL-2.5-2B} & 
                \multicolumn{3}{c}{Centurio-Qwen} &
                \multicolumn{3}{c}{4o-mini} \\
                \cmidrule(lr){2-4} \cmidrule(lr){5-7} \cmidrule(lr){8-10} \cmidrule(lr){11-13} \cmidrule(lr){14-16} \cmidrule(lr){17-19}
                \textit{k} & 
                1 & 3 & 5 & 
                1 & 3 & 5 & 
                1 & 3 & 5 & 
                1 & 3 & 5 & 
                1 & 3 & 5 & 
                1 & 3 & 5 \\
                \midrule
\texttt{lao\_Laoo} & 38.6 & 52.4 & 59.0 & 61.2 & 75.9 & 82.9 & 67.8 & 82.4 & 87.9 & 55.3 & 72.5 & 80.7 & 77.5 & 91.4 & 94.8 & 83.8 & 94.4 & 97.3 \\
\texttt{lij\_Latn} & 72.5 & 89.3 & 94.3 & 80.4 & 91.7 & 96.2 & 82.6 & 95.3 & 98.1 & 83.0 & 95.5 & 98.7 & 83.8 & 95.7 & 98.0 & 88.4 & 97.3 & 99.0 \\
\texttt{lim\_Latn} & 70.2 & 89.4 & 94.8 & 78.5 & 92.0 & 95.9 & 82.1 & 95.8 & 97.8 & 80.8 & 95.2 & 98.4 & 82.7 & 96.1 & 98.2 & 87.5 & 97.5 & 99.1 \\
\texttt{lin\_Latn} & 50.9 & 67.4 & 77.3 & 61.4 & 78.3 & 84.1 & 65.4 & 84.0 & 89.6 & 62.7 & 83.8 & 91.7 & 75.4 & 89.8 & 93.3 & 80.3 & 95.7 & 98.5 \\
\texttt{lit\_Latn} & 67.2 & 87.1 & 93.8 & 81.7 & 92.4 & 96.0 & 76.6 & 92.2 & 95.9 & 77.7 & 94.9 & 98.2 & 84.2 & 94.9 & 97.2 & 91.1 & 98.6 & 99.4 \\
\texttt{lmo\_Latn} & 72.1 & 90.6 & 95.2 & 81.2 & 93.0 & 96.4 & 83.7 & 96.0 & 98.6 & 83.3 & 95.8 & 98.7 & 84.3 & 96.0 & 98.0 & 89.1 & 98.0 & 99.2 \\
\bottomrule
            \end{tabular}
        \end{adjustbox}
    \end{minipage}
\end{table*}

\end{document}